\documentclass[11pt]{article}
\usepackage{graphicx}
\usepackage{amssymb}
\usepackage{amscd}
\usepackage{mathrsfs}
\usepackage{longtable,lscape}
\usepackage{amsthm}
\usepackage{amsfonts}
\usepackage{amsmath}
\usepackage{bbm}
\usepackage{float}
\usepackage{url}

\newcommand{\compl}{{\mathbb C}}

\newcommand{\captionfonts}{\footnotesize}
\makeatletter  
\long\def\@makecaption#1#2{%
  \vskip\abovecaptionskip
  \sbox\@tempboxa{{\captionfonts #1: #2}}%
  \ifdim \wd\@tempboxa >\hsize
    {\captionfonts #1: #2\par}
  \else
    \hbox to\hsize{\hfil\box\@tempboxa\hfil}%
  \fi
  \vskip\belowcaptionskip}
\makeatother 
\begin{document}
\title{Quantum Structure of Negation and Conjunction \\ in Human Thought}
\author{Diederik Aerts$^1$, Sandro Sozzo$^{2}$ and Tomas Veloz$^{3}$ \vspace{0.5 cm} \\ 
        \normalsize\itshape
        $^1$ Center Leo Apostel for Interdisciplinary Studies, Brussels Free University \\ 
        \normalsize\itshape
         Krijgskundestraat 33, 1160 Brussels, Belgium \\
        \normalsize
        E-Mail: \url{diraerts@vub.ac.be}
          \vspace{0.5 cm} \\ 
        \normalsize\itshape
        $^2$ School of Management and IQSCS, University of Leicester \\ 
        \normalsize\itshape
         University Road, LE1 7RH Leicester, United Kingdom \\
        \normalsize
        E-Mail: \url{ss831@le.ac.uk}
          \vspace{0.5 cm} \\ 
        \normalsize\itshape
        $^3$ Department of Mathematics, University of British Columbia, Okanagan Campus \\
        \normalsize\itshape
        3333 University Way, Kelowna, BC Canada V1V 1V7
        \\
        \normalsize\itshape
        and, Instituto de Filosof\'ia y Ciencias de la Complejidad (IFICC)  \\
        \normalsize\itshape
        Los Alerces 3024, \~Nu\~noa, Santiago, Chile  \\
        \normalsize
        E-Mail: \url{tomas.veloz@ubc.ca} \\
              }
\date{}
\maketitle
\begin{abstract}
\noindent
We analyse in this paper the data collected in a set of experiments performed on human subjects on the combination of natural concepts. We investigate the mutual influence of conceptual conjunction and negation by measuring the membership weights of a list of exemplars with respect to two concepts, e.g., {\it Fruits} and {\it Vegetables}, and their conjunction {\it Fruits And Vegetables}, but also their conjunction when one or both concepts are negated, namely, {\it Fruits And Not Vegetables}, {\it Not Fruits And Vegetables} and {\it Not Fruits And Not Vegetables}. Our findings sharpen existing analysis on conceptual combinations, revealing systematic and remarkable deviations from classical (fuzzy set) logic and probability theory. And, more important, our results give further considerable evidence to the validity of our quantum-theoretic framework for the combination of two concepts. Indeed, the representation of conceptual negation naturally arises from the general assumptions of our two-sector Fock space model, and this representation faithfully agrees with the collected data. In addition, we find a further significant deviation and a priori unexpected from classicality, which can exactly be explained by assuming that human reasoning is the superposition of an `emergent reasoning' and a `logical reasoning', and that these two processes can be successfully represented in a Fock space algebraic structure.
\end{abstract}

\section{Introduction\label{intro}}
During the last decade there has been an increasing evidence of  presence of quantum structures in 
processes that find their origin in human behaviour and cognition,
more specifically, in situations of decision making and in the structure of language (see, e.g., \cite{aa1995,aabg2000,ac2004,vr2004,ag2005a,ag2005b,a2009a,a2009b,pb2009,k2010,as2011,Aerts2011b,Aerts2012,bpft2011,bb2012,abgs2013,ags2013,hk2013,pb2013,wbap2013,as2014,ast2014,s2014a,pnas,plosone,s2014b,as2014b}).
 The success of this quantum modeling is interpreted as
due to `descriptive effectiveness of the mathematical apparatus of quantum theory as formal instrument to model cognitive dynamics and structures in situations where classical set-based approaches are problematical', without 
a priori a direct or precise connection with the validity of quantum laws in the microscopic world,
although also recently a reflection connecting the quantum modeling in the micro-world with these new quantum cognition approaches has been put forward \cite{as2014b}. In particular, the mathematics of quantum theory in Hilbert space has proved very successful in modeling conceptual combinations, 
i.e. conjunctions and disjunctions of two concepts (see, e.g., 
 \cite{ag2005a,ag2005b,a2009a,a2009b,as2011,abgs2013,ags2013,as2014,s2014a,s2014b}).

The `combination problem', that is, the problem of how the combination of two or more natural concepts can be represented starting from the representation of the component concepts, has been studied experimentally and within classical concept theories in great detail in the last thirty years. More specifically:

(i) The `Guppy effect' in concept conjunction, also known as the `Pet-Fish problem' \cite{os1981,os1982}. If one measures the typicality of specific exemplars with respect to the concepts {\it Pet} and {\it Fish} and their conjunction {\it Pet-Fish}, then one experimentally finds that an exemplar such as {\it Guppy} is a very typical example of {\it Pet-Fish}, while it is neither a very typical example of {\it Pet} nor of {\it Fish}.

(ii) The deviation from classical (fuzzy) set-theoretic membership weights of exemplars with respect to pairs of concepts and their conjunction or disjunction \cite{h1988a,h1988b}. If one measures the membership weight of an exemplar with respect to a pair of concepts and their conjunction (disjunction), then one experimentally finds that the membership weight of the exemplar with respect to the conjunction (disjunction) is greater (less) than the membership weight of the exemplar with respect to at least one of the component concepts.

(iii) The existence of `borderline contradictions' in sentences expressing vague properties \cite{bovw1999,ap2011}. Roughly speaking, a borderline contradiction is a sentence of the form ${\mathscr P}(x) \land \lnot {\mathscr P}(x)$, for a vague predicate ${\mathscr P}$ and a borderline case $x$, e.g., the sentence ``John is tall and John is not tall''.

What one typically finds in the above situations is a failure of set-theoretic approaches (classical set, fuzzy set, Kolmogorovian probability) to supply satisfactory theoretic models for the experimentally observed patterns. Indeed, all traditional approaches to concept theory (mainly, `prototype theory' \cite{r1973,r1978,r1983}, `exemplar theory' \cite{n1988,n1992} and `theory theory'  \cite{mm1985,rn1988}) and concept representation (mainly, `extensional' membership-based \cite{z1982,r1995} and `intensional' attribute-based \cite{h1988b,m1975,h1997}) 
have structural difficulties to cope with the experimental data exactly where  
the `graded', or `vague' nature of these data violates (fuzzy) set theoretic structures abundantly
\cite{os1982,z1982,z1965}, indicating that this violation of set theoretic structures is the core of the problem. 
More specifically, none of the traditional approaches, 
when representing the membership weights and typicalities expressing such gradeness in a classical (fuzzy) set-theoretic model, where conceptual conjunctions are represented by logical conjunctions and conceptual disjunctions are represented by logical disjunctions,
puts forward even moderately successful models with respect to the experimental data.
This situation is experienced as one of the major problems in the domain of traditional concept theories and an obstacle for progress 
 \cite{r1995,h1997,k1992,f1994,kp1995,os1997}.

Important results in concept research and modeling have been obtained in the last decade  within the approach of `quantum cognition' in which our research group has contributed substantially. The fundamentals of our approach can be resumed in the following progressive steps.

(a) The structural aspects of the approach rest on the results of older 
research on the foundations of quantum theory \cite{a1999}, the origins of quantum probability \cite{a1986,p1989} and the identification of typically quantum aspects outside the microscopic domain of quantum physics \cite{aa1995,aabg2000}. 

(b) A first major step was taken in considering a concept as an `entity in a specific state changing under the influence of a context', rather than as a `container of instantiations', and this led to the development of a {\it `State Context Property' formalism} ({\it SCoP}), and allowed the authors to provide a quantum representation of the guppy effect \cite{ag2005a,ag2005b}. 

(c) Continuing in this approach to concepts representation, the mathematical formalism of quantum theory was employed to model the overextension and underextension of membership weights measured in \cite{h1988a,h1988b}. More specifically, the overextension for conjunctions of concepts measured in \cite{h1988a} was described as an effect of quantum interference,
due to quantum superposition \cite{a2009a,ags2013}, which also  plays a primary role in the description of both overextension and underextension for disjunctions of concepts \cite{h1988b}.

(d) A two-sector Fock space structure enabled a complete representation of data on conjunctions and disjunctions of two concepts 
\cite{a2009a,ags2013}.

(e) Specific conceptual combinations experimentally revealed the presence of further genuine quantum effects, namely, entanglement  \cite{a2009b,as2011,abgs2013,ags2013,as2014} and quantum-type indistinguishability \cite{IQSA1}.

(f) This quantum-theoretic framework was successfully applied to describe borderline vagueness \cite{s2014a} and the effects of negation on conceptual conjunction \cite{s2014b}.

(g) Other phenomena related to concept combination, such as `Ellsberg and Machina decision making paradoxes    \cite{e1961,m2009} were sucessfully modeled in the same quantum-conceptual framework \cite{Aerts2011b,Aerts2012}.   

In the present paper we extend the collection of data in \cite{s2014b} with the aim of further exploring the use of negation in conceptual combinations and, more generally,  the underlying logical structures being at work in human thought in
the course of cognitive processes \cite{IQSA2}.  Let us first put forward a specific comment with respect to the `negation of a concept'. From the perspective of prototype theory, for quite some concepts the negation of a concept can be considered as a `singular concept', since it does not have a well defined prototype. In fact, while it is plain to determine the  
non-membership of, e.g., {\it Fruit}, this does not seem to lead to the determination involving a similarity with some prototype of {\it Not Fruit}. Some authors maintain, for this reason, that single negated concepts have little meaning and that conceptual negations can be evaluated only in conjunctions of the form {\it Fruits Which Are Not Vegetables} \cite{h1997}. We agree that there is an asymmetry between the way subjects estimate the membership of an exemplar, e.g., {\it Apple}, with respect to a positive concept, e.g., {\it Fruits}, and the way subjects estimate the membership of the same exemplar with respect to its negative counterpart, e.g., {\it Not Fruits}. Notwithstanding this, we believe it is meaningful to explicitly introduce the concept {\it Not Fruits} in our research. First of all, because we do not confine our concept modeling to prototype theory, on the contrary, our approach is more general, the basic structure of prototype theory can be recovered if we limit the concepts to be in their ground states \cite{ag2005a,ag2005b}. Secondly,  we will see that the quantum modeling elaborated in the present paper sheds light exactly on this problem, namely, the `negated concept' only appears as a full concept in `one half of the representation', while 
is treated as `non-membership with respect to the positive concept' in the other half. Hence, quantum cognition, in the form of the model we work out copes with this problem in a natural way. 

There has been very little research on how human beings interpret and combine negated concepts. In a seminal study, Hampton \cite{h1997} considered in a set of experiments both conjunctions of the form {\it Games Which Are Also Sports} and conjunctions of the form {\it Games Which Are Not Sports}.
His work confirmed overextension in both types of conjunctions, also showing a violation of Boolean classical logical rules for the negation, which has recenly been confirmed by ourselves \cite{s2014b}. These results were the starting point for our research in this paper, whose content can be summarized as follows.

In Section \ref{experiment} we illustrate design (Section \ref{participants}) and procedure (Section \ref{procedure}) of the four cognitive experiments we performed. In the first experiment $e_{AB}$, we tested the membership weights of four different sets of exemplars with respect to four pairs $(A,B)$ of concepts and their conjunction `$A \ {\rm and} \ B$'. In the second experiment $e_{AB'}$, we tested the membership weights of the same four sets of exemplars with respect to the same four pairs $(A,B)$ of concepts, but negating the second concept, hence actually considering $A$, $B'$ and the conjunction `$A \ {\rm and} B'$'. In the third experiment $e_{A'B}$, we tested the membership weights obtained considering $A'$, $B$ and the conjunction `$A' \ {\rm and} \ B$'. Finally, in the fourth experiment $e_{A'B'}$, we considered the membership weights obtained by negating both concepts, hence actually considering $A'$, $B'$ and the conjunction `$A'$ and $B'$'.

We investigate the representability of the collected data, reported in Appendix \ref{tables}, in a `unique classical Kolmogorovian probability space' \cite{k1933}. Basic notions and results on probability measures and classical modeling are briefly reviewed in Appendix \ref{classical}. We prove  theorems providing necessary and sufficient conditions for the modeling of the conceptual conjunctions `$A$ and $B$', `$A$ and $B'$', `$A'$ and $B$', and `$A'$ and $B'$' in such a single classical Kolmogorovian framework in Section \ref{experimentalresults}. Then, we observe in Section \ref{res-experiments} that the data in Appendix \ref{tables} significantly violate our theorems.

One usually identifies four `types of deviations from classicality' when conceptual conjunctions and negations are considered together as specified in our experiments, namely, (1) `conjunction minimum rule deviation', or `overextension', (2) `Kolmogorovian conjunction factor deviation', (3) `double overextension', (4) `conceptual negation deviation'. Our experimental data exhibit deviations from classicality of type (1), (3) and (4), while (2) is not present. Moreover, type-(4) deviations are generally weaker than deviations of type-(1). However, our general analysis of classicality for the presence of conjunction and negation together (Section \ref{experimentalresults}), leads to five classicality conditions to be satisfied of the data to fit into one classical probability setting together. When we analyse the deviations of our data with respect to these five conditions we find a very strong systematic deviation, which is very stable, i.e. gives rise to the same numerical values for the deviation even over the different pairs of concepts that we have experimented on. Analysing this regular pattern of violation, we show how it constitutes a very strong evidence for the 
presence and dominance of what we have called `conceptual emergence' in our earlier work on the Fock space model for the combination of concepts \cite{a2009a,ags2013,abgs2013,s2014a,s2014b}. That the violation is numerically the same independent of the considered pair of concepts indicates that we have identified a non-classical mechanism in human thought which is linked to the depth of concept formation itself, independent of the specific meaning for a specific pair of concepts and a specific set of considered exemplars. This was for ourselves the most surprising and unexpected result of the investigation that we present in the present article, and we also consider it as one of the most important results of this investigation. In our opinion it inescapably proves that, whenever uncertainty is involved, human thought  does not follow the rules of classical probability, and that this deviation of classical probability is strong and takes place on a deep structural conceptual level.  Let us stress in this respect that this overwhelming systematics of deviation of classicality was `not' able to be seen or detected in the foregoing studies of conjunctions of pairs of concepts, since a focus on `overextension' or `Kolmogorovian conjunction factor deviation', which was the focus in all earlier investigations \cite{h1988a,a2009a,ags2013,abgs2013,s2014a,s2014b}, could structurally `not' reveal the systematic deviation we find here now, by lack of symmetry. It was necessary to experiment on conjunction and negation together and derive the five classicality conditions containing the necessary symmetry, to be able to identify this strong and stable pattern of deviation of classicality. The second most important finding of the research we present in this article is related to the same identification of the `deviation of classicality pattern' over all the considered pairs of concepts and their negations. Indeed, like we will show in Section \ref{experimentalresults}, not only do we find a strong and stable numerical deviation independent of the considered concepts and exemplars, additionally, the numerical size of the deviation is `almost' equal to the deviation of our five classicality conditions `if we would substitute the theoretical values for an average quantum model by means of our Fock space model of the situation'. Hence, as a second, equally unexpected and for ourselves surprising result, the data indicate in a very strong way that the deviation is exactly the one that would theoretically be found in case the situation is modelled quantum mechanically. Hence, we believe our finding to be a strong support for the validity of our quantum Fock space model, where this conceptual emergence is described in the first sector of this Fock space.
 
In Section \ref{brussels} we elaborate this quantum-theoretic modeling in Fock space for conceptual negations and conjunctions which naturally extends the modeling in \cite{a2009a} and follows the general lines traced in \cite{s2014a,s2014b}. It is however important to remark that the simultaneous modeling of conjunction and negation requires the introduction of two 
new conceptual steps which were not needed for the modeling of conjunction pairs: (i) the introduction of entangled states in the second sector, which enables formalizing the situation where probabilities in second sector of Fock space can be formed by a `product procedure', even if they are not independent -- this is an aspect of the Fock space model we had not understood in our earlier modeling, hence we could consider it a further new surprising finding of the investigation presented in this paper, i.e. that we can model dependent probabilities keeping the procedure for the conjunction of concepts to be a product procedure in the second sector, and it are the natural possibility for the conjunction to be in an entangled state that allows us to do so; (ii) the handling of `negation' in the second sector by `logical inversion', similarly like we handled conjunction in the second sector by `product', more concretely, an experiment with `negation' with respect to a concept is treated by `negating logically' an experiment on the concept itself. This is also the way in which the Fock space model naturally copes with the general non-prototypicality of a negated concept.

When elaborating our Fock space model for the four pairs, conjunction and negation, we proceed as follows.  We firstly inquire into the representability of the data in only one sector of Fock space (individual Hilbert space) in Section \ref{onesector}. Secondly, we allow the possibility of representing the data in a larger two-sector Fock space in Section \ref{twosector}. In Section \ref{solconditions} we discuss the conditions that should be satisfied by a collection of experimental data to be modeled in this two-sector Fock space. In Section \ref{entanglement} we show that all the conditions in Section \ref{experiment} for a classical representation hold in second sector of Fock space (tensor product Hilbert space) if a suitable entangled state is chosen in this sector. This entails that typical logical rules are satisfied, though in a probabilistic setting or, equivalently, a specific quantum logical structure can be detected, in this sector.

We see in Section \ref{model} that a large amount of data fall outside these classicality conditions and illustrate some specific cases that are classically problematical and that can instead be faithfully represented in our quantum-mechanical framework. The detailed representation is provided in the Supplementary Material attached to the present paper. Finally, in Section \ref{results} we comment on our results, by extensively discussing novelties and confirmations of our quantum-theoretic approach, and we clarify in Section \ref{quantumlogic} how `standard formal quantum logic' may appear in second sector of Fock space, and we explore its connections with classical probabilistic logic in the same sector.

\section{Description of experiments and classicality analysis\label{experiment}}
James Hampton identified in his cognitive tests systematic deviations from classical set (fuzzy set) predictions of conjunctions and disjunctions of two concepts, and named these deviations  `overextensions' and `underextensions' \cite{h1988a,h1988b}. Cases of `double overextension' were also observed. More explicitly, if the membership weight of an exemplar $x$ with respect to the conjunction `$A \ {\rm and} \ B$' of two concepts $A$ and $B$ is higher than the membership weight of $x$ with respect to one concept (both concepts), we say that the membership weight of $x$ is `overextended' (`double overextended') with respect to the conjunction (by abuse of language, we say that $x$ is overextended with respect to the conjunction, in this case). If the membership weight of an exemplar $x$ with respect to the disjunction `$A \ {\rm or} \ B$' of two concepts $A$ and $B$ is less than the membership weight of $x$ with respect to one concept, we say that the membership weight of $x$ is `underextended' with respect to the disjunction (by abuse of language, we say that $x$ is underextended with respect to the disjunction, in this case). 

Similar effects were identified by Hampton in his experiments on conjunction and negation of two concepts \cite{h1997}. The analysis in \cite{a2009a} revealed further deviations from classicality in Hampton's experiments, due to the impossibility to generally represent the collected data in a classical (Kolmogorovian) probability framework. In \cite{s2014b} we moved along this direction and performed an experiment in which we tested both conjunctions of the form `$A$ and $B$' and conjunctions of the form `$A$ and $B'$', for specific pairs $(A,B)$ of natural concepts. We showed that very similar devations from classicality are observed in our experiment on human subjects.

In the present paper we aim to generalize the results in \cite{s2014b}, providing an extensive analysis of conceptual conjunction and negation and investigating their reciprocal influences. To this end we complete the experiment in \cite{s2014b} by performing a more general cognitive test, as described in the following sections.

\subsection{Participants and design\label{participants}}
We asked 40 subjects, chosen among colleagues and friends, to fill in a questionnaire in which they had to estimate the membership of four different sets 
of exemplars with respect to four different pairs $(A,B)$ of natural concepts, and their conjunctions `$A$ and $B$',  `$A$ and $B'$',  `$A'$ and $B$' and `$A'$ and $B'$', where $A'$ and $B'$ denote the negations of the concepts $A$ and $B$, respectively. We devised a `within-subjects design' for our experiments, hence all participants were exposed to every treatment or condition.

We considered four pairs of natural concepts, namely ({\it Home Furnishing}, {\it Furniture}), ({\it Spices}, {\it Herbs}), ({\it Pets}, {\it Farmyard Animals}) and ({\it Fruits}, {\it Vegetables}). For each pair, we considered 24 exemplars and measured their membership with respect to these pairs of concepts and the conjunctions of these pairs mentioned above. The choice of the exemplars and pairs of concepts was inspired by the experiments Hampton conducted on the disjunction of two concepts \cite{h1988b}. We chose four of the eight pairs of concepts tested by Hampton, together with the same set of exemplars, for each pair. This choice was part of a more general research program in which we want to investigate the validity of specific logical rules, such as de Morgan's laws, in concept combinations.

Conceptual membership was estimated by using a `7-point scale'. The tested subjects were asked to choose a number from the set $\{+3, +2, +1, 0, -1, -2, -3 \}$, where the positive numbers $+1$, $+2$ and $+3$ meant that they considered `the exemplar to be a member of the concept' -- $+3$ indicated a strong membership, $+1$ a relatively weak membership. The negative numbers $-1$, $-2$ and $-3$ meant that the subject considered `the exemplar to be a non-member of the concept' -- $-3$ indicated a strong non-membership, $-1$ a relatively weak non-membership.

Although our experiment tests `amount of membership' on the used Likert scale, for the investigation presented in the present paper, we only use the data of a sub experiment, namely the one tested for `membership' or `non-membership' -- our plan is to use the `amount of membership data' for a following study leading to a graphical representation of the data, as we did with Hampton's data for the disjunction in earlier work \cite{ags2013,abgs2013}. A second reason for this specific form of the experiment is, because we wanted to stay as close as possible to the disjunction experiments of James Hampton \cite{h1988b}, since we plan to investigate later the connections of our conjunction data with Hampton's disjunction data, for example to investigate the way in which the `de Morgan Laws' take form in our modeling of the data. So, for these two reasons we performed the full experiment measuring `amount of membership' and also testing simultaneously for `membership or non-membership', while it is only the latter sub experiment that we make use of in the investigation presented in this article. The data of this sub experiment give rise to relative frequencies for testing membership or not membership, which means that in the limit of large numbers in the modeling we get probabilities. More concretely, $\mu(A \ {\rm and} \ B)$ is the large number limit of the relative frequency for $x$ to be a member of `$A \ {\rm and} \ B$' in the performed experiment. We get to this by converting the values collected on the 7-point scale by associating a value $+1$ to each positive value on the 7-point scale, $-1$ to each negative number, and $0.5$ to each $0$ on the same 7-point scale.

\subsection{Procedure and materials\label{procedure}}
For each pair $(A,B)$ of natural concepts, the 40 participants were involved in four subsequent experiments, $e_{AB}$, $e_{AB'}$, $e_{A'B}$ and $e_{A'B'}$, corresponding to the conjunctions `$A$ and $B$', `$A$ and $B'$', `$A'$ and $B$' and `$A'$ and $B'$', respectively. More specifically, the four sequential experiments can be illustrated as follows.

For the conceptual pair ({\it Home Furnishing}, {\it Furniture}), we firstly asked the 40 subjects to estimate the membership of the first set of 24 exemplars with respect to the concepts {\it Home Furnishing}, {\it Furniture}, and their conjunction {\it Home Furnishing And Furniture}. Then, we asked the same 40 subjects to estimate the membership of the same set of 24 exemplars with respect to the concept  {\it Home Furnishing}, the negation {\it Not Furniture} of the concept {\it Furniture}, and their conjunction {\it Home Furnishing And Not Furniture}. Subsequently, we asked the 40 subjects to estimate the membership of the 24 exemplars with respect to the negation {\it Not Home Furnishing} of the concept  {\it Home Furnishing}, the concept {\it Furniture}, and their conjunction {\it Not Home Furnishing And Furniture}. Finally,  we asked the 40 subjects to estimate the membership of the 24 exemplars with respect to the negations {\it Not Home Furnishing},  {\it Not Furniture}, and their conjunction {\it Not Home Furnishing And Not Furniture}. The corresponding membership weights are reported in Table 1, Appendix  \ref{tables}.

For the conceptual pair ({\it Spices}, {\it Herbs}), we firstly asked the 40 subjects to estimate the membership of the second set of 24 exemplars with respect to the concepts {\it Spices}, {\it Herbs}, and their conjunction {\it Spices And Herbs}. Then, we asked the same 40 subjects to estimate the membership of the same set of 24 exemplars with respect to the concept  {\it Spices}, the negation {\it Not Herbs} of the concept {\it Herbs}, and their conjunction {\it Spices And Not Herbs}. Subsequently, we asked the 40 subjects to estimate the membership of the 24 exemplars with respect to the negation {\it Not Spices} of the concept  {\it Spices}, the concept {\it Herbs}, and their conjunction {\it Not Spices And Herbs}. Finally,  we asked the 40 subjects to estimate the membership of the 24 exemplars with respect to the negations {\it Not Spices},  {\it Not Herbs}, and their conjunction {\it Not Spices And Not Herbs}. The corresponding membership weights are reported in Table 2, Appendix \ref{tables}.

For the conceptual pair ({\it Pets}, {\it Farmyard Animals}), we firstly asked the 40 subjects to estimate the membership of the third set of 24 exemplars with respect to the concepts {\it Pets}, {\it Farmyard Animals}, and their conjunction {\it Pets And Farmyard Animals}. Then, we asked the same 40 subjects to estimate the membership of the same set of 24 exemplars with respect to the concept  {\it Pets}, the negation {\it Not Farmyard Animals} of the concept {\it Farmyard Animals}, and their conjunction {\it Pets And Not Farmyard Animals}. Subsequently, we asked the 40 subjects to estimate the membership of the 24 exemplars with respect to the negation {\it Not Pets} of the concept  {\it Pets}, the concept {\it Farmyard Animals}, and their conjunction {\it Not Pets And Farmyard Animals}. Finally,  we asked the 40 subjects to estimate the membership of the 24 exemplars with respect to the negations {\it Not Pets},  {\it Not Farmyard Animals}, and their conjunction {\it Not Pets And Not Farmyard Animals}. The corresponding membership weights are reported in Table 3, Appendix \ref{tables}.

For the conceptual pair ({\it Fruits}, {\it Vegetables}), we firstly asked the 40 subjects to estimate the membership of the third set of 24 exemplars with respect to the concepts {\it Fruits}, {\it Vegetables}, and their conjunction {\it Fruits And Vegetables}. Then, we asked the same 40 subjects to estimate the membership of the same set of 24 exemplars with respect to the concept  {\it Fruits}, the negation {\it Not Vegetables} of the concept {\it Vegetables}, and their conjunction {\it Fruits And Not Vegetables}. Subsequently, we asked the 40 subjects to estimate the membership of the 24 exemplars with respect to the negation {\it Not Fruits} of the concept  {\it Fruits}, the concept {\it Vegetables}, and their conjunction {\it Not Fruits And Vegetables}. Finally,  we asked the 40 subjects to estimate the membership of the 24 exemplars with respect to the negations {\it Not Fruits},  {\it Not Vegetables}, and their conjunction {\it Not Fruits And Not Vegetables}. The corresponding membership weights are reported in Table 4, Appendix \ref{tables}.

\subsection{Methodology\label{experimentalresults}}
A first inspection of tables Tables 1--4 in Appendix \ref{tables} already reveals that some exemplars present overextension with respect to all conjunctions `$A  \ {\rm and} \ B$', `$A \ {\rm and} \ B'$', `$A' \ {\rm and} \ B$', `$A' \ {\rm and} \ B'$'. This is the case, e.g., for the exemplar
{\it Lamp} with respect to the concepts {\it Home Furnishing} and {\it Furniture} (Table 1), the exemplar {\it Salt} with respect to {\it Spices} and {\it Herbs} (Table 2), the exemplar {\it Goldfish} with respect to {\it Pets} and {\it Farmyard Animals} (Table 3) and the exemplar {\it Mustard} with respect to {\it Fruits} and {\it Vegetables} (Table 4). Hence, manifest deviations from classicality occurred in our experiments. When we say `deviations from classicality', we actually mean that the collected data behave in such a way that they cannot generally be modeled by using the usual connectives of classical (fuzzy set) logic for conceptual conjunctions, neither the rules of classical probability for their membership weights. In order to systematically identify such deviations from classicality we need however a characterization of the representability of these data in a classical probability space. To this end 
we derive in the following step by step conditions that will give us an overall picture of the classicality of conceptual conjunctions and negations. Finally, we arrive to a set of five conditions, formulated in Theorem 3 as a set of necessary and sufficient conditions of classicality for a pair of concepts, its negations and conjunctions to be representable within a classical Kolmogorovian probability model. Symbols and notions are introduced in Appendix \ref{classical}. Let us mention that to our knowledge the `necessary and sufficient conditions for probabilities $\mu(A), \mu(B), \mu(A'), \mu(B')$, $\mu(A\ {\rm and}\ B)$, $\mu(A\ {\rm and}\ B')$, $\mu(A'\ {\rm and}\ B)$ and $\mu(A'\ {\rm and}\ B')$ to be represented within in classical Kolmogorovian probability model, have not yet been systematically derived, and hence are not known in the literature. However the `necessary and sufficient conditions for probabilities $\mu(A), \mu(B)$ and $\mu(A\ {\rm and}\ B)$ to be represented within in classical Kolmogorovian probability model have been systematically studied \cite{p1989}, their direct derivation can for example be found in \cite{a2009a}, theorem 1 of section 1.3. We will start our investigation of the classicality condition by making use of the conditions that could be derived for $\mu(A), \mu(B)$ and $\mu(A\ {\rm and}\ B)$ and applying them additionally to $\mu(A'), \mu(B')$ and $\mu(A'\ {\rm and}\ B')$, and to add some intermediate conditions connecting $\mu(A), \mu(B)$ and $\mu(A\ {\rm and}\ B)$ and $\mu(A'), \mu(B')$ and $\mu(A'\ {\rm and}\ B')$, to also imply classicality for the mixed situations such as $\mu(A), \mu(B')$ and $\mu(A\ {\rm and}\ B')$. We can prove the following theorem.

\bigskip
\noindent
{\bf Theorem 1.} {\it The membership weights $\mu(A), \mu(B), \mu(A'), \mu(B')$, $\mu(A\ {\rm and}\ B)$, $\mu(A\ {\rm and}\ B')$, $\mu(A'\ {\rm and}\ B)$ and $\mu(A'\ {\rm and}\ B')$ of an exemplar $x$ with respect to the concepts $A$, $B$, the negations `not $A$', `not $B$', the conjunctions `$A$ and $B$', `$A$ and $B'$', `$A'$ and $B$' and `$A'$ and $B'$' are classical conjunction data
i.e. can be represented in a classical Kolmogorovian probability model, if and only if they satisfy the following conditions.}
\begin{eqnarray} \label{ineq01}
&0 \le \mu(A\ {\rm and}\ B) \le \mu(A) \le 1 \\ \label{ineq02}
&0 \le \mu(A\ {\rm and}\ B) \le \mu(B) \le 1 \\ \label{ineq03}
&0 \le \mu(A'\ {\rm and}\ B') \le \mu(A') \le 1 \\ \label{ineq04}
&0 \le \mu(A'\ {\rm and}\ B') \le \mu(B') \le 1 \\ \label{ineq05}
&\mu(A) - \mu(A\ {\rm and}\ B)=\mu(B')-\mu(A'\ {\rm and}\ B')=\mu(A \ {\rm and}\ B') \\ \label{ineq06}
&\mu(B) - \mu(A\ {\rm and}\ B)=\mu(A')-\mu(A'\ {\rm and}\ B')=\mu(A' \ {\rm and}\ B) \\ \label{ineq07}
&1-\mu(A)-\mu(B)+\mu(A\ {\rm and}\ B)=\mu(A'\ {\rm and}\ B') \\ 
&1-\mu(A')-\mu(B')+\mu(A'\ {\rm and}\ B')=\mu(A\ {\rm and}\ B)\label{ineq08}
\end{eqnarray}
 
\medskip
\noindent
{\bf Proof.} If $\mu(A), \mu(B), \mu(A'), \mu(B')$ and $\mu(A\ {\rm and}\ B), \mu(A\ {\rm and}\ B'), \mu(A'\ {\rm and}\ B), \mu(A'\ {\rm and}\ B')$ are classical conjunction and negation data, then there exists a Kolmogorovian probability space $(\Omega,\sigma(\Omega),P)$ and events $E_A, E_B \in \sigma(\Omega)$ such that $P(E_A) = \mu(A)$, $P(E_B) = \mu(B)$, $P(\Omega \setminus E_A) = \mu(A')$, $P(\Omega \setminus E_B) = \mu(B')$, $P(E_A \cap E_B) = \mu(A\ {\rm and}\ B)$, $P(E_A \cap((\Omega \setminus E_B)) = \mu(A\ {\rm and}\ B')$, $P((\Omega \setminus E_A) \cap E_B) = \mu(A'\ {\rm and}\ B)$ and $P((\Omega \setminus E_A) \cap (\Omega \setminus E_B)) = \mu(A'\ {\rm and}\ B')$. From the general properties of a Kolmogorovian probability space it follows that (\ref{ineq01}), (\ref{ineq02}), (\ref{ineq03}), (\ref{ineq04}), (\ref{ineq05}), (\ref{ineq06}), (\ref{ineq07}) and (\ref{ineq08}) are satisfied.

Now suppose that $x$ is such that its membership weights $\mu(A), \mu(B), \mu(A'), \mu(B')$, $\mu(A\ {\rm and}\ B)$, $\mu(A\ {\rm and}\ B')$, $\mu(A'\ {\rm and}\ B)$ and $\mu(A'\ {\rm and}\ B')$ with respect to the concepts $A$, $B$, $A'$, $B'$, `$A$ and $B$', `$A$ and $B'$', `$A'$ and $B$' and `$A'$ and $B'$', respectively, satisfy Equations (\ref{ineq01}), (\ref{ineq02}), (\ref{ineq03}), (\ref{ineq04}), (\ref{ineq05}), (\ref{ineq06}), (\ref{ineq07}) and (\ref{ineq08}). We will prove that as a consequence $\mu(A), \mu(B), \mu(A'), \mu(B')$, $\mu(A\ {\rm and}\ B)$, $\mu(A\ {\rm and}\ B')$, $\mu(A'\ {\rm and}\ B)$ and $\mu(A'\ {\rm and}\ B')$ are classical conjunction and negation data, in the sense that `there exists a classical Kolmogorovian probability space, such that we can represent all of them as measures on event sets of this space'. We make our proof by explicitly construct a Kolmogorovian probability space that models these data. Consider the set $\Omega=\{1, 2, 3, 4\}$ and $\sigma(\Omega) = {\cal P}(\Omega)$, the set of all subsets of $\Omega$. We define
\begin{eqnarray} \label{eqtheorem101}
&P(\{1\}) = \mu(A\ {\rm and}\ B) \\ \label{eqtheorem102}
&P(\{2\}) = \mu(A\ {\rm and}\ B')=\mu(A) - \mu(A\ {\rm and}\ B) \\ \label{eqtheorem103}
&P(\{3\}) = \mu(A'\ {\rm and}\ B)=\mu(A') - \mu(A'\ {\rm and}\ B') \\ \label{eqtheorem104}
&P(\{4\}) = \mu(A'\ {\rm and}\ B')
\end{eqnarray}
and further for an arbitrary subset $S \subseteq \{1,2,3,4\}$ we define
\begin{equation} \label{defarbitrarysubset}
P(S) = \sum_{a\in S}P(\{a\})
\end{equation}
Let us prove that $P: \sigma(\Omega) \rightarrow [0,1]$ is a probability measure. For this purpose, we need to prove that $P(S) \in [0,1]$ for an arbitrary subset $S \subseteq \Omega$, and that the `sum formula' for a probability measure is satisfied. The sum formula for a probability measure is satisfied because of definition (\ref{defarbitrarysubset}). What remains to be proved is that $P(S) \in [0,1]$ for an arbitrary subset $S \subseteq \Omega$, and that all different subsets that can be formed are contained in $\sigma(\Omega)$. $P(\{1\}), P(\{2\}), P(\{3\})$ and $P(\{4\})$ are contained in $[0,1]$ as a consequence of equations (\ref{ineq01}), (\ref{ineq03}), (\ref{ineq05}) and (\ref{ineq06}). Using (\ref{ineq05}) 
we have that $P(\{1,2\})=\mu(A\ {\rm and}\ B)+\mu(A\ {\rm and}\ B')=\mu(A\ {\rm and}\ B)+\mu(A) - \mu(A\ {\rm and}\ B)=\mu(A)$. Using (\ref{ineq06})
we have that $P(\{3,4\})=\mu(A'\ {\rm and}\ B')+\mu(A'\ {\rm and}\ B)=\mu(A'\ {\rm and}\ B')+\mu(A')-\mu(A'\ {\rm and}\ B')=\mu(A')$. Again using (\ref{ineq06}) we have 
that $P(\{1,3\})=\mu(A\ {\rm and}\ B)+\mu(A'\ {\rm and}\ B)=\mu(A\ {\rm and}\ B)+\mu(B) - \mu(A\ {\rm and}\ B)=\mu(B)$, and using again (\ref{ineq05}) we have 
that $P(\{2,4\})=\mu(A\ {\rm and}\ B')+\mu(A'\ {\rm and}\ B')=\mu(B')-\mu(A'\ {\rm and}\ B')+\mu(A'\ {\rm and}\ B')=\mu(B')$. Moreover, $P(\{1,2\}), P(\{3,4\}), P(\{1,3\})$ and $P(\{2,4\})$ are all contained in $[0,1]$ as a consequence of equations (\ref{ineq01}), (\ref{ineq02}), (\ref{ineq03}) and (\ref{ineq04}). We 
have already found the representatives of all elements and their conjunctions in $\sigma(\Omega)$. But we have not yet considered all subsets of $\Omega$. Indeed, let us consider $\mu(\{1,2,3\})=\mu(A\ {\rm and}\ B)+\mu(A\ {\rm and}\ B')+\mu(A'\ {\rm and}\ B)=1-\mu(A'\ {\rm and}\ B')$. And from (\ref{ineq07}) it follows that this is contained in $[0.1]$. In an analogous way we prove that $\mu(\{1,2,4\})=1-\mu(A'\ {\rm and}\ B)$, $\mu(\{1,3,4\})=1-\mu(A\ {\rm and}\ B')$, and $\mu(\{2,3,4\})=1-\mu(A\ {\rm and}\ B)$. We almost have all 
subsets of 
$\Omega$. Remains 
$\{1,4\}$ and  
$\{2,3\}$. Since by construction we have $\mu(\{1\}) \le \mu(\{1,4\}) \le  \mu(\{1,2,4\})$ and $\mu(\{2\}) \le \mu(\{2,3\}) \le \mu(\{2,3,4\})$, it follows that both 
$\mu(\{1,4\})$ and $\mu(\{2,3\})$ are contained in $[0,1]$. The last subset to control is $\Omega$ itself. We have $P(\Omega)=P(\{1\}) + P(\{2\}) + P(\{3\}) + P(\{4\}) = 1$, following the calculation we made above. We have verified all subsets $S \subseteq \Omega$, and hence proved that $P$ is a probability measure. All subsets for which we have gathered data are represented in this $\sigma$-algebra, which completes our proof.
\qed

\bigskip
\noindent
It is important to stress that we did not use (\ref{ineq08}) in the proof of Theorem 1, which means that inequalities (\ref{ineq05}), (\ref{ineq06}), (\ref{ineq07}) and (\ref{ineq08}) are not independent. By using, for example, (\ref{ineq05}), (\ref{ineq06}), and then (\ref{ineq07}) we get
\begin{eqnarray}
&&1-\mu(A')-\mu(B')+\mu(A'\ {\rm and}\ B')=1-\mu(B)-\mu(B')+\mu(A\ {\rm and}\ B) \nonumber \\
&=&1-\mu(B)-\mu(A'\ {\rm and}\ B')-\mu(A)+\mu(A\ {\rm and}\ B)+\mu(A\ {\rm and}\ B) \nonumber \\
&=&\mu(A'\ {\rm and}\ B')-\mu(A'\ {\rm and}\ B')+\mu(A\ {\rm and}\ B) \nonumber \\
&=&\mu(A\ {\rm and}\ B) \nonumber
\end{eqnarray}
which proves indeed that (\ref{ineq08}) can be derived from (\ref{ineq05}), (\ref{ineq06}) and then (\ref{ineq07}), and can be left out as a condition.

Let us now prove a result which is useful for our purposes. Following (\ref{ineq05}) and (\ref{ineq06}) we have that $\mu(A \ {\rm and}\ B')+\mu(A\ {\rm and}\ B)+\mu(A' \ {\rm and}\ B)=\mu(A)+\mu(B) - \mu(A\ {\rm and}\ B)$. Moreover, by using (\ref{ineq07}), we get $\mu(A \ {\rm and}\ B')+\mu(A\ {\rm and}\ B)+\mu(A' \ {\rm and}\ B)+\mu(A'\ {\rm and}\ B')=\mu(A)+\mu(B) - \mu(A\ {\rm and}\ B)+1-\mu(A)-\mu(B)+\mu(A\ {\rm and}\ B)=1$. 

The equality
\begin{eqnarray} \label{sumunity}
\mu(A \ {\rm and}\ B')+\mu(A\ {\rm and}\ B)+\mu(A' \ {\rm and}\ B)+\mu(A'\ {\rm and}\ B')=1
\end{eqnarray}
can be used, together with (\ref{ineq05}) and (\ref{ineq06}), as follows.
\begin{eqnarray}
\mu(A)+\mu(A')=\mu(A \ {\rm and}\ B)+\mu(A \ {\rm and}\ B')+\mu(A' \ {\rm and}\ B)+\mu(A' \ {\rm and}\ B')=1 \nonumber \\
\mu(B)+\mu(B')=\mu(A \ {\rm and}\ B)+\mu(A' \ {\rm and}\ B)+\mu(A \ {\rm and}\ B')+\mu(A' \ {\rm and}\ B')=1 \nonumber
\end{eqnarray}
This means that from (\ref{ineq01}), and hence $0 \le \mu(A) \le 1$, follows that $0 \le 1-\mu(A) \le 1$, and hence $0 \le \mu(A') \le 1$. And from (\ref{ineq02}), and hence $0 \le \mu(B) \le 1$, follows that $0 \le 1-\mu(B) \le 1$, and hence $0 \le \mu(B') \le 1$. Suppose now that (\ref{ineq01}) and (\ref{ineq02}) are satisfied. This means that $0 \le \mu(A)-\mu(A \ {\rm and}\ B)=\mu(B')-\mu(A' \ {\rm and}\ B')$ and hence $\mu(A' \ {\rm and}\ B') \le \mu(B')$, and $0 \le \mu(B)-\mu(A \ {\rm and}\ B)=\mu(A')-\mu(A' \ {\rm and}\ B')$ and hence $\mu(A' \ {\rm and}\ B') \le \mu(A')$. The only condition that lacks to have derived (\ref{ineq03}) and (\ref{ineq04}) from (\ref{ineq01}) and (\ref{ineq02}), is that $0 \le \mu(A' \ {\rm and}\ B')$. We can add this as a requirement to (\ref{ineq07}). 

Hence, we have proved the following theorem. 

\bigskip
\noindent 
{\bf Theorem 2.} {\it The membership weights $\mu(A), \mu(B), \mu(A'), \mu(B')$, $\mu(A\ {\rm and}\ B)$, $\mu(A\ {\rm and}\ B')$, $\mu(A'\ {\rm and}\ B)$ and $\mu(A'\ {\rm and}\ B')$ of an exemplar $x$ with respect to the concepts $A$, $B$, $A'$ and $B'$ and the conjunctions `$A$ and $B$', `$A$ and $B'$', `$A'$ and $B$' and `$A'$ and $B'$' are classical conjunction data if and only if they satisfy the following conditions.}
\begin{eqnarray} \label{bis-ineq01}
&0 \le \mu(A\ {\rm and}\ B) \le \mu(A) \le 1 \\ \label{bis-ineq02}
&0 \le \mu(A\ {\rm and}\ B) \le \mu(B) \le 1 \\ \label{bis-ineq03}
&\mu(A) - \mu(A\ {\rm and}\ B)=\mu(B')-\mu(A'\ {\rm and}\ B')=\mu(A \ {\rm and}\ B') \\ \label{bis-ineq06}
&\mu(B) - \mu(A\ {\rm and}\ B)=\mu(A')-\mu(A'\ {\rm and}\ B')=\mu(A' \ {\rm and}\ B) \\ \label{bis-ineq07}
&0 \le 1-\mu(A)-\mu(B)+\mu(A\ {\rm and}\ B)=\mu(A'\ {\rm and}\ B')
\end{eqnarray}

\bigskip
\noindent
The classicality requirements in Theorems 1 and 2 are not symmetric with respect to the exchange of $A$ with $A'$ and $B$ with $B'$. Thus, we can look for equivalent and more symmetric sets of requirements. These include validity of the `marginal law' of classical probability.

\bigskip
\noindent
{\bf Lemma 1.} The four equalities defined in (\ref{ineq05}) and (\ref{ineq06}) are equivalent with the following four equalities expressing the marginal law for all elements to be satisfied.
\begin{eqnarray} \label{marg01}
\mu(A)&=&\mu(A\ {\rm and}\ B)+\mu(A \ {\rm and}\ B') \\ \label{marg02}
\mu(B)&=&\mu(A\ {\rm and}\ B)+\mu(A' \ {\rm and}\ B) \\ \label{marg03}
\mu(A')&=&\mu(A'\ {\rm and}\ B')+\mu(A' \ {\rm and}\ B) \\ \label{marg04}
\mu(B')&=&\mu(A'\ {\rm and}\ B')+\mu(A \ {\rm and}\ B')
\end{eqnarray}

\medskip
\noindent
{\bf Proof.} That (\ref{marg01}), (\ref{marg02}), (\ref{marg03}) and (\ref{marg04}) follow from (\ref{ineq05}) and (\ref{ineq06}) follows from a simple reshuffling of the terms.
Suppose now that (\ref{marg01}), (\ref{marg02}), (\ref{marg03}) and (\ref{marg04}) are satisfied. The inverse reshuffling of the same terms proves that (\ref{ineq05}) and (\ref{ineq06}) are satisfied.
\qed

\bigskip
\noindent
Lemma 1 entails that we can substitute (\ref{ineq05}) and (\ref{ineq06}) by the four equations expressing the marginal law to be satisfied. 

A further simplification is possible. Indeed, (\ref{ineq05}), (\ref{ineq06}) and (\ref{ineq07}) are equivalent with (\ref{marg01}), (\ref{marg02}), (\ref{marg03}), (\ref{marg04}) and (\ref{sumunity}). We have proved above that (\ref{ineq05}), (\ref{ineq06}) and (\ref{ineq07}) imply (\ref{sumunity}). Let us prove the inverse. Hence suppose that (\ref{marg01}), (\ref{marg02}), (\ref{marg03}), (\ref{marg04}) and (\ref{sumunity}) are satisfied, and let us proof (\ref{ineq07}). We have
\begin{eqnarray}
&&1-\mu(A)-\mu(B)+\mu(A\ {\rm and}\ B) \nonumber \\
&=&1-\mu(A\ {\rm and}\ B)-\mu(A \ {\rm and}\ B')-\mu(A\ {\rm and}\ B)-\mu(A' \ {\rm and}\ B)+\mu(A\ {\rm and}\ B) \nonumber \\
&=&1-\mu(A\ {\rm and}\ B)-\mu(A \ {\rm and}\ B')-\mu(A' \ {\rm and}\ B) \nonumber \\
&=&\mu(A'\ {\rm and}\ B') \nonumber
\end{eqnarray}
which proves that Equation (\ref{ineq07}) holds.

We have thus proved the following theorem, stating a new and more symmetric set of classicality conditions.

\bigskip
\noindent 
{\bf Theorem 3.} {\it The membership weights $\mu(A), \mu(B), \mu(A'), \mu(B')$, $\mu(A\ {\rm and}\ B)$, $\mu(A\ {\rm and}\ B')$, $\mu(A'\ {\rm and}\ B)$ and $\mu(A'\ {\rm and}\ B')$ of an exemplar $x$ with respect to the concepts $A$, $B$, $A'$ and $B'$ and the conjunctions `$A$ and $B$', `$A$ and $B'$', `$A'$ and $B$' and `$A'$ and $B'$' are classical conjunction data if and only if they satisfy the following conditions.}
\begin{eqnarray} \label{cond01}
&0 \le \mu(A\ {\rm and}\ B) \le \mu(A) \le 1 \\ \label{cond02}
&0 \le \mu(A\ {\rm and}\ B) \le \mu(B) \le 1 \\ \label{cond03}
&\mu(A)=\mu(A\ {\rm and}\ B)+\mu(A \ {\rm and}\ B') \\ \label{cond04}
&\mu(B)=\mu(A\ {\rm and}\ B)+\mu(A' \ {\rm and}\ B) \\ \label{cond05}
&\mu(A')=\mu(A'\ {\rm and}\ B')+\mu(A' \ {\rm and}\ B) \\ \label{cond06}
&\mu(B')=\mu(A'\ {\rm and}\ B')+\mu(A \ {\rm and}\ B') \\ \label{cond07}
&0 \le 1-\mu(A\ {\rm and}\ B)-\mu(A\ {\rm and}\ B')-\mu(A'\ {\rm and}\ B)=\mu(A'\ {\rm and}\ B')
\end{eqnarray}

\bigskip
\noindent
The conditions above can be further simplified by observing that the membership weights we collected in our experiments are large number limits of relative frequencies, thus all measured quantities are already contained in the interval $[0, 1]$. Therefore, we have
\begin{eqnarray} \label{interval01}
\mu(A), \mu(B), \mu(A'), \mu(B'), \mu(A\ {\rm and}\ B), \mu(A\ {\rm and}\ B') , \mu(A'\ {\rm and}\ B) , \mu(A'\ {\rm and}\ B') \in [0,1] 
\end{eqnarray}
Now, when (\ref{interval01}) is satisfied, we have that from (\ref{cond04}) and (\ref{cond05}) follows that
\begin{eqnarray}
\mu(A\ {\rm and}\ B) \le \mu(A) \nonumber \\
\mu(A\ {\rm and}\ B) \le \mu(B) \nonumber 
\end{eqnarray}
This entails that (\ref{cond01}) and (\ref{cond02}) are satisfied, when (\ref{cond04}) and (\ref{cond05}) are. Hence, we can amazingly enough formulate Theorem 3 a new, with only five conditions to be satisfied -- four conditions expressing the marginal law.

\bigskip
\noindent 
{\bf Theorem 3$'$.} {\it If the membership weights $\mu(A), \mu(B), \mu(A'), \mu(B')$, $\mu(A\ {\rm and}\ B)$, $\mu(A\ {\rm and}\ B')$, $\mu(A'\ {\rm and}\ B)$ and $\mu(A'\ {\rm and}\ B')$ of an exemplar $x$ with respect to the concepts $A$, $B$, $A'$ and $B'$ and the conjunctions `$A$ and $B$', `$A$ and $B'$', `$A'$ and $B$' and `$A'$ and $B'$' are all contained in the interval $[0,1]$, they are classical conjunction data if and only if they satisfy the following conditions.}
\begin{eqnarray} \label{condbis01}
&\mu(A)=\mu(A\ {\rm and}\ B)+\mu(A \ {\rm and}\ B') \\ \label{condbis02}
&\mu(B)=\mu(A\ {\rm and}\ B)+\mu(A' \ {\rm and}\ B) \\ \label{condbis03}
&\mu(A')=\mu(A'\ {\rm and}\ B')+\mu(A' \ {\rm and}\ B) \\ \label{condbis04}
&\mu(B')=\mu(A'\ {\rm and}\ B')+\mu(A \ {\rm and}\ B') \\ \label{condbis05}
&\mu(A\ {\rm and}\ B)+\mu(A\ {\rm and}\ B')+\mu(A'\ {\rm and}\ B)+\mu(A'\ {\rm and}\ B')=1
\end{eqnarray}

\bigskip
\noindent
Equations (\ref{condbis01})--(\ref{condbis05}) express classicality conditions in their most symmetric form. A more traditional way to quantify deviations from classical conjunction in real data is resorting to the following parameters.
\begin{eqnarray}
&\Delta_{AB}=\mu(A\ {\rm and}\ B)-\min \{ \mu(A),\mu(B) \} \\
&\Delta_{AB'}=\mu(A\ {\rm and}\ B')-\min \{ \mu(A),\mu(B') \} \\
&\Delta_{A'B}=\mu(A'\ {\rm and}\ B)-\min \{\mu(A'),\mu(B) \} \\
&\Delta_{A'B'}=\mu(A'\ {\rm and}\ B')-\min\{ \mu(A'),\mu(B') \} \\
\end{eqnarray}
In fact, the quantities $\Delta_{AB}$, $\Delta_{AB'}$, $\Delta_{A'B}$ and $\Delta_{A'B'}$ typically measure overextension with respect to the conjunctions `$A$ and $B$', `$A$ and $B'$', `$A'$ and $B$' and `$A'$ and $B'$', respectively \cite{h1988a}. 
However, overextension-type deviations are generally not the only way in which membership for conjunction of concepts can deviate from classicality. Let us now introduce the following quantities: 
\begin{eqnarray}
&k_{AB}=1-\mu(A)-\mu(B)+\mu(A\ {\rm and}\ B) \\
&k_{AB'}=1-\mu(A)-\mu(B')+\mu(A\ {\rm and}\ B') \\
&k_{A'B}=1-\mu(A')-\mu(B)+\mu(A'\ {\rm and}\ B) \\
&k_{A'B'}=1-\mu(A')-\mu(B')+\mu(A'\ {\rm and}\ B')
\end{eqnarray}
The quantities $k_{AB}$, $k_{AB'}$, $k_{A'B}$ and $k_{A'B'}$ have been named `Kolmogorovian conjunction factors' and studied in detail in  \cite{a2009a}. The Kolmogorovian factors measure a deviation that can be understood as of `opposite type' than the deviation measured by the overextension. Namely, the condition for $k_{AB}$ is violated when both $\mu(A)$ and $\mu(B)$ are `too large' compared with $\mu(A ~{\rm and}~B)$. Finally, we introduce a new type of quantities that measure 
the deviations of classicality as expressed by (\ref{condbis02})--(\ref{condbis05}), hence essentially deviations from the marginal law of classical probability:\footnote{Remark that, if we set $I_{AA'}=1-\mu(A)-\mu(A')$ and $I_{BB'}=1-\mu(B)-\mu(B')$, we have
$I_{AA'}=I_{ABA'B'}-I_A-I_{A'}$ and $I_{BB'}=I_{ABA'B'}-I_B-I_{B'}$, which means that the parameters  $I_{AA'}$ and $I_{BB'}$ used in  \cite{s2014b} can be derived from the parameters $I_{ABA'B'}$, $I_{A}$, $I_{B}$, $I_{A'}$ and $I_{B'}$.}
\begin{eqnarray}
&I_{ABA'B'}=1-\mu(A\ {\rm and}\ B)-\mu(A\ {\rm and}\ B')-\mu(A'\ {\rm and}\ B)-\mu(A'\ {\rm and}\ B') \label{normalization} \\
&I_{A}=\mu(A)-\mu(A\ {\rm and}\ B)-\mu(A \ {\rm and}\ B') \label{negationA} \\
&I_{B}=\mu(B)-\mu(A\ {\rm and}\ B)-\mu(A' \ {\rm and}\ B) \label{negationB}\\
&I_{A'}=\mu(A')-\mu(A'\ {\rm and}\ B')-\mu(A' \ {\rm and}\ B) \label{negationA'}\\
&I_{B'}=\mu(B')-\mu(A'\ {\rm and}\ B')-\mu(A \ {\rm and}\ B') \label{negationB'}
\end{eqnarray}

It is finally interesting to note that Theorem 3$'$ can be reformulated by means of the introduced parameters as follows.

\bigskip
\noindent {\bf Theorem 3$''$.} {\it If the membership weights $\mu(A), \mu(B), \mu(A'), \mu(B')$, $\mu(A\ {\rm and}\ B)$, $\mu(A\ {\rm and}\ B')$, $\mu(A'\ {\rm and}\ B)$ and $\mu(A'\ {\rm and}\ B')$ of an exemplar $x$ with respect to concepts $A$, $B$, $A'$ and $B'$ and the conjunctions `$A$ and $B$', `$A$ and $B'$', `$A'$ and $B$' and `$A'$ and $B'$', are all contained in the interval $[0,1]$, they are classical conjunction data if and only if}
\begin{eqnarray} 
I_{ABA'B'}=I_{A}=I_{B}=I_{A'}=I_{B'}=0
\end{eqnarray}

\subsection{Results \label{res-experiments}}
Let us now come back to our experiments. One observes by pure inspection that Theorems 1--3 are manisfestly violated in several cases. To see this systematically we report in Appendix \ref{tables} the relevant conditions that should hold in a classical setting. Since the conditions $k_{AB}>0$, $k_{AB'}>0$, $k_{A'B}>0$ and $k_{A'B'}>0$ are always satisfied, they are not explicitly inserted in Tables 1--4. On the contrary, $\Delta_{XY}$, $I_{X}$, $I_{Y}$, $X=A,A',Y=B,B'$, and $I_{ABA'B'}$ are frequently violated.  This means that deviations from a classical probability model in our experimental data are due to both overextension in the conjunctions and violations of classicality in the negations. We consider some relevant cases in the following.

The exemplar {\it Apple} scores $\mu(A)=1$ with respect to the concept {\it Fruits}, $\mu(B)=0.23$ with respect to the concept {\it Vegetables}, and $\mu(A \ {\rm and} \ B)=0.6$ with respect to the conjunction {\it Fruits And Vegetables}, hence it has $\Delta_{AB}=0.38$ (Table 4). The exemplar {\it Prize Bull} scores $\mu(A)=0.13$ with respect to {\it Pets}, $\mu(B)=0.76$ with respect to the concept {\it Farmyard Animals}, and $\mu(A \ {\rm and} \ B)=0.43$ with respect to the conjunction {\it Pets And Farmyard Animals}, hence it has $\Delta_{AB}=0.29$ (Table 3). The membership weight of {\it Chili Pepper} with respect to {\it Spices} is 0.98, with respect to {\it Herbs} is	0.53, while its membership weight with respect to the conjunction {\it Spices And Herbs} is 0.8, hence $\Delta_{AB}=0.27$, thus giving rise to overextension (Table 2). Even stronger deviations are observed in the combination {\it Fruits And Vegetables}. For example, the exemplar {\it Broccoli} scores 0.09 with respect to {\it Fruits}, 1 with respect to {\it Vegetables}, and 0.59 with respect to {\it Fruits And Vegetables} ($\Delta_{AB}=0.49$). A similar pattern is observed for {\it Parsley}, which scores 0.02 with respect to {\it Fruits},	0.78 with respect to {\it Vegetables} and 0.45 with respect to {\it Fruits And Vegetables} ($\Delta_{AB}=0.43$, Table 4).

Overextension is present when one concept is negated. More explicitly:

(i) in the conjunction `$A$ and $B'$'. Indeed, the membership weights of {\it Shelves} with respect to {\it Home Furnishing}, {\it Not Furniture} and {\it Home Furnishing And Not Furniture} is 0.85, 0.13 and 0.39, respectively, for a $\Delta_{AB'}=0.26$ (Table 1). Then,  {\it Pepper} scores 0.99 with respect to {\it Spices},  0.58 with respect to {\it Not Herbs}, and	0.9 with respect to {\it Spices and Not Herbs}, for a $\Delta_{AB'}=0.32$ (Table 2). Finally, {\it Doberman Guard Dog} gives 0.88 and 0.27 with respect to {\it Pets} and {\it Not Farmyard Animals}, respectively, while it scores	0.55 with respect to {\it Pets And Not Farmyard Animals}, hence it scores $\Delta_{AB'}=0.28$ (Table 3).

(ii) in the conjunction `$A'$ and $B$'. Indeed, the membership weights of {\it Desk} with respect to {\it Not Home Furnishing}, {\it Furniture} and {\it Not Home Furnishing And Furniture} is 0.31, 0.95 and	0.75, respectively, for a $\Delta_{A'B}=0.44$ (Table 1). The exemplar {\it Oregano} scores 0.21 with respect to {\it Not Spices},  0.86 with respect to {\it Herbs}, and	0.5 with respect to {\it Not Spices and Herbs}, for a $\Delta_{A'B}=0.29$ (Table 2). Finally, again {\it Doberman Guard Dog} gives 0.14 and 0.76 with respect to {\it Not Pets} and {\it Farmyard Animals}, respectively, while it scores 0.45 with respect to {\it Not Pets And Farmyard Animals}, hence it scores $\Delta_{A'B}=0.45$ (Table 3).

When two concepts are negated -- `$A'$ and $B'$' -- we have, for example, $\mu(A')=0.12$, $\mu(B')=0.81$ and $\mu(A' \ {\rm and} \ B')=0.43$ for {\it Goldfish}, with respect to {\it Not Pets} and {\it Not Farmyard Animals}, hence $\Delta_{A'B'}=0.31$, in this case (Table 3). More, the exemplar {\it Garlic} scores $\mu(A')=0.88$ with respect to {\it Not Fruits} and $\mu(B')=0.24$ with respect to {\it Not Vegetables}, and $\mu(A' \ {\rm and} \ B')=0.45$ with respect to {\it Not Fruits And Not Vegetables}, for a $\Delta_{A'B'}=0.21$ (Table 4).

Double overextension is also present in various cases. For example, the membership weight of {\it Olive} with respect to {\it Fruits And Vegetables} is 0.65, which is greater than both 0.53 and 0.63, i.e. the membership weights of {\it Olive} with respect to {\it Fruits} and {\it Vegetables}, respectively (Table 4).	Furthermore, {\it Prize Bull} scores 0.13 with respect to {\it Pets} and 0.26 with respect to {\it Not Farmyard Animals}, but its membership weight with respect to {\it Pets And Not Farmayard Animals} is 0.28 (Table 3). Also, {\it Door Bell} gives 0.32 with respect to {\it Not Home Furnishing} and 0.33 with respect to {\it Furniture}, while it gives 0.34 with respect to {\it Not Home Furnishing And Furniture}.

Significant deviations from classicality are also due to conceptual negation, in the form of violation of the marginal law of classical probability theory. By again referring to Tables 1--4, we have that the exemplar {\it Field Mouse} has $I_{ABA'B'}$ in Equation (\ref{normalization}) equal to $-0.46$ (Table 3), while the exemplar {\it Doberman Guard Dog} has $I_{ABA'B'}=-1.03$ (Table 3). Both exemplars thus violate Equation (\ref{condbis05}). Analogously, {\it Chili Pepper} has $I_{A}$ in Equation (\ref{negationA}) equal to $-0.73$ (Table 2), hence it violates Equation (\ref{condbis01}), while {\it Pumpkin} has $I_{B'}$ in Equation (\ref{negationB'}) equal to $-0.13$ (Table 4), hence it violates Equation (\ref{condbis04}). 

We performed a statistical analysis of the data, estimating the probability that the experimentally identified deviations from classicality would be due to chance. We specifically considered the classicality conditions (\ref{condbis01})--(\ref{condbis05}) with the aim to prove that the deviations $I_{X}$, $X=A,A'$, $I_{Y}$, $Y=B,B'$ and $I_{ABA'B'}$ in Equations (\ref{normalization})--(\ref{negationB'}) were statistically significant. We firstly performed a `two-tail t-test for paired two samples for means' to test deviations from the marginal law of classical probability, that is, we tested violations of Equations (\ref{condbis01})--(\ref{condbis04}) by comparing $\mu(X)$ with respect to $\sum_{Y=B,B'}\mu(X,Y)$, $X=A,A'$, and $\mu(Y)$ with respect to $\sum_{X=A,A'}\mu(X \ {\rm and} \ Y)$, $Y=B,B'$. Then, we performed a `two-tail t-test for one sample for means' to test $\sum_{X=A,A'}\sum_{Y=B,B'}\mu(X {\rm and} \ Y)$ with respect to 1. The corresponding p-values for ${\rm df}=37$ are reported in Tables 5a--e, Appendix \ref{tables}. Due to the high number of multiple comparisons -- 24 null hypotheses were tested for each pair $(X,Y)$ of concepts -- we applied a `Bonferroni correction procedure' to avoid the so-called `family-wise error rate' (FWER). Hence, we compared the obtained p-values with the reference value $0.05/24\approx 0.002$. Notwithstanding the fact that the Bonferroni procedure is considered as conservative, we found p-values systematically much lower than this reference value, for all exemplars and pairs of concepts, which makes it possible to conclude that the experimentally tested deviations from classicality are not due to chance. 

It is finally interesting to notice that the deviations from Equations (\ref{condbis01})--(\ref{condbis05}) show similar patterns, characterized by approximately constant numerical values. This observation led us to suspect that $I_{X}$, $I_{Y}$, $X=A,A',Y=B,B'$ and $I_{ABA'B'}$ are constant functions across all exemplars. To this end we first, performed a `linear regression analysis' of the data, sorted from smaller to larger, so we can check whether these quantities can be represented by a line of the form $y=mx+q$, with $m=0$. This was indeed the case.  For $I_{A}$, we obtained $m=3.0 \cdot 10^{-3}$ with $R^{2}=0.94$; for $I_{B}$, we obtained $m=2.9 \cdot 10^{-3}$ with $R^{2}=0.93$; for $I_{A'}$, we obtained $m=2.6 \cdot 10^{-3}$ with $R^{2}=0.96$; for $I_{B'}$, we obtained $m=3.1 \cdot 10^{-3}$ with $R^{2}=0.98$; for $I_{ABA'B'}$, we obtained $m=4 \cdot 10^{-3}$ with $R^{2}=0.92$. Next, we computed the $95\%$-confidence interval for these parameters and obtained interval $(-0.51, -0.33)$ for $I_A$, interval $(-0.42, -0.28)$ for $I_{A'}$, interval $(-0.52, -0.34)$ for $I_B$, interval $(-0.40, -0.26)$ for $I_{B'}$, and interval $(-0.97, -0.64)$ for $I_{AA'BB'}$. Hence, we can conclude that the measured parameters systematically fall within a narrow band centered at very similar values. 

This pattern is incredibly stable, systematic and regular, because it is independent on the specific exemplar, the pair of concepts and the type of conjunction that are considered. And we believe that this deviation from classical rules occurs at a deeper, more fundamental, level than the known deviation due to overextension, as mentioned in Section \ref{intro}. In our opinion, it expresses a fundamental mechanism through which the human mind forms concepts, hence it reveals aspects of conceptual formation, even before aspects of conceptual combination. We will see in Section  \ref{results} that the stability of this violation can exactly be explained in a two-sector Fock space quantum framework.

The results obtained in our experiments indicate systematic deviations from the rules of classical (fuzzy set) logic and probability theory, exactly as we have encountered in other situations \cite{a2009a,s2014a,s2014b}. We will see in the next sections these `non-classicalities' are of a quantum-type, and hence they can be described within the mathematical formalism of quantum theory. The essentials of this formalism are reviewed in Appendix  \ref{quantum} and we refer to it for symbols and notation.

\section{Quantum modeling conceptual conjunctions and negations\label{brussels}}
In \cite{a2009a} we proved that a big amount of the experimental data collected in \cite{h1988a,h1988b} on conjunctions and disjunctions of two concepts can be modeled by using the mathematical formalism of quantum theory. A two-sector Fock space then provided an optimal algebraic setting for this modeling (see Appendix \ref{quantum}). In \cite{s2014a} we proved that this quantum-theoretic framework was suitable to model the data collected in \cite{ap2011} on `borderline contradictions', and in \cite{s2014b} we were able to prove that also the experimental data collected on conjunctions of the form `$A \ {\rm and} \ B$' and `$A \ {\rm and} \ B'$', for specific pairs $(A,B)$ of concepts can be represented by using the same quantum mathematics. However, a complete modeling of data on both conjunctions and negations requires performing new experiments, where the conceptual conjunctions `$A \ {\rm and} \ B$' and `$A \ {\rm and} \ B'$' are tested together with the conceptual conjunctions `$A' \ {\rm and} \ B$' and `$A' \ {\rm and} \ B'$'. The complete collection of these experiments has been discussed in Section \ref{experiment}. As anticipated in Section \ref{intro}, we undertake this modeling task here. It is natural to observe that the modeling in \cite{a2009a} needs a suitable generalization, since conceptual negation should be taken into account as well. But, we will see later in this section that such a generalization is completely compatible with the original model, because it rests on the assumption that (probabilistic versions of) logical rules hold only in second sector of Fock space. By introducing this quite natural assumption, we were able to model conceptual conjunctions and disjunctions in Fock space. We show now that conceptual conjunctions and negations can be modeled in Fock space by introducing the same assumption.

To model conceptual negations we also need a new theoretical step which was not necessary in our previous formulations, namely, the introduction of `entangled states' in second sector of Fock space, which implies that membership weights are not independent. This introduction, together with the application of quantum logical rules in second sector of Fock space, are compatible with previous formulations, but they make our generalization in this paper highly non-obvious. We will extensively discuss the novelties of the present modeling in the next sections. Let us first proceed with our mathematical construction.

Let us denote by $\mu(A)$, $\mu(B)$, $\mu(A')$, $\mu(B')$, $\mu(A \ {\rm and} \ B)$, $\mu(A \ {\rm and} \ B')$, $\mu(A' \ {\rm and} \ B)$ and $\mu(A' \ {\rm and} \ B')$ the membership weights of a given exemplar $x$ with respect to the concepts $A$, $B$, the negations $A'$, $B'$ and the conjunctions `$A \ {\rm and} \ B$', `$A \ {\rm and} \ B'$', `$A' \ {\rm and}\ B$' and `$A' \ {\rm and} \ B'$', respectively.

The decision measurement testing whether a specific exemplar $x$ is a member or not a member of a concept $A$ is represented by the spectral decomposition consisting of the two orthogonal projection operators $M$ and $\mathbbmss{1}-M$ defined in a complex Hilbert space ${\cal H}$. Let us represent the concepts $A$ and $B$ by the unit vectors $|A\rangle$ and $|B\rangle$, respectively, of ${\cal H}$, and assume they are orthogonal.\footnote{The choice of orthogonal vectors is linked to the quantum mathematics-type of our modeling. In quantum physics, indeed, two collapsed states resulting from different measurement outcomes are represented by orthogonal vectors. This choice also allows to make a precise link  between interference effects and complex amplitudes in our model (see Appendix \ref{quantum}).} This means that we have
\begin{equation}
\langle A|A\rangle=\langle B|B\rangle=1 \quad \langle A|B\rangle =0
\end{equation}
By using standard rules for quantum probabilities (see Appendix \ref{quantum}), we get that the following two requirements need to be satisfied.
\begin{equation}
\label{BornRule}
\mu(A)=\langle A|M|A\rangle \quad \mu(B)=\langle B|M|B\rangle
\end{equation}
where $\mu(A)$ and $\mu(B)$ are the measured membership weights of $x$ with respect to the concepts $A$ and $B$, respectively, in the performed experiment.

Next we consider two unit vectors $|A'\rangle$, $|B'\rangle$ in ${\cal H}$, such that the set $\{|A\rangle, |B\rangle, |A'\rangle, |B'\rangle\}$, is an orthonormal set. The conceptual negations $A'$ and $B'$, which we interpret as emergent concepts, are represented by $|A'\rangle$ and $|B'\rangle$, respectively. Therefore, we need two additional requirements to be satisfied, namely,
\begin{equation}
\label{BornRuleNeg}
\mu(A')=\langle A'|M|A'\rangle \quad \mu(B')=\langle B'|M|B'\rangle
\end{equation}
where $\mu(A')$ and $\mu(B')$ are the measured membership weights of $x$ with respect to the negations $A'$ and $B'$ of the  concepts $A$ and $B$, respectively, in the performed experiment.

\subsection{The first sector analysis\label{onesector}}
Let us first analyse the situation were we look for a modeling solution only in the Hilbert space ${\cal H}$, or first sector of Fock space (see Appendix \ref{quantum}). In this case, the concepts `$A$ and $B$', `$A$ and $B'$', `$A'$ and $B$', and `$A'$ and $B'$' are respectively represented  by the superposition vectors\footnote{We introduce in this model a superposition vector with equal weights on the two vectors, for the sake of simplicity. The most general case of a weighted superposition can be considered in future investigation, and it is an interesting line of research in itself, as the interpretation of the weights is not trivial.} ${1 \over \sqrt{2}}(|A\rangle + |B\rangle)$, ${1 \over \sqrt{2}}(|A\rangle + |B'\rangle)$, ${1 \over \sqrt{2}}(|A'\rangle + |B\rangle)$, and ${1 \over \sqrt{2}}(|A'\rangle + |B'\rangle)$.

Let us analyse in detail the aspects of this situation with the aim of resulting in a view on the possible solutions. Geometric considerations induce to observe that, if we look for a solution in the complex Hilbert space $\compl^8$, we will find the most general type of solution. Indeed, since we consider four vectors $|A\rangle$, $|A'\rangle$, $|B\rangle$ and $|B'\rangle$, and the measurement entails two projection operators to measure membership, $M$ and its orthogonal projector $\mathbbmss{1}-M$, then the maximal number of subspaces we can obtain when applying the operators to the vectors is eight. Thus, the choice of a Hilbert space with more than eight dimensions would not add degrees of freedom that can give rise to additional solutions to those that can be found in $\compl^8$. Hence, we explicitly use this Hilbert space in what follows reminding, however, that our results hold in any higher dimensional Hilbert space. 
Let $\{ |1\rangle=(1,\ldots,0)$, $|2\rangle=(0,1,\ldots,0)$, \ldots, $|8\rangle=(0,0,\ldots,1)\} $ denote the canonical base of $\compl^8$.
We construct a representation in $\compl^8$ where $M$ projects on the subspace $\compl^4$ generated by the last four vectors of this canonical base, and $\mathbbmss{1}-M$ on the  subspace $\compl^4$ generator by the first four vectors of it. If we set
\begin{eqnarray}
&&|A\rangle=e^{i\phi_A}(a_1, a_2, a_3, a_4, a_5, a_6, a_7, a_8) \\
&&|A'\rangle=e^{i\phi_{A'}}(a'_1, a'_2, a'_3, a'_4, a'_5, a'_6, a'_7, a'_8) \\
&&|B\rangle=e^{i\phi_B}(b_1, b_2, b_3, b_4, b_5, b_6, b_7, b_8) \\
&&|B'\rangle=e^{i\phi_{B'}}(b'_1, b'_2, b'_3, b'_4, b'_5, b'_6, b'_7, b'_8)
\end{eqnarray}
then, Equations~\eqref{BornRule} and~\eqref{BornRuleNeg} become
\begin{eqnarray} \label{firstsectoreq01}  
&&\mu(A)=\langle A|M|A\rangle=a_5^2+a_6^2+a_7^2+a_8^2 \\ \label{firstsectoreq02}
&&1-\mu(A)=\langle A|\mathbbmss{1}-M|A\rangle=a_1^2+a_2^2+a_3^2+a_4^2 \\ \label{firstsectoreq03}
&&\mu(A')=\langle A'|M|A'\rangle={a'}_5^2+{a'}_6^2+{a'}_7^2+{a'}_8^2 \\ \label{firstsectoreq04}
&&1-\mu(A')=\langle A'|\mathbbmss{1}-M|A'\rangle={a'}_1^2+{a'}_2^2+{a'}_3^2+{a'}_4^2 \\ \label{firstsectoreq05}
&&\mu(B)=\langle B|M|B\rangle=b_5^2+b_6^2+b_7^2+b_8^2 \\ \label{firstsectoreq06}
&&1-\mu(B)=\langle B|\mathbbmss{1}-M|B\rangle=b_1^2+b_2^2+b_3^2+b_4^2 \\ \label{firstsectoreq07}
&&\mu(B')=\langle B'|M|B'\rangle={b'}_5^2+{b'}_6^2+{b'}_7^2+{b'}_8^2 \\ \label{firstsectoreq08}
&&1-\mu(B')=\langle B'|\mathbbmss{1}-M|B'\rangle={b'}_1^2+{b'}_2^2+{b'}_3^2+{b'}_4^2,
\end{eqnarray}
and the orthogonality conditions become
\begin{eqnarray}
\label{firstsectoreq09}
&&0=\langle A|A'\rangle=a_1a'_1+a_2a'_2+a_3a'_3+a_4a'_4+a_5a'_5+a_6a'_6+a_7a'_7+a_8a'_8 \\ \label{firstsectoreq10}
&&0=\langle B|B'\rangle=b_1b'_1+b_2b'_2+b_3b'_3+b_4b'_4+b_5b'_5+b_6b'_6+b_7b'_7+b_8b'_8 \\ \label{firstsectoreq11}
&&0=\langle A|B\rangle=a_1b_1+a_2b_2+a_3b_3+a_4b_4+a_5b_5+a_6b_6+a_7b_7+a_8b_8 \\ \label{firstsectoreq12}
&&0=\langle A|B'\rangle=a_1b'_1+a_2b'_2+a_3b'_3+a_4b'_4+a_5b'_5+a_6b'_6+a_7b'_7+a_8b'_8 \\ \label{firstsectoreq13}
&&0=\langle A'|B\rangle=a'_1b_1+a'_2b_2+a'_3b_3+a'_4b_4+a'_5b_5+a'_6b_6+a'_7b_7+a'_8b_8 \\ \label{firstsectoreq14}
&&0=\langle A'|B'\rangle=a'_1b'_1+a'_2b'_2+a'_3b'_3+a'_4b'_4+a'_5b'_5+a'_6b'_6+a'_7b'_7+a'_8b'_8. 
\end{eqnarray}
A solution of equations (\ref{firstsectoreq01})--
(\ref{firstsectoreq14}), gives us a configuration of the four orthonormal vectors $|A\rangle$, $|A'\rangle$, $|B\rangle$ and $|B'\rangle$ in $\compl^8$, such that self-adjoint operator formed by the spectral decomposition of the two orthogonal projections $M$ and $\mathbbmss{1}-M$ give rise to the values $\mu(A)$, $1-\mu(A)$, $\mu(A')$, $1-\mu(A')$, $\mu(B)$, $1-\mu(B)$ and $\mu(B')$, $1-\mu(B')$, corresponding to the measured data related to the concepts $A$ and $B$, and their negations  `$A'$'  and `$B'$'.

To compute the membership weight of the conjunction of two concepts, we apply the quantum probability rule for the membership operator on the state vectors of the respective concept combinations. Such application lead to the following equations
\begin{eqnarray}
\mu(A\ {\rm and}\ B)&=&{1 \over 2}(\mu(A)+\mu(B))+\Re\langle A|M|B\rangle \\
\mu(A\ {\rm and}\ B')&=&{1 \over 2}(\mu(A)+\mu(B'))+\Re\langle A|M|B'\rangle \\
\mu(A'\ {\rm and}\ B)&=&{1 \over 2}(\mu(A')+\mu(B))+\Re\langle A'|M|B\rangle \\
\mu(A'\ {\rm and}\ B')&=&{1 \over 2}(\mu(A')+\mu(B'))+\Re\langle A'|M|B'\rangle
\end{eqnarray}
Hence, in our $8-$dimensional complex Hilbert space model these equations become
\begin{eqnarray}
 \label{firstsectoreq15}
&&\mu(A\ {\rm and}\ B)={1 \over 2}(\mu(A)+\mu(B))+\Re\langle A|M|B\rangle \nonumber \\
&&={1 \over 2}(\mu(A)+\mu(B))+(a_5b_5+a_6b_6+a_7b_7+a_8b_8)\cos(\phi_B-\phi_A) \\ \label{firstsectoreq16}
&&\mu(A\ {\rm and}\ B')={1 \over 2}(\mu(A)+\mu(B'))+\Re\langle A|M|B'\rangle \nonumber \\ 
&&={1 \over 2}(\mu(A)+\mu(B'))+(a_5b'_5+a_6b'_6+a_7b'_7+a_8b'_8)\cos(\phi_{B'}-\phi_A) \\ \label{firstsectoreq17}
&&\mu(A'\ {\rm and}\ B)={1 \over 2}(\mu(A')+\mu(B))+\Re\langle A'|M|B\rangle \nonumber \\
&&={1 \over 2}(\mu(A')+\mu(B))+(a'_5b_5+a'_6b_6+a'_7b_7+a'_8b_8)\cos(\phi_B-\phi_{A'}) \\ \label{firstsectoreq18}
&&\mu(A'\ {\rm and}\ B')={1 \over 2}(\mu(A')+\mu(B'))+\Re\langle A'|M|B'\rangle \nonumber \\
&&={1 \over 2}(\mu(A')+\mu(B'))+(a'_5b'_5+a'_6b'_6+a'_7b'_7+a'_8b'_8)\cos(\phi_{B'}-\phi_{A'})
\end{eqnarray}
We can now analyze whether or not the solution of Equations \eqref{firstsectoreq01}--\eqref{firstsectoreq14} is compatible with Equations \eqref{firstsectoreq15}--\eqref{firstsectoreq18}. To this end note that the right hand side of Equations \eqref{firstsectoreq15}--\eqref{firstsectoreq18} correspond to the average of the probabilities of the former concepts, plus the so called `interference term', which depends on (i) how the vectors representing the former concepts in the combination, when restricted to the subspace determined by $M$, project into each other, and (ii) on the phase angles of the vectors \cite{a2009a}. Note that the configuration of the phase angles $\phi_A,\phi_B,\phi_{A'},\phi_{B'}$ allow to model a variety of interference situations. In fact, when the difference between these phase angles is close to $0$ or $\pi$, we have a maximal amount of interference while, when the difference between these phase angles is close to $\frac{\pi}{2}$ or $\frac{3\pi}{2}$, we have a minimal amount of interference. We can then characterize a set of `compatibility intervals' for the solution of Equations \eqref{firstsectoreq01}--\eqref{firstsectoreq14} and \eqref{firstsectoreq15}--\eqref{firstsectoreq18} which determines the modeling capacity of this Hilbert space model. Let,
\begin{eqnarray}
i(A,B)_1&=&{1 \over 2}(\mu(A)+\mu(B))-|a_5b_5+a_6b_6+a_7b_7+a_8b_8| \\
i(A,B)_2&=&{1 \over 2}(\mu(A)+\mu(B))+|a_5b_5+a_6b_6+a_7b_7+a_8b_8| \\
i(A,B')_1&=&{1 \over 2}(\mu(A)+\mu(B'))-|a_5b'_5+a_6b'_6+a_7b'_7+a_8b'_8| \\
i(A,B')_2&=&{1 \over 2}(\mu(A)+\mu(B'))+|a_5b'_5+a_6b'_6+a_7b'_7+a_8b'_8|\\
i(A',B)_1&=&{1 \over 2}(\mu(A')+\mu(B))-|a'_5b_5+a'_6b_6+a'_7b_7+a'_8b_8|\\
i(A',B)_2&=&{1 \over 2}(\mu(A')+\mu(B))+|a'_5b_5+a'_6b_6+a'_7b_7+a'_8b_8|\\
i(A',B')_1&=&{1 \over 2}(\mu(A')+\mu(B'))-|a'_5b'_5+a'_6b'_6+a'_7b'_7+a'_8b'_8|\\
i(A',B')_2&=&{1 \over 2}(\mu(A')+\mu(B'))+|a'_5b'_5+a'_6b'_6+a'_7b'_7+a'_8b'_8|,
\end{eqnarray}
and let us define the following `solution intervals'
\begin{eqnarray} \label{firstintsolAB} 
{\mathscr I}_{AB}&=&[i(A,B)_1,i(A,B)_2] \\ \label{firstintsolAB'} 
{\mathscr I}_{AB'}&=&[i(A,B')_1,i(A,B')_2] \\ \label{firstintsolA'B} 
{\mathscr I}_{A'B}&=&[i(A',B)_1,i(A',B)_2] \\ \label{firstintsolA'B'} 
{\mathscr I}_{A'B'}&=&[i(A',B')_1,i(A',B')_2]. 
\end{eqnarray}
A solution of Equations~\eqref{firstsectoreq01}-\eqref{firstsectoreq18} exists if and only if the membership weights $\mu(A\ {\rm and}\ B),$ $\mu(A'\ {\rm and}\ B),$ $\mu(A\ {\rm and}\ B')$ and $\mu(A\ {\rm and}\ B)$ are respectively contained in the intervals ${\mathscr I}_{AB}$, ${\mathscr I}_{A'B}$, ${\mathscr I}_{AB'}$ and ${\mathscr I}_{A'B'}$.

By analogy with what we found in \cite{a2009a} and \cite{s2014a,s2014b}, it is reasonable to expect that our experimental data in Appendix \ref{tables} cannot be generally modeled in the complex Hilbert space $\compl^8$, or first sector of Fock space, but a second sector $\compl^8 \otimes \compl^8$ of Fock space is also needed (see Appendix \ref{quantum}). Consider for example a simple case that applies for classical logics $\mu(A)=1, \mu(B)=0$ and $\mu(A~\mbox{and}~B)=0$. This case is consistent with the minimum conjunction rule \cite{z1965} but not with our first sector of Fock space (Hilbert space) model. 
We will see that this type of cases is compatible with second sector of Fock space, and show that a framework that encompasses  both `logical' and  `emergent' reasoning about membership in these situations requires the general properties of a Fock space. To make our description complete however, we first have to introduce a new, conceptually relevant, ingredient.

\subsection{Introducing entanglement in conceptual combinations\label{entanglement}}
In \cite{a2009a} and \cite{s2014a,s2014b} we successfully modeled conjunctions of the form `$A$ and $B$' in a Fock space constructed as the direct sum of an individual Hilbert space ${\cal H}$, or `first sector of Fock space', and a tensor product Hilbert space ${\cal H} \otimes {\cal H}$, or `second sector of Fock space' (see Appendix \ref{quantum}). The concepts $A$ and $B$ were respectively represented by the unit vectors $|A\rangle$ and $|B\rangle$ of ${\cal H}$, while the conjunction `$A$ and $B$' was represented by the unit vector ${1 \over \sqrt{2}}(|A\rangle+|B\rangle)$ in first sector, and by the tensor product vector $|A\rangle \otimes |B\rangle$ in second sector. The decision measurement of a subject who estimates whether a given exemplar $x$ is a member of `$A$ and $B$'  was represented by the orthogonal projection operator $M$ in first sector, and by the tensor product projection operator $M \otimes M$ in second sector. The conjunction `$A$ and $B$' was represented by a unit vector of the form $\psi(A,B)=me^{i \theta}|A\rangle \otimes |B\rangle+ne^{i \rho}{1 \over \sqrt{2}}(|A\rangle+|B\rangle)$ in the Fock space ${\cal H} \oplus {\cal H} \otimes {\cal H}$, while the decision measurement was represented by the orthogonal projection operator $M \oplus  (M \otimes M)$ in the same Fock space. By using quantum probabilistic rules, one could then write the membership weight of $x$ with respect to `$A$ and $B$' as $\mu(A \ {\rm and} \ B)=\langle \psi(A,B)|(M \otimes M) \oplus M|\psi(A,B)\rangle=m^2\mu(A)\mu(B)+n^2( {1 \over 2}(\mu(A)+\mu(B))+\Re \langle A|M|B\rangle)$. This treatment needs now to be generalized to the decision measurement of the concepts $A$, $B$, the negations $A'$, $B'$ and the conjunctions `$A \ {\rm and} \ B$', `$A \ {\rm and} \ B'$, `$A' \ {\rm and} \ B$' and `$A' \ {\rm and}\ B'$'. The first sector situation has already been analysed in Section \ref{onesector} where we have also constructed an explicit representation in the complex Hilbert space $\compl^8$. In these two sections we instead analyse the second sector situation, but we allow for the possibility of representing concepts by entangled states too (see Appendix \ref{quantum}).

Is it possible to introduce some `type of entanglement' in second sector $\compl^8 \otimes \compl^8$? This question is interesting, in our opinion, since it is reasonable to believe that the outcomes of experiments for $A$ are not independent of the outcomes of experiments for $B$. For example, in case a specific exemplar $x$ is strongly a member of {\it Fruits}, this will influence the strength of membership of {\it Vegetables}, and viceversa, because the meanings of {\it Fruits} and {\it Vegetables} are not independent. And this apparently occurs for all human concepts. Suppose we combined, for example, {\it Fruits} with {\it Not Fruits}, then one would expect to exist, for any exemplar, a substantial amount of anti-correlation between it being a member of {\it Fruits} and it being a member of {\it Not Fruits}. How can we express the general situation, where anti-correlation, as well as correlation, are possible to be increased or decreased by parameters? In the foregoing modeling \cite{a2009a,ags2013} we chose the simplest representation for the situation in second sector, namely the product state $|A\rangle \otimes |B\rangle$, which leads to a situation of complete independence between $A$ and $B$, for whatever exemplar tested. Let us investigate what would be a situation for a general entangled state, and how a two-sector Fock space already incorporates this possibility.

Suppose that the concept `$A$ and $B$' is not represented by the product state vector $|A\rangle \otimes |B\rangle$ in second sector of Fock space $\compl^8 \otimes \compl^8$, but by a general entangled state vector $|C\rangle$ of  $\compl^8 \otimes \compl^8$. We remind that $\compl^8$ is the concrete Hilbert space we constructed in Section \ref{onesector}. In the canonical base  $\{|i\rangle\}_{i=1,\ldots,8}$ of $\compl^8$, we have
\begin{eqnarray}
|C\rangle=\sum_{i,j=1}^{8} c_{ij}e^{i\gamma_{ij}}|i\rangle \otimes |j\rangle
\end{eqnarray}
and
\begin{eqnarray}
1=\langle C|C\rangle&=&(\sum_{k,l=1}^{8} c_{kl}e^{-i\gamma_{kl}}\langle k| \otimes \langle l|)(\sum_{i,j=1}^8 c_{ij}e^{i\gamma_{ij}}|i\rangle \otimes |j\rangle)=\sum_{k,l=1}^8\sum_{i,j=1}^8 c_{kl}c_{ij}e^{i(\gamma_{ij}-\gamma_{kl})}\langle k|i\rangle \langle l|j\rangle \nonumber \\
&=&\sum_{i,j=1}^8 c_{ij}c_{ij}e^{i(\gamma_{ij}-\gamma_{ij})}=\sum_{i,j=1}^8 c_{ij}^2
\end{eqnarray}
We express the effect, as described in second sector, of the experiments where subjects were asked to decide for membership (or not membership) of a specific exemplar with respect to the concepts $A$ and $B$, as follows. Membership with respect to $A$, as a yes-no measurement, is described by the orthogonal projection operators $M \otimes \mathbbmss{1}$, $(\mathbbmss{1}-M) \otimes \mathbbmss{1}$, as spectral family of the corresponding self-adjoint operator. Hence in second sector, tests on concept $A$, are described in the first component of the tensor product Hilbert space $\compl^8 \otimes \compl^8$, which forms second sector. In an analogous way, tests on concept $B$, are described in the second component of the tensor product, by the orthogonal projection operators $\mathbbmss{1} \otimes M$, $\mathbbmss{1} \otimes (\mathbbmss{1}-M)$, as spectral family of the corresponding self-adjoint operator. Remark that we do not introduce in any way the concepts $A'$ and $B'$ for the second sector description, and also not conjunction of $A$ and $B$ with them, at least the aspect of these conjunctions that represent new emergent concepts. The concept $A'$ and $B'$ are indeed `emergent entities' because the negation on a concept to give rise to a new concept, namely the negation concept.

Also the experimental data collected on $A'$, $B'$ and combinations of them with $A$ and $B$ do not appear in second sector. All emergence is indeed modeled in first sector. This means that also the conjunction of $A$ and $B$ as a new emergent concept does not appear in second sector, it only appears in first sector modeled there by the superposition. All non-emergent equivalents of these are described by the tensor product of the two self-adjoint operators corresponding to the yes-no experiments with respect to membership performed on concepts $A$ and consequently on concept $B$, exactly as in our real life experiment that gave rise to our data on $A$ and $B$. This means that the orthogonal projectors of the spectral family of this tensor product self adjoint operator describe all cases of non emergence. This family consists of $\{M \otimes M, M \otimes (\mathbbmss{1}-M), (\mathbbmss{1}-M) \otimes M, (\mathbbmss{1}-M) \otimes (\mathbbmss{1}-M)\}$. 
Let us express this on a general entangled state $|C\rangle$. We have
\begin{eqnarray}
\mu(A)&=&\langle C|M\otimes \mathbbmss{1}|C\rangle \nonumber \\ &=&(\sum_{k,l=1}^8 c_{kl}e^{-i\gamma_{kl}}\langle k| \otimes \langle l|) M \otimes \mathbbmss{1}|(\sum_{i,j=1}^8 c_{ij}e^{i\gamma_{ij}}|i\rangle \otimes |j\rangle)=\sum_{k,l=1}^8\sum_{i,j=1}^8 c_{kl}c_{ij}e^{i(\gamma_{ij}-\gamma_{kl})}\langle k|M|i\rangle \langle l|\mathbbmss{1}|j\rangle \nonumber \\ \label{entangledstate01}
&=&\sum_{i,j,k=1}^8 c_{kj}c_{ij}e^{i(\gamma_{ij}-\gamma_{kj})}\langle k|M|i\rangle=\sum_{i=5}^8\sum_{j=1}^8c_{ij}c_{ij}e^{i(\gamma_{ij}-\gamma_{ij})}=\sum_{i=5}^8\sum_{j=1}^8c_{ij}^2  \\
\mu(B)&=&\langle C|\mathbbmss{1}\otimes M|C\rangle \nonumber \\ &=& (\sum_{k,l=1}^8 c_{kl}e^{-i\gamma_{kl}}\langle k| \otimes \langle l|) \mathbbmss{1} \otimes M|(\sum_{i,j=1}^8 c_{ij}e^{i\gamma_{ij}}|i\rangle \otimes |j\rangle)=\sum_{k,l=1}^8\sum_{i,j=1}^8 c_{kl}c_{ij}e^{i(\gamma_{ij}-\gamma_{kl})}\langle k|\mathbbmss{1}|i\rangle \langle l|M|j\rangle \nonumber \\ \label{entangledstate02}
&=&\sum_{i,j,l=1}^8 c_{il}c_{ij}e^{i(\gamma_{ij}-\gamma_{il})} \langle l|M|j\rangle=\sum_{i=1}^8\sum_{j=5}^8c_{ij}c_{ij}e^{i(\gamma_{ij}-\gamma_{ij})}=\sum_{i=1}^8\sum_{j=5}^8c_{ij}^2
\end{eqnarray}
We also have
\begin{eqnarray}
1-\mu(A)&=&\langle C|(\mathbbmss{1}-M)\otimes \mathbbmss{1}|C\rangle=(\sum_{k,l=1}^8 c_{kl}e^{-i\gamma_{kl}}\langle k| \otimes \langle l|) (\mathbbmss{1}-M) \otimes \mathbbmss{1}|(\sum_{i,j=1}^8 c_{ij}e^{i\gamma_{ij}}|i\rangle \otimes |j\rangle) \nonumber \\
&=&\sum_{k,l=1}^8\sum_{i,j=1}^8 c_{kl}c_{ij}e^{i(\gamma_{ij}-\gamma_{kl})}\langle k|\mathbbmss{1}-M|i\rangle \langle l|\mathbbmss{1}|j\rangle=\sum_{i,j,k=1}^8 c_{kj}c_{ij}e^{i(\gamma_{ij}-\gamma_{kj})}\langle k|\mathbbmss{1}-M|i\rangle \nonumber \\ \label{entangledstate03}
&=&\sum_{i=1}^4\sum_{j=1}^8 c_{ij}c_{ij}e^{i(\gamma_{ij}-\gamma_{ij})}=\sum_{i=1}^4\sum_{j=1}^8 c_{ij}^2 \\
1-\mu(B)&=&\langle C|\mathbbmss{1}\otimes (\mathbbmss{1}-M)|C\rangle=(\sum_{k,l=1}^8 c_{kl}e^{-i\gamma_{kl}}\langle k| \otimes \langle l|) \mathbbmss{1} \otimes (\mathbbmss{1}-M)|(\sum_{i,j=1}^8 c_{ij}e^{i\gamma_{ij}}|i\rangle \otimes |j\rangle) \nonumber \\
&=&\sum_{k,l=1}^8\sum_{i,j=1}^8 c_{kl}c_{ij}e^{i(\gamma_{ij}-\gamma_{kl})}\langle k|\mathbbmss{1}|i\rangle \langle l|\mathbbmss{1}-M|j\rangle=\sum_{i,j,l=1}^8 c_{il}c_{ij}e^{i(\gamma_{ij}-\gamma_{il})} \langle l|\mathbbmss{1}-M|j\rangle \nonumber \\ \label{entangledstate04}
&=&\sum_{i=1}^8\sum_{j=1}^4 c_{ij}c_{ij}e^{i(\gamma_{ij}-\gamma_{ij})}=\sum_{i=1}^8\sum_{j=1}^4 c_{ij}^2
\end{eqnarray}
And further we have
\begin{eqnarray}
\langle C|M\otimes M|C\rangle&=&(\sum_{k,l=1}^8 c_{kl}e^{-i\gamma_{kl}}\langle k| \otimes \langle l|) M \otimes M|(\sum_{i,j=1}^8 c_{ij}e^{i\gamma_{ij}}|i\rangle \otimes |j\rangle)\nonumber \\
&=&\sum_{k,l=1}^8\sum_{i,j=1}^8 c_{kl}c_{ij}e^{i(\gamma_{ij}-\gamma_{kl})}\langle k|M|i\rangle \langle l|M|j\rangle\nonumber \\ \label{entangledstate05}
&=&\sum_{i=5}^8\sum_{j=5}^8 c_{ij}c_{ij}e^{i(\gamma_{ij}-\gamma_{ij})}=\sum_{i=5}^8\sum_{j=5}^8 c_{ij}^2 \\ \label{entangledstate06}
\langle C|M\otimes (\mathbbmss{1}-M)|C\rangle&=&\sum_{i=5}^8\sum_{j=1}^4 c_{ij}^2 \\ \label{entangledstate07}
\langle C|(\mathbbmss{1}-M)\otimes M|C\rangle&=&\sum_{i=1}^4\sum_{j=5}^8 c_{ij}^2 \\ \label{entangledstate08}
\langle C|(\mathbbmss{1}-M)\otimes (\mathbbmss{1}-M)|C\rangle&=&\sum_{i=1}^4\sum_{j=1}^4 c_{ij}^2
\end{eqnarray}
The values of $\langle C|M\otimes M|C\rangle$, $\langle C|M\otimes (\mathbbmss{1}-M)|C\rangle$, $\langle C|(\mathbbmss{1}-M)\otimes M|C\rangle$ and $\langle C|(\mathbbmss{1}-M)\otimes (\mathbbmss{1}-M)|C\rangle$ will respectively represent the amounts that within our Fock space model second sector contributes to the values of $\mu(A\ {\rm and}\ B)$, $\mu(A\ {\rm and}\ B')$, $\mu(A'\ {\rm and}\ B)$ and $\mu(A'\ {\rm and}\ B')$.
We can prove that the second sector theoretical values, allowing the state to be a general entangled state in our $\compl^8\otimes \compl^8$ Hilbert space model, reach exactly the values to be found for the case the classicality conditions (\ref{condbis01})--(\ref{condbis05}) in Theorem 3$'$ are satisfied. More explicitly, the following theorem holds.

\bigskip
\noindent
{\bf Theorem 4.} \emph{If the experimentally collected membership weights $\mu(A)$, $\mu(B)$, $\mu(A')$, $\mu(B')$, $\mu(A\ {\rm and}\ B)$, $\mu(A \ {\rm and}\ B')$, $\mu(A'\ {\rm and}\ B)$ and $\mu(A'\ {\rm and}\ B')$ can be represented in second sector of Fock space for a given choice of the entangled state vector $|C\rangle$ and the decision measurement projection operator $M$, then the membership weights satisfy (\ref{condbis01})--(\ref{condbis05}), hence they are classical data. Viceversa, if $\mu(A)$, $\mu(B)$, $\mu(A')$, $\mu(B')$, $\mu(A\ {\rm and}\ B)$, $\mu(A \ {\rm and}\ B')$, $\mu(A'\ {\rm and}\ B)$ and $\mu(A'\ {\rm and}\ B')$ satisfy (\ref{condbis01})--(\ref{condbis05}), hence they are classical data, then an entangled state vector $|C\rangle$ and a decision measurement projection operator $M$ can always be found such that $\mu(A)$, $\mu(B)$, $\mu(A')$, $\mu(B')$, $\mu(A\ {\rm and}\ B)$, $\mu(A \ {\rm and}\ B')$, $\mu(A'\ {\rm and}\ B)$ and $\mu(A'\ {\rm and}\ B')$ can be represented in second sector of Fock space.}

\medskip
\noindent
{\bf Proof.} 
Suppose that the theoretical values for a modeling only in second sector of Fock space are given. This means that $|C\rangle$ and $M$ are given, and the values of $\mu(A)$, $\mu(B)$, $\mu(A')$, $\mu(B')$, $\mu(A\ {\rm and}\ B)$, $\mu(A\ {\rm and}\ B')$, $\mu(A'\ {\rm and}\ B)$ and $\mu(A'\ {\rm and}\ B')$ are given respectively by $\langle C|M\otimes \mathbbmss{1}|C\rangle$, $\langle C|\mathbbmss{1}\otimes M|C\rangle$, $\langle C|(\mathbbmss{1}-M)\otimes \mathbbmss{1}|C\rangle$, $\langle C|\mathbbmss{1}\otimes (\mathbbmss{1}-M)|C\rangle$,$\langle C|M\otimes M|C\rangle$, $\langle C|M\otimes (\mathbbmss{1}-M)|C\rangle$, $\langle C|(\mathbbmss{1}-M)\otimes M|C\rangle$ and $\langle C|(\mathbbmss{1}-M)\otimes (\mathbbmss{1}-M)|C\rangle$. If we use the results calculated in (\ref{entangledstate01}), (\ref{entangledstate02}), (\ref{entangledstate03}), (\ref{entangledstate04}), (\ref{entangledstate05}), (\ref{entangledstate06}), (\ref{entangledstate07}) and (\ref{entangledstate08}), we can easely prove all the classicality conditions  (\ref{condbis01})--(\ref{condbis05}) to be satisfied. Let us prove one of them explicitly. For example, $\mu(A'\ {\rm and}\ B)+\mu(A'\ {\rm and}\ B')=\langle C|(\mathbbmss{1}-M)\otimes M|C\rangle+\langle C|(\mathbbmss{1}-M)\otimes (\mathbbmss{1}-M)|C\rangle=\langle C|(\mathbbmss{1}-M)\otimes M+(\mathbbmss{1}-M)|C\rangle=\langle C|(\mathbbmss{1}-M)\otimes \mathbbmss{1}|C\rangle=\mu(A')$. 

Let us prove the other implication. Hence suppose that we have available data satisfying the classicality conditions (\ref{condbis01})--(\ref{condbis05}), let us prove that we can find a state $|C\rangle$ and an orthogonal projector $M$, such that second sector models these data. It is a straightforward verification that an entangled vector $|C\rangle$, such that $c_{ij}={1 \over 16}\sqrt{\mu(A\ {\rm and}\ B)}$ for $5 \le i \le 8$ and $5 \le j \le 8$, $c_{ij}={1 \over 16}\sqrt{\mu(A\ {\rm and}\ B')}$ for $5 \le i \le 8$ and $1 \le j \le 4$, $c_{ij}={1 \over 16}\sqrt{\mu(A'\ {\rm and}\ B)}$ for $1 \le i \le 4$ and $5 \le j \le 8$ and $c_{ij}={1 \over 16}\sqrt{\mu(A'\ {\rm and}\ B')}$ for $1 \le i \le 4$ and $1 \le j \le 4$, is a solution.
\qed

\bigskip
\noindent
It is interesting to notice that Theorem 4 implies that the tensor product Hilbert space model (second sector of Fock space) has exactly the same generality as the most general classical conditions for conjunction and negation. More specifically, given any data that satisfy the five classicality conditions of Theorem 3$'$, we can construct an entangled state such that in second sector `exactly' these classicality conditions are satisfied. Moreover, it clarifies that entangled states in our general Fock space modeling of data on conceptual conjunction and negation play a fundamental unexpected role in the combination of human concepts. The fact that classical logical rules are satisfied, in a probabilistic form, in second sector of Fock space provides an important confirmation to our two-sector quantum framework, as we will see in Section \ref{results}.

\subsection{A complete modeling in Fock space\label{twosector}}
In Section \ref{onesector} we considered the situation of first sector of Fock space, representing the starting concepts $A$ and $B$ by the state vectors $|A\rangle$ and $|B\rangle$, respectively, of a Hilbert space ${\cal H}$. Then, we introduced the state vectors $|A'\rangle$ and $|B'\rangle$, which represent the conceptual negations `not $A$' and `not $B$', respectively. Since first sector of Fock space describes `emergence', the state vectors $|A'\rangle$ and $|B'\rangle$ can be interpreted as representing the newly emergent concepts `not $A$' and `not $B$', respectively, compatibly with the core of the approach we developed in our quantum modeling of combinations of concepts. We also saw in Section \ref{onesector} that also the newly emergent concepts which is the conjunction `$A$ and $B$', or one of the other conjunction combinations, `$A$ and $B'$', `$A'$ and $B$', and `$A'$ and $B'$', are directly represented in this first sector of Fock space  by state vectors, more specifically by the superposition state vectors of the corresponding state vectors, namely ${1 \over \sqrt{2}}(|A\rangle+|B\rangle)$, ${1 \over \sqrt{2}}(|A\rangle+|B'\rangle)$, ${1 \over \sqrt{2}}(|A'\rangle+|B\rangle)$ and ${1 \over \sqrt{2}}(|A'\rangle+|B'\rangle)$, respectively.

Following \cite{a2009a} and \cite{s2014a,s2014b}, we should also take into account the logical aspects of conceptual conjunctions and negations in second sector of Fock space, mathematically formed by the tensor product of the Hilbert space ${\cal H}$ of first sector. In the approach we followed till now we represented the state of the concept $A$ and $B$ in second sector by the product vector $|A\rangle \otimes |B\rangle$ of this tensor product ${\cal H} \otimes {\cal H}$. However, we saw in Section \ref{entanglement}, this leads inevitably to the probability for the conjunction $\mu(A\ {\rm and}\ B)$ to be equal to the product $\mu(A)\mu(B)$. In classical probability theory this means that the probabilities are probabilistically independent. Now, quite obviously, since the concepts $A$ and $B$ are related by their meaning, these probabilities are not probabilistically independent. Suppose that $B$ were `not $A$', like in the situation of the borderline effect \cite{s2014a}, then obviously we would have an anti-correlation between $\mu(A)$ and $\mu(B)$. But, even in this not simple case, any meaning connection between $A$ and $B$ would give rise to probabilities that are not independent. On the other hand, we have seen in Section \ref{entanglement} that we can model any type of classical probabilistic dependence by introducing the proper entangled state for the concept representation of $A$ and of $B$ in second sector. This means that we should not in principle use $|A\rangle \otimes |B\rangle$ to represent the concepts in second sector, but a properly chosen entangled state.

Let us denote, following our analysis in Section \ref{entanglement}, such a general entangled state in $\compl^8$ by means of
\begin{eqnarray}
|C\rangle=\sum_{i,j=1}^{8}c_{ij}e^{\gamma_{ij}}|i\rangle \otimes |j\rangle
\end{eqnarray}
where $| i \rangle$ and $| j \rangle$ are the canonical base vectors of $\compl^8$.

The state vector representing the concept `$A$ and $B$' in its totality, hence its first sector part, describing emergent human thought, i.e. the formation of the new concept `$A$ and $B$', and its second sector part, describing quantum logical human thought, i.e. the conjunctive connective structure `$A$ and $B$', is then the following
\begin{equation} \label{fockstateAB}
\psi(A,B)=m_{AB}e^{i\theta}|C\rangle + {n_{AB}e^{i\rho} \over \sqrt{2}}(|A\rangle + |B\rangle)
\end{equation}
with $m_{AB}^2+n_{AB}^2=1$. It is indeed the direct sum of two vectors, one vector given by ${1 \over \sqrt{2}}(|A\rangle + |B\rangle)$ in first sector of Fock space, accounting for the emergent part of human thought with respect to the conjunction, and a second vector given by $|C\rangle$ in second sector of Fock space, accounting for the quantum logical part of human thought with respect to the conjunction.

We then get the following general expression for the membership weight of the conjunction
\begin{eqnarray}
\mu(A\ {\rm and}\ B)&=& \Big (m_{AB}e^{-i\theta}\langle C| + {n_{AB}e^{-i\rho}\over \sqrt{2}}(\langle A|+\langle B|) \Big )(M \otimes M \oplus M) \Big ( m_{AB}e^{i\theta}|C\rangle + {n_{AB}e^{i\rho}\over \sqrt{2}}(|A\rangle+|B\rangle) \Big ) \nonumber \\
&=&m_{AB}^2(\langle C|)M \otimes M|(|C\rangle)+{n_{AB}^2 \over 2}(\langle A|+\langle B|)M(|A\rangle+|B\rangle) \nonumber \\
&=&m_{AB}^2\sum_{i,j=5}^8c^2_{ij}+{n_{AB}^2 \over 2}(\langle A|M|A\rangle+\langle B|M|B\rangle+\langle A|M|B\rangle+\langle B|M|A\rangle) \nonumber \\ \label{fockspaceconjunctionAB}
&=&m_{AB}^2\sum_{i,j=5}^8c^2_{ij}+n_{AB}^2({1 \over 2}(\mu(A)+\mu(B))+\Re\langle A|M|B\rangle) \nonumber \\ \label{FockSpaceSolutionAB}
&=&m_{AB}^2\sum_{i,j=5}^8c^2_{ij}+n_{AB}^2({1 \over 2}(\mu(A)+\mu(B))+(a_5b_5+a_6b_6+a_7b_7+a_8b_8)\cos(\phi_B-\phi_A))
\end{eqnarray}
where we have used Equation (\ref{firstsectoreq15}) in the last line of Equation (\ref{FockSpaceSolutionAB}).

What is the procedure corresponding to emergent and quantum logical parts of human thought when we also take into account negations, i.e. when we consider the conjunctions `$A$ and $B'$', ` $A'$ and $B$' and `$A'$ and $B'$'? In first sector of Fock space, we already made it explicit in the foregoing section. As we saw above, the new emergent concepts `$A'$' and `$B'$' are described by the state vectors $|A'\rangle$ and $|B'\rangle$, and the respective conjunctions, i.e. their emergent aspects as a new concept, each time by means of the corresponding superposition state vector. This is the way the emergence of negation and conjunction are jointly modeled in first sector of Fock space -- new state vectors model the new emergent concepts due to negation, and the emergent conjunctions are modeled by the respective superpositions.

In second sector of Fock space, we however have a specific situation to solve. Namely, exactly as we did for the conjunction, we need to identify what is the quantum logical structure related with negation, independent of its provoking the emergence of a new concept, i.e. the negation of the original concept. In second sector of Fock space we indeed only express the quantum logical reasoning in human thought and not the emergent reasoning. For the conjunction `$A$ and $B$' we did this by means of the entangled state $|C\rangle$. Let us reflect about the negation, for example, with respect to the concept $B$. To make things clear let us introduce the following two expressions. We are in the experimental situation where the membership of an exemplar $x$, or the non-membership of this exemplar, is to be decided about, by a test subject participating in the experiment. The concept $B$ can be involved, and the concept $B'$ can be involved.

\medskip
\noindent
{\it Expression 1.} The considered exemplar $x$ is a member of the concept $B'$.

\medskip
\noindent
{\it Expression 2.} The considered exemplar $x$ is `not' a member of the concept $B$.

\medskip
\noindent
Our theoretic proposal is that:

(1) the first expression describes what happens in a human mind when emergent thought is dominant with respect to a concept $B$, its negation $B'$ and an exemplar $x$. Indeed, the focus is on `membership' of this exemplar $x$ with respect to the new emergent concept `$B'$';

(2) the second expression describes what happens in a human mind when quantum logical thought is dominant with respect to a concept $B$, its negation $B'$ and an exemplar $x$. Indeed, the focus is on `non-membership' of this exemplar $x$ with respect to the old existing concept $B$.

Expressions (1) and (2) are two structurally speaking subtle deeply different possibilities of reasoning related to a concept and its negation. 

Our third theoretic proposal is that: 

(3) human thought, when confronted with this situation, follows a dynamics described by a quantum superposition of the two modes (1) and (2).
 
We will see in the following that the mathematical structure of Fock space enables modeling this in an impecable way. 

Indeed, expression (1) will be modeled in first sector of our Fock space, and it is mathematically realised by making $M$ work on $|B'\rangle$. Expression (2) will instead be modeled in second sector of Fock space, and it is mathematically realised by making $\mathbbmss{1}\otimes(\mathbbmss{1}-M)$ work on $|C\rangle$. In the complete Fock space, direct sum of its first and second sectors, mathematically a superposition of the whole dynamics can be realised, by considering the superposition state which we already specified in Equation (\ref{fockstateAB}), and consider different structures of the projection operator on the whole of Fock space. More specifically, $(M\otimes M) \oplus M$ for `$A$ and $B$', $(M\otimes (\mathbbmss{1}-M)) \oplus M$ for `$A$ and $B'$', $((\mathbbmss{1}-M)\otimes M) \oplus M$ for `$A'$ and $B$', and $((\mathbbmss{1}-M)\otimes(\mathbbmss{1}-M)) \oplus M$ for `$A'$ and $B'$'. However, for each of the combinations the vector representing the combination in second sector of Fock space will be $|C\rangle$. So, no vector appears in second sector of Fock space, since the negation is expressed quantum logically here, hence by $M$ becoming $\mathbbmss{1}-M$. While in first sector of Fock space, the negation is expressed emergently, hence by $|A\rangle$ becoming $|A'\rangle$ and $|B\rangle$ becoming $|B'\rangle$, and $M$ remaining $M$, since the focus in this first sector of Fock space, with emergent reasoning of human thought, is always on `membership', while in second sector, with quantum logical reasoning, the focus of negation is on `non-membership', described by $\mathbbmss{1}-M$.   

The above conceptual analysis makes it possible for us to write the complete Fock space formulas for the other combinations. More specifically, if we represent the concept `$A \ {\rm and} \ B'$' by the unit vector
\begin{equation} \label{fockstateAB'}
\psi(A,B')=m_{AB'}e^{i\theta}|C\rangle + {n_{AB'}e^{i\rho} \over \sqrt{2}}(|A\rangle + |B'\rangle)
\end{equation}
with $m_{AB'}^2+n_{AB'}^2=1$, then
\begin{eqnarray}
\mu(A\ {\rm and}\ B')
&=& (m_{AB'}e^{-i\theta}\langle C| + {n_{AB'}e^{-i\rho}\over \sqrt{2}}(\langle A|+\langle B'|)) M \otimes (\mathbbmss{1}-M) \oplus M(m_{AB'}e^{i\theta}|C\rangle + {n_{AB'}e^{i\rho}\over \sqrt{2}}(|A\rangle+|B'\rangle)) \nonumber \\
&=&m_{AB'}^2(\langle C|) M \otimes (\mathbbmss{1}-M)|(|C\rangle)+{n_{AB'}^2 \over 2}(\langle A|+\langle B'|)M(|A\rangle+|B'\rangle) \nonumber \\
&=&m_{AB'}^2\sum_{i=5}^8\sum_{j=1}^4c^2_{ij}+{n_{AB'}^2 \over 2}(\langle A|M|A\rangle+\langle B'|M|B'\rangle+\langle A|M|B'\rangle+\langle B'|M|A\rangle) \nonumber \\ 
&=&m_{AB'}^2\sum_{i=5}^8\sum_{j=1}^4c^2_{ij}+n_{AB'}^2({1 \over 2}(\mu(A)+\mu(B'))+\Re\langle A|M|B'\rangle) \nonumber \\ \label{FockSpaceSolutionAB'}
&=&m_{AB'}^2\sum_{i=5}^8\sum_{j=1}^4c^2_{ij}+n_{AB'}^2({1 \over 2}(\mu(A)+\mu(B'))+(a_5b'_5+a_6b'_6+a_7b'_7+a_8b'_8)\cos(\phi_{B'}-\phi_A))
\end{eqnarray}
where we have used Equation (\ref{firstsectoreq16}) in the last line of Equation (\ref{FockSpaceSolutionAB'}). Analogously,
if we represent the concept `$A' \ {\rm and} \ B$' by the unit vector
\begin{equation} \label{fockstateA'B}
\psi(A',B)=m_{A'B}e^{i\theta}|C\rangle + {n_{A'B}e^{i\rho} \over \sqrt{2}}(|A'\rangle + |B\rangle)
\end{equation}
with $m_{A'B}^2+n_{A'B}^2=1$, then
\begin{eqnarray}
\mu(A'\ {\rm and}\ B)&=& (m_{A'B}e^{-i\theta}\langle C| + {n_{A'B}e^{-i\rho}\over \sqrt{2}}(\langle A'|+\langle B|))(\mathbbmss{1}-M) \otimes M \oplus M (m_{A'B}e^{i\theta}|C\rangle + {n_{A'B}e^{i\rho}\over \sqrt{2}}(|A'\rangle+|B\rangle)) \nonumber \\
&=&m_{A'B}^2(\langle C|) (\mathbbmss{1}-M) \otimes M|(|C\rangle)+{n_{A'B}^2 \over 2}(\langle A'|+\langle B|)M(|A'\rangle+|B\rangle) \nonumber \\
&=&m_{A'B}^2\sum_{i=1}^4\sum_{j=5}^8c^2_{ij}+{n_{A'B}^2 \over 2}(\langle A'|M|A'\rangle+\langle B|M|B\rangle+\langle A'|M|B\rangle+\langle B|M|A'\rangle) \nonumber \\ \label{fockspaceconjunctionA'B}
&=&m_{A'B}^2\sum_{i=1}^4\sum_{j=5}^8c^2_{ij}+n_{A'B}^2({1 \over 2}(\mu(A')+\mu(B))+\Re\langle A'|M|B\rangle) \nonumber \\ \label{FockSpaceSolutionA'B}
&=&m_{A'B}^2\sum_{i=1}^4\sum_{j=5}^8c^2_{ij}+n_{A'B}^2({1 \over 2}(\mu(A')+\mu(B))+(a'_5b_5+a'_6b_6+a'_7b_7+a'_8b_8)\cos(\phi_B-\phi_{A'}))
\end{eqnarray}
where we have used Equation (\ref{firstsectoreq17}) in the last line of Equation (\ref{FockSpaceSolutionA'B}). Finally,
if we represent the concept `$A' \ {\rm and} \ B'$' by the unit vector
\begin{equation} \label{fockstateA'B'}
\psi(A',B')=m_{A'B'}e^{i\theta}|C\rangle + {n_{A'B'}e^{i\rho} \over \sqrt{2}}(|A'\rangle + |B'\rangle)
\end{equation}
with $m_{A'B'}^2+n_{A'B'}^2=1$, then
\begin{eqnarray}
&&\mu(A'\ {\rm and}\ B') \nonumber \\
&=& (m_{A'B'}e^{-i\theta}\langle C| + {n{A'B'}e^{-i\rho}\over \sqrt{2}}(\langle A'|+\langle B'|)) (\mathbbmss{1}-M) \otimes (\mathbbmss{1}-M) \oplus M(m_{A'B'}e^{i\theta}|C\rangle + {n_{A'B'}e^{i\rho}\over \sqrt{2}}(|A'\rangle+|B'\rangle)) \nonumber \\
&=&m_{A'B'}^2(\langle C|) (\mathbbmss{1}-M) \otimes (\mathbbmss{1}-M)|(|C\rangle)+{n_{A'B'}^2 \over 2}(\langle A'|+\langle B'|)M(|A'\rangle+|B'\rangle) \nonumber \\
&=&m_{A'B'}^2\sum_{i,j=1}^4c^2_{ij}+{n_{A'B'}^2 \over 2}(\langle A'|M|A'\rangle+\langle B'|M|B'\rangle+\langle A'|M|B'\rangle+\langle B'|M|A'\rangle) \nonumber \\ 
&=&m_{A'B'}^2\sum_{i,j=1}^4c^2_{ij}+n_{A'B'}^2({1 \over 2}(\mu(A)+\mu(B'))+\Re\langle A'|M|B'\rangle) \nonumber \\ \label{FockSpaceSolutionA'B'}
&=&m_{A'B'}^2\sum_{i,j=1}^4c^2_{ij}+n_{A'B'}^2({1 \over 2}(\mu(A')+\mu(B'))+(a'_5b'_5+a'_6b'_6+a'_7b'_7+a'_8b'_8)\cos(\phi_{B'}-\phi_{A'}))
\end{eqnarray}
where we have used Equation (\ref{firstsectoreq18}) in the last line of Equation (\ref{FockSpaceSolutionA'B'}).

Equations (\ref{FockSpaceSolutionAB}), (\ref{FockSpaceSolutionAB'}), (\ref{FockSpaceSolutionA'B}) and (\ref{FockSpaceSolutionA'B'}) contain the probabilistic expressions for simultaneously representing experimental data on conjunctions and negations of two concepts in a quantum-theoretic framework. These equations express the membership weights of the conjunctions `$A$ and $B$', `$A$ and $B'$', `$A'$ and $B$' and `$A'$ and $B'$' in terms of the memership weights of $A$, $B$, $A'$ and $B'$, for suitable values of the following modeling parameters:\footnote{We remind that $n_{XY}^2=1-m_{XY}^2$, $X=A,A',Y=B,B'$. In addition, only the sums $\sum_{i,j=5}^{8}c_{ij}^2$, $\sum_{i=5}^{8}\sum_{j=1}^{4}c_{ij}^2$, $\sum_{i=1}^{4}\sum_{j=5}^{8}c_{ij}^2$ and $\sum_{i,j=5}^{8}c_{ij}^2$ appear in Equations (\ref{FockSpaceSolutionAB}), (\ref{FockSpaceSolutionAB'}), (\ref{FockSpaceSolutionA'B}) and (\ref{FockSpaceSolutionA'B'}), respectively.}

(i) the angles $\phi_{B}-\phi_{A}$, $\phi_{B'}-\phi_{A}$, $\phi_{B}-\phi_{A'}$ and $\phi_{B'}-\phi_{A'}$, 

(ii) the pairs of convex coefficients $(m_{AB},n_{AB})$, $(m_{AB'},n_{AB'})$, $(m_{A'B},n_{A'B})$ and $(m_{A'B'},n_{A'B'})$, 

(iii) the normalized coefficients $c_{11}^2$, \ldots, $c_{88}^2$. 

As we can see, our two-sector Fock space framework is able to cope with conceptual negation in a very natural way. In fact, the latter negation is modeled by using the general assumption that emergent aspects of a concept are represented in first sector of Fock space, while logical aspects of a concept are represented in second sector. This will be made explicit in Section \ref{results}.  
It is however important to stress that, for a given experiment $e_{XY}$, with $X=A,A'$, $Y=B,B'$ described in Section \ref{experiment}, there is no guarantee that sets of these paramters can be found such that Equations (\ref{FockSpaceSolutionAB})--(\ref{FockSpaceSolutionA'B'}) are simultaneously satisfied. For this reason, we will provide in the next section the conditions that should be satisfied by the experimental data $\mu(A)$, $\mu(B)$, \ldots, $\mu(A' \ {\rm and} \ B)$, $\mu(A' \ {\rm and} \ B)$ such that these sets exist.

\subsection{The modeling procedure\label{solconditions}}
Let us consider the data collected in the experiments $e_{XY}$, $X=A,A'$, $Y=B,B'$, of Section \ref{experiment}. These data are explicitly reported in Tables 1--4, Appendix \ref{tables}. Since the existence of solutions for Equations (\ref{FockSpaceSolutionAB})--(\ref{FockSpaceSolutionA'B'}) depends, for a given experiment $e_{XY}$, on the expemplar $x$ and the pair $(A,B)$ of concepts that are considered, we explicitly report such dependence for all the relevant variables that are considered in this section. Hence, for each considered exemplar $x$, we collected the eight membership weights $\mu_{x}(A)$, $\mu_{x}(B)$, $\mu_{x}(A')$, $\mu_{x}(B')$, $\mu_{x}(A \ {\rm and} \ B)$, $\mu_{x}(A\ {\rm and} \ B')$, $\mu_{x}(A' \ {\rm and} \ B)$, and $\mu_{x}(A' \ {\rm and} \ B')$.

The analysis we made in the foregoing sections makes it possible for us to propose a general modeling procedure. For what concerns solutions that can be found on first sector alone, we determined the intervals of solutions as explained in (\ref{firstintsolAB}), (\ref{firstintsolAB'}), (\ref{firstintsolA'B}) and (\ref{firstintsolA'B'}). We can now easily determine the general intervals of solutions, including the extra solutions made possible by second sector. Therefore we need to consider the following quantities $\sum_{i,j=5}^8c^2_{ij}$, $\sum_{i=5}^8\sum_{j=1}^4c^2_{ij}$,$\sum_{i=1}^4\sum_{j=5}^8c^2_{ij}$ and $\sum_{i=5}^8\sum_{j=1}^4c^2_{ij}$, respectively for the combinations `$A$ and $B$', `$A$ and not $B$', `not $A$ and $B$' and `not $A$ and not $B$'. To be able to express the intervals of Fock space solutions, we introduce the following quantities.
\begin{eqnarray}
s(A,B,x)&=&\min(\sum_{i,j=5}^8c^2_{ij}, i(A,B)_1) \\
t(A,B,x)&=&\max(\sum_{i,j=5}^8c^2_{ij}, i(A,B)_2)
\end{eqnarray}
Then, the interval
\begin{equation}\label{intsolAB}
U_{sol}(AB,x)=[s(A,B,x),t(A,B,x)]
\end{equation}
is the solution interval for the general Fock space model. Hence, in case the experimental value $\mu(A\ {\rm and}\ B)$ is contained in this interval, a solution exists. As a second step we can then see whether a solution in first sector alone exists, which consists of veryfying whether the experimental value $\mu(A\ {\rm and}\ B)$ is contained in $I_{AB}$. Suppose that the anwer is `yes', then we can caculate the angle $\phi_B-\phi_A$ that gives rise to this solution in first sector. This angle is then an indication of which angle to choose for the general Fock space solution. Usually different choices are possible. If there is no solution in first sector, we anyhow can choose an angle $\phi_B-\phi_A$, such that a choice of this angle $\phi_B-\phi_A$, and a choice of $m_{AB}$ and $n_{AB}$ gives a solution. The possible values of the angle and the coefficients $m_{AB}$ and $n_{AB}$ are calculated by solving the general Equation (\ref{FockSpaceSolutionAB}).

We can analyse the other combinations in an equivalent way. Let us start with the combination `$A$ and not $B$'. We have:
\begin{eqnarray}
s(A,B',x)&=&\min(\sum_{i=5}^8\sum_{j=1}^4c^2_{ij}, i(A,B')_1) \\
t(A,B',x)&=&\max(\sum_{i=5}^8\sum_{j=1}^4c^2_{ij},i(A,B')_2)
\end{eqnarray}
Then, the interval
\begin{equation}\label{intsolAB'}
U_{sol}(AB',x)=[s(A,B',x),t(A,B',x)]
\end{equation}
is the solution interval for the general Fock space model. The equation to be used to calculate the angle $\phi_{B'}-\phi_A$, and the coefficients $m_{AB'}$ and $n_{AB'}$ is Equation (\ref{FockSpaceSolutionAB'}).

For the combination `$A'$ and $B$', we have:
\begin{eqnarray}
s(A',B,x)&=&\min(\sum_{i=1}^4\sum_{j=5}^8c^2_{ij}, i(A',B)_1) \\
t(A',B,x)&=&\max(\sum_{i=1}^4\sum_{j=5}^8c^2_{ij}, i(A',B)_2)
\end{eqnarray}
Then, the interval
\begin{equation}\label{intsolA'B}
U_{sol}(A'B,x)=[s(A',B,x),t(A',B,x)]
\end{equation}
is the solution interval for the general Fock space model. The equation to be used to calculate the angle $\phi_B-\phi_{A'}$, and the coefficients $m_{A'B}$ and $n_{A'B}$ is Equation (\ref{FockSpaceSolutionA'B}).

Finally, for the combination `not $A$ and not $B$', we have:
\begin{eqnarray}
s(A',B',x)&=&\min(\sum_{i=5}^8\sum_{j=1}^4c^2_{ij}, i(A',B')_1) \\
t(A,B',x)&=&\max(\sum_{i=5}^8\sum_{j=1}^4c^2_{ij}, i(A',B')_2)
\end{eqnarray}
Then, the interval
\begin{equation}\label{intsolA'B'}
U_{sol}(A'B',x)=[s(A',B',x),t(A',B',x)]
\end{equation}
is the solution interval for the general Fock space model. The equation to be used to calculate the angle $\phi_{B'}-\phi_{A'}$, and the coefficients $m_{A'B'}$ and $n_{A'B'}$ is Equation (\ref{FockSpaceSolutionA'B'}).

The conclusion we draw from the analysis above is that finding solutions for a given set of experimental data in our quantum-theoretic modeling it is highly non-obvious, which makes the results in the next section even more significant.

\section{Representation of experimental data in Fock space\label{model}}
The data in our experiments on conjunctions and negations of two concepts are presented in Appendix \ref{tables}, Tables 1--4. Most of these data are compatible with the intervals in Equations (\ref{intsolAB}), (\ref{intsolAB'}), (\ref{intsolA'B}) and (\ref{intsolA'B'}). Hence, almost all our data can be successfully modeled by using the quantum probabilistic equations in (\ref{FockSpaceSolutionAB}), (\ref{FockSpaceSolutionAB'}), (\ref{FockSpaceSolutionA'B}) and (\ref{FockSpaceSolutionA'B'}). Let us consider some interesting cases, distinguishing them by: (i) situations with double overextension, (ii) situations with complete overextension, (iii) situations requiring both sectors of Fock space and/or entanglement, (iv) partially classical situations. Complete modeling is presented in the Supplementary Material attached to this article.

\medskip

(i) Let us start with exemplars that are double overextended.

{\it Olive}, with respect to ({\it Fruits}, {\it Vegetables}) (double overextension with respect to {\it Fruits And Vegetables}). {\it Olive} scored $\mu(A)=0.53$ with respect to {\it Fruits}, $\mu(B)=0.63$ with respect to {\it Vegetables}, $\mu(A')=0.47$ with respect to {\it Not Fruits}, $\mu(B')=0.44$ with respect to {\it Not Vegetables}, $\mu(A \ {\rm and} \ B)=0.65$ with respect to {\it Fruits And Vegetables}, $\mu(A \ {\rm and} \ B')=0.34$ with respect to {\it Fruits And Not Vegetables}, $\mu(A' \ {\rm and} \ B)=0.51$ with respect to {\it Not Fruits And Vegetables}, and $\mu(A' \ {\rm and} \ B')=0.36$ with respect to {\it Not Fruits And Not Vegetables}. If one first looks for a representation of {\it Olive} in the Hilbert space $\mathbb{C}^8$, then the concepts {\it Fruits} and {\it Vegetables} are represented by  	the unit vectors
$|A\rangle=e^{i \phi_{A}}(-0.02,-0.47,0.5,-0.02,-0.07,-0.31,-0.18,-0.63)$
and
$|B\rangle=e^{i \phi_{B}}(0.04,0.02,-0.6,0.03,-0.26,0.35,-0.39,\- -0.53)$, respectively,
and their negations {\it Not Fruits} and {\it Not Vegetables} by the unit vectors
$|A'\rangle= e^{i \phi_{A'}} (0.06,-0.47, \- -0.55,0.03,-0.02,-0.64,-0.06,0.25)$,
and
$|B'\rangle=e^{i \phi_{B'}}(-0.03,0.75,-0.01,-0.01,-0.08,-0.6,-0.18,-0.19)$, respectively.

The interference angles $\phi_{AB}=\phi_{B}-\phi_{A}=57.31^{\circ}$, $\phi_{AB'}=\phi_{B'}-\phi_{A}=95.32^{\circ}$, $\phi_{A'B}=\phi_{B}-\phi_{A'}=103.43^{\circ}$ and $\phi_{A'B'}=\phi_{B'}-\phi_{A'}=85.56^{\circ}$ complete the Hilbert space representation in $\mathbb{C}^8$. A complete modeling in the Fock space $\mathbb{C}^8 \oplus (\mathbb{C}^8\otimes \mathbb{C}^8)$ satisfying Equations (\ref{FockSpaceSolutionAB}), (\ref{FockSpaceSolutionAB'}), (\ref{FockSpaceSolutionA'B}) and (\ref{FockSpaceSolutionA'B'}) is given by an entangled state characterized by $\sum_{i,j=5}^{8}c_{ij}^2=0.44^2$, $\sum_{i=5}^{8}\sum_{i=1}^{4}c_{ij}^{2}=0.58^2$, $\sum_{i=1}^{4}\sum_{j=5}^{8}c_{ij}^2=0.66^2$ and $\sum_{i,j=1}^{4}c_{ij}^{2}=0.18^2$, and convex weights 
$m_{AB}=0.42$,
$n_{AB}=0.91$, 
$m_{AB'}=0.1$, 
$n_{AB'}=0$, 
$m_{A'B}=0.78$,  $n_{A'B}=0.63$, $m_{A'B'}=0.52$, and $n_{A'B'}=0.86$.
 	
{\it Prize Bull}, with respect to ({\it Pets}, {\it Farmyard Animals}) (double overextension with respect to {\it Pets And Not Farmyard Animals}). {\it Prize Bull} scored $\mu(A)=0.13$ with respect to {\it Pets}, $\mu(B)=0.76$ with respect to {\it Farmyard Animals}, $\mu(A')=0.88$ with respect to {\it Not Pets}, $\mu(B')=0.26$ with respect to {\it Not Farmyard Animals}, $\mu(A \ {\rm and} \ B)=0.43$ with respect to {\it Pets And Farmyard Animals}, $\mu(A \ {\rm and} \ B')=0.28$ with respect to {\it Pets And Not Farmyard Animals}, $\mu(A' \ {\rm and} \ B)=0.83$ with respect to {\it Not Pets And Farmyard Animals}, and $\mu(A' \ {\rm and} \ B')=0.34$ with respect to {\it Not Pets And Not Farmyard Animals}. If one first looks for a representation of {\it Prize Bull} in the Hilbert space $\mathbb{C}^8$, then the concepts {\it Pets} and {\it Farmyard Animals}, and their negations {\it Not Pets} and {\it Not Farmyard Animals} are respectively represented by the unit vectors  
 $|A\rangle=e^{i \phi_{A}}(0.07,-0.39,-0.84,0.03,-0.06,-0.35,0.04,-0.01)$
 and
$|B\rangle=e^{i \phi_{B}}(0.03,	0.21,	-0.44,	0.01,	0.01,	0.81,	-0.2,	-0.25)$,
and
$|A'\rangle=e^{i \phi_{A'}}(0.01,0.29,	-0.19,	0,	0.11,	0.06,	-0.2,	0.91)$
and
$|B'\rangle=e^{i \phi_{B'}}(0.01,0.84,	-0.19,	0,	-0.17,	-0.41,	-0.01,	-0.26)$.

The interference angles $\phi_{AB}=\phi_{B}-\phi_{A}=105.71^{\circ}$, $\phi_{AB'}=\phi_{B'}-\phi_{A}=40.23^{\circ}$, $\phi_{A'B}=\phi_{B}-\phi_{A'}=111.25^{\circ}$ and $\phi_{A'B'}=\phi_{B'}-\phi_{A'}=52.51^{\circ}$ complete the Hilbert space representation in $\mathbb{C}^8$. A complete modeling in the Fock space $\mathbb{C}^8 \oplus (\mathbb{C}^8\otimes \mathbb{C}^8)$ satisfying Equations (\ref{FockSpaceSolutionAB}), (\ref{FockSpaceSolutionAB'}), (\ref{FockSpaceSolutionA'B}) and (\ref{FockSpaceSolutionA'B'}) is given by an entangled state characterized by $\sum_{i,j=5}^{8}c_{ij}^2=0.24^2$, $\sum_{i=5}^{8}\sum_{i=1}^{4}c_{ij}^{2}=0.27^2$, $\sum_{i=1}^{4}\sum_{j=5}^{8}c_{ij}^2=0.84^2$ and $\sum_{i,j=1}^{4}c_{ij}^{2}=0.41^2$, and convex weights 
$m_{AB}=0.46$,
$n_{AB}=0.89$, 
$m_{AB'}=0.41$, 
$n_{AB'}=0.91$, 
$m_{A'B}=0.54$,  $n_{A'B}=0.84$, $m_{A'B'}=0.52$, and $n_{A'B'}=0.85$.

{\it Door Bell}, with respect to ({\it Home Furnishing}, {\it Furniture}) (double overextension with respect to {\it Not Home Furnishing And Furniture}). {\it Door Bell} scored $\mu(A)=0.75$ with respect to {\it Home Furnishing}, $\mu(B)=0.33$ with respect to {\it Furniture}, $\mu(A')=0.32$ with respect to {\it Not Home Furnishing}, $\mu(B')=0.79$ with respect to {\it Not Furniture}, $\mu(A \ {\rm and} \ B)=0.5$ with respect to {\it Home Furnnishing And Furniture}, $\mu(A \ {\rm and} \ B')=0.64$ with respect to {\it Home Furnishing And Not Furniture}, $\mu(A' \ {\rm and} \ B)=0.34$ with respect to {\it Not Home Furnishing And Furniture}, and $\mu(A' \ {\rm and} \ B')=0.51$ with respect to {\it Not Home Furnishing And Not Furniture}. If one first looks for a representation of {\it Door Bell} in the Hilbert space $\mathbb{C}^8$, then the concepts {\it Home Furnishing} and {\it Furniture}, and their negations {\it Not Home Furnishing} and {\it Not Furniture} are respectively represented by the unit vectors  
$|A\rangle=e^{i \phi_{A}}(0,0.33,0.37,-0.05,0.04,-0.29,0,0.81)$
	and
$|B\rangle=e^{i \phi_{B}}(-0.14,	0.77,0.17,	-0.16,	0.24, -0.19,0.07,	-0.48)$,
and
$|A'\rangle=e^{i \phi_{A'}}(0.21,-0.43,0.66,0.13,0.22,-0.39,$ $0.22,-0.27)$
	and
$|B'\rangle=e^{i \phi_{B'}}(-0.08,-0.03,-0.45,-0.02,-0.17,-0.52,0.7,0.04)$.
									
The interference angles $\phi_{AB}=\phi_{B}-\phi_{A}=102.81^{\circ}$, $\phi_{AB'}=\phi_{B'}-\phi_{A}=117.67^{\circ}$, $\phi_{A'B}=\phi_{B}-\phi_{A'}=67.37^{\circ}$ and $\phi_{A'B'}=\phi_{B'}-\phi_{A'}=77.65^{\circ}$ complete the Hilbert space representation in $\mathbb{C}^8$. A complete modeling in the Fock space $\mathbb{C}^8 \oplus (\mathbb{C}^8\otimes \mathbb{C}^8)$ satisfying Equations (\ref{FockSpaceSolutionAB}), (\ref{FockSpaceSolutionAB'}), (\ref{FockSpaceSolutionA'B}) and (\ref{FockSpaceSolutionA'B'}) is given by an entangled state characterized by $\sum_{i,j=5}^{8}c_{ij}^2=0.35^2$, $\sum_{i=5}^{8}\sum_{i=1}^{4}c_{ij}^{2}=0.79^2$, $\sum_{i=1}^{4}\sum_{j=5}^{8}c_{ij}^2=0.46^2$ and $\sum_{i,j=1}^{4}c_{ij}^{2}=0.2^2$, and convex weights 
$m_{AB}=0.48$,
$n_{AB}=0.88$, 
$m_{AB'}=0.91$, 
$n_{AB'}=0.41$, 
$m_{A'B}=0.65$,  $n_{A'B}=0.76$, $m_{A'B'}=0.43$, and $n_{A'B'}=0.9$.

\medskip
 
(ii) Let us now come to the exemplars that present complete overextension, that is, exemplars that are overextended in all experiments.
 
{\it Goldfish}, with respect to ({\it Pets}, {\it Farmyard Animals}) (big overextension in all experiments, but also double overextension with respect to {\it Not Pets And Farmyard Animals}). {\it Goldfish} scored $\mu(A)=0.93$ with respect to {\it Pets}, $\mu(B)=0.17$ with respect to {\it Farmyard Animals}, $\mu(A')=0.12$ with respect to {\it Not Pets}, $\mu(B')=0.81$ with respect to {\it Not Farmyard Animals}, $\mu(A \ {\rm and} \ B)=0.43$ with respect to {\it Pets And Farmyard Animals}, $\mu(A \ {\rm and} \ B')=0.91$ with respect to {\it Pets And Not Farmyard Animals}, $\mu(A' \ {\rm and} \ B)=0.18$ with respect to {\it Not Pets And Farmyard Animals}, and $\mu(A' \ {\rm and} \ B')=0.43$ with respect to {\it Not Pets And Not Farmyard Animals}. If one first looks for a representation of {\it Goldfish} in the Hilbert space $\mathbb{C}^8$, then the concepts {\it Pets} and {\it Farmyard Animals}, and their negations {\it Not Pets} and {\it Not Farmyard Animals} are respectively represented by the unit vectors  
$|A\rangle=e^{i \phi_{A}}(-0.05,0.16,-0.21,-0.01,-0.71,0.22,0.33,0.51)$
and
$|B\rangle=e^{i \phi_{B}}(-0.24,0.26,-0.84,-0.07,0.38,-0.11,-0.01,0.12)$,
and
$|A'\rangle=e^{i \phi_{A'}}(0.18,0.85,0.35,0.09,0.2,-0.12,$ $-0.03,0.25)$
and
$|B'\rangle=e^{i \phi_{B'}}(0.01,-0.41,0.14,-0.01,0.27,-0.32,-0.13,0.79)$.

The interference angles $\phi_{AB}=\phi_{B}-\phi_{A}=78.9^{\circ}$, $\phi_{AB'}=\phi_{B'}-\phi_{A}=43.15^{\circ}$, $\phi_{A'B}=\phi_{B}-\phi_{A'}=54.74^{\circ}$ and $\phi_{A'B'}=\phi_{B'}-\phi_{A'}=77.94^{\circ}$ complete the Hilbert space representation in $\mathbb{C}^8$. A complete modeling in the Fock space $\mathbb{C}^8 \oplus (\mathbb{C}^8\otimes \mathbb{C}^8)$ satisfying Equations (\ref{FockSpaceSolutionAB}), (\ref{FockSpaceSolutionAB'}), (\ref{FockSpaceSolutionA'B}) and (\ref{FockSpaceSolutionA'B'}) is given by an entangled state characterized by $\sum_{i,j=5}^{8}c_{ij}^2=0.35^2$, $\sum_{i=5}^{8}\sum_{i=1}^{4}c_{ij}^{2}=0.9^2$, $\sum_{i=1}^{4}\sum_{j=5}^{8}c_{ij}^2=0.22^2$ and $\sum_{i,j=1}^{4}c_{ij}^{2}=0.17^2$, and convex weights 
$m_{AB}=0.45$,
$n_{AB}=0.89$, 
$m_{AB'}=0.45$, 
$n_{AB'}=0.9$, 
$m_{A'B}=0.48$,  $n_{A'B}=0.88$, $m_{A'B'}=0.45$, and $n_{A'B'}=0.89$. 
 
{\it Parsley}, with respect to ({\it Spices}, {\it Herbs}) (overextension in all experiments). {\it Parsley} scored $\mu(A)=0.54$ with respect to {\it Spices}, $\mu(B)=0.9$ with respect to {\it Herbs}, $\mu(A')=0.54$ with respect to {\it Not Spices}, $\mu(B')=0.09$ with respect to {\it Not Herbs}, $\mu(A \ {\rm and} \ B)=0.68$ with respect to {\it Spices And Herbs}, $\mu(A \ {\rm and} \ B')=0.26$ with respect to {\it Spices And Not Herbs}, $\mu(A' \ {\rm and} \ B)=0.73$ with respect to {\it Not Spices And Herbs}, and $\mu(A' \ {\rm and} \ B')=0.18$ with respect to {\it Not Spices And Not Herbs}. If one first looks for a representation of {\it Parsley} in the Hilbert space $\mathbb{C}^8$, then the concepts {\it Spices} and {\it Herbs}, and their negations {\it Not Spices} and {\it Not Herbs} are respectively represented by the unit vectors   
$|A\rangle=e^{i \phi_{A}}(0,0.25,-0.63,-0.02,-0.02,0.5,-0.06,0.54)$
	and
$|B\rangle=e^{i \phi_{B}}(0,0.02,-0.32,-0.01,0.09,-0.84,-0.23,0.37)$,
	and
$|A'\rangle=e^{i \phi_{A'}}(0,0.17,0.66,0.02,-0.17,0.01,$ $0.14,0.7)$
$|B'\rangle=e^{i \phi_{B'}}(0,-0.95,-0.06,-0.01,-0.04,0.11,0.02,0.27)$.
							
The interference angles $\phi_{AB}=\phi_{B}-\phi_{A}=97.66^{\circ}$, $\phi_{AB'}=\phi_{B'}-\phi_{A}=84.49^{\circ}$, $\phi_{A'B}=\phi_{B}-\phi_{A'}=68.25^{\circ}$ and $\phi_{A'B'}=\phi_{B'}-\phi_{A'}=113.49^{\circ}$ complete the Hilbert space representation in $\mathbb{C}^8$. A complete modeling in the Fock space $\mathbb{C}^8 \oplus (\mathbb{C}^8\otimes \mathbb{C}^8)$ satisfying Equations (\ref{FockSpaceSolutionAB}), (\ref{FockSpaceSolutionAB'}), (\ref{FockSpaceSolutionA'B}) and (\ref{FockSpaceSolutionA'B'}) is given by an entangled state characterized by $\sum_{i,j=5}^{8}c_{ij}^2=0.66^2$, $\sum_{i=5}^{8}\sum_{i=1}^{4}c_{ij}^{2}=0.32^2$, $\sum_{i=1}^{4}\sum_{j=5}^{8}c_{ij}^2=0.68^2$ and $\sum_{i,j=1}^{4}c_{ij}^{2}=0$, and convex weights 
$m_{AB}=0.48$,
$n_{AB}=0.88$, 
$m_{AB'}=0.55$, 
$n_{AB'}=0.84$, 
$m_{A'B}=0.46$,  $n_{A'B}=0.89$, $m_{A'B'}=0.5$, and $n_{A'B'}=0.87$.

\medskip

(iii) Let us then illustrate some relevant examplers that either cannot be modeled in a pure Hilbert space framework, or cannot be represented by product states in second sector of Fock space.

{\it Raisin}, with respect to ({\it Fruits}, {\it Vegetables}). {\it Raisin} scored $\mu(A)=0.88$ with respect to {\it Fruits}, $\mu(B)=0.27$ with respect to {\it Vegetables}, $\mu(A')=0.13$ with respect to {\it Not Fruits}, $\mu(B')=0.76$ with respect to {\it Not Vegetables}, $\mu(A \ {\rm and} \ B)=0.53$ with respect to {\it Fruits And Vegetables}, $\mu(A \ {\rm and} \ B')=0.75$ with respect to {\it Fruits And Not Vegetables}, $\mu(A' \ {\rm and} \ B)=0.25$ with respect to {\it Not Fruits And Vegetables}, and $\mu(A' \ {\rm and} \ B')=0.34$ with respect to {\it Not FruitsAnd Not Vegetables}. If one first looks for a representation of {\it Raisin} in the Hilbert space $\mathbb{C}^8$, then the concepts {\it Fruits} and {\it Vegetables}, and their negations {\it Not Fruits} and {\it Not Vegetables} are respectively represented by the unit vectors  	
$|A\rangle=e^{i \phi_{A}}(0.05,-0.01,0.34,0.01,-0.1,-0.51,0.23,-0.75)$
	and
$|B\rangle=e^{i \phi_{B}}	(-0.41,-0.15,-0.73,-0.1,-0.38,-0.17,$ $-0.19,-0.25)$,
	and
$|A'\rangle=e^{i \phi_{A'}}(0.56,-0.73,-0.09,0.1,-0.08,-0.28,-0.15,0.16)$
	and
$|B'\rangle=e^{i \phi_{B'}}(0.07,0.46,$ $-0.14,0.04,0.13,-0.76,-0.11,0.4)$.

However, a complete representation satisfying Equations (\ref{FockSpaceSolutionAB}), (\ref{FockSpaceSolutionAB'}), (\ref{FockSpaceSolutionA'B}) and (\ref{FockSpaceSolutionA'B'}) can only be worked out in the Fock space $\mathbb{C}^8 \oplus (\mathbb{C}^8\otimes \mathbb{C}^8)$. This occurs for interference angles $\phi_{AB}=\phi_{B}-\phi_{A}=80.79^{\circ}$, $\phi_{AB'}=\phi_{B'}-\phi_{A}=160^{\circ}$, $\phi_{A'B}=\phi_{B}-\phi_{A'}=18.15^{\circ}$ and $\phi_{A'B'}=\phi_{B'}-\phi_{A'}=92.88^{\circ}$, and for an entangled state characterized by $\sum_{i,j=5}^{8}c_{ij}^2=0.41^2$, $\sum_{i=5}^{8}\sum_{i=1}^{4}c_{ij}^{2}=0.85^2$, $\sum_{i=1}^{4}\sum_{j=5}^{8}c_{ij}^2=0.32^2$ and $\sum_{i,j=1}^{4}c_{ij}^{2}=0.13^2$, and convex weights 
$m_{AB}=0.45$,
$n_{AB}=0.89$, 
$m_{AB'}=0.65$, 
$n_{AB'}=0.76$, 
$m_{A'B}=0.26$,  $n_{A'B}=0.97$, $m_{A'B'}=0.48$, and $n_{A'B'}=0.88$.							

{\it Fox}, with respect to ({\it Pets}, {\it Farmyard Animals}). {\it Fox} scored $\mu(A)=0.13$ with respect to {\it Pets}, $\mu(B)=0.3$ with respect to {\it Farmyard Animals}, $\mu(A')=0.86$ with respect to {\it Not Pets}, $\mu(B')=0.68$ with respect to {\it Not Farmyard Animals}, $\mu(A \ {\rm and} \ B)=0.18$ with respect to {\it Pets And Farmyard Animals}, $\mu(A \ {\rm and} \ B')=0.29$ with respect to {\it Pets And Not Farmyard Animals}, $\mu(A' \ {\rm and} \ B)=0.46$ with respect to {\it Not Pets And Farmyard Animals}, and $\mu(A' \ {\rm and} \ B')=0.59$ with respect to {\it Not Pets And Not Farmyard Animals}. If one first looks for a representation of {\it Fox} in the Hilbert space $\mathbb{C}^8$, then the concepts {\it Pets} and {\it Farmyard Animals}, and their negations {\it Not Pets} and {\it Not Farmyard Animals} are respectively represented by the unit vectors  
$|A\rangle=e^{i \phi_{A}} (-0.07,-0.84,-0.39,-0.03,-0.02,-0.31,0.02,0.19)$
and
$|B\rangle=e^{i \phi_{B}}(-0.01,0.17,-0.82,0.01,-0.01,0.28,-0.01,-0.47)$,
and
$|A'\rangle=e^{i \phi_{A'}}(-0.05,0.19,-0.31,-0.02,0.12,0.39,$ $-0.02,0.83)$
and
$|B'\rangle=e^{i \phi_{B'}}(-0.14,0.47,-0.26,-0.08,-0.08,-0.8,0.04,0.17)$.
 
However, a complete representation satisfying Equations (\ref{FockSpaceSolutionAB}), (\ref{FockSpaceSolutionAB'}), (\ref{FockSpaceSolutionA'B}) and (\ref{FockSpaceSolutionA'B'}) can only be worked out in the Fock space $\mathbb{C}^8 \oplus (\mathbb{C}^8\otimes \mathbb{C}^8)$. This occurs for interference angles $\phi_{AB}=\phi_{B}-\phi_{A}=96.58^{\circ}$, $\phi_{AB'}=\phi_{B'}-\phi_{A}=95.05^{\circ}$, $\phi_{A'B}=\phi_{B}-\phi_{A'}=85.68^{\circ}$ and $\phi_{A'B'}=\phi_{B'}-\phi_{A'}=-20^{\circ}$, and for an entangled state characterized by $\sum_{i,j=5}^{8}c_{ij}^2=0.05^2$, $\sum_{i=5}^{8}\sum_{i=1}^{4}c_{ij}^{2}=0.36^2$, $\sum_{i=1}^{4}\sum_{j=5}^{8}c_{ij}^2=0.55^2$ and $\sum_{i,j=1}^{4}c_{ij}^{2}=0.76^2$, and convex weights 
$m_{AB}=0.51$,
$n_{AB}=0.86$, 
$m_{AB'}=0.61$, 
$n_{AB'}=0.79$, 
$m_{A'B}=0.61$,  $n_{A'B}=0.79$, $m_{A'B'}=0.66$, and $n_{A'B'}=0.75$.		

\medskip

(iv) Let us finally describe the quantum-theoretic representation of an exemplar that does not present overextension in any conjunction, but still does not admit a representation in a classical Kolmogorovian probability framework.

{\it Window Seat}, with respect to ({\it Home Furnishing}, {\it Furniture}). {\it Window Seat} scored $\mu(A)=0.5$ with respect to {\it Home Furnishing}, $\mu(B)=0.48$ with respect to {\it Furniture}, $\mu(A')=0.47$ with respect to {\it Not Home Furnishing}, $\mu(B')=0.55$ with respect to {\it Not Furniture}, $\mu(A \ {\rm and} \ B)=0.45$ with respect to {\it Home Furnnishing And Furniture}, $\mu(A \ {\rm and} \ B')=0.49$ with respect to {\it Home Furnishing And Not Furniture}, $\mu(A' \ {\rm and} \ B)=0.39$ with respect to {\it Not Home Furnishing And Furniture}, and $\mu(A' \ {\rm and} \ B')=0.41$ with respect to {\it Not Home Furnishing And Not Furniture}. If one first looks for a representation of {\it Window Seat} in the Hilbert space $\mathbb{C}^8$, then the concepts {\it Home Furnishing} and {\it Furniture}, and their negations {\it Not Home Furnishing} and {\it Not Furniture} are respectively represented by the unit vectors
$|A\rangle=e^{i \phi_{A}}	(-0.01,0.69,0.14,-0.01,-0.13,-0.66,-0.2,0.11)$
and
$|B\rangle=e^{i \phi_{B}}	(-0.08,-0.39,-0.6,0,-0.03,-0.4,$ $-0.17,0.54)$,
and
$|A'\rangle=e^{i \phi_{A'}}(0.13,-0.19,0.69,0.01,0.09,0.05,-0.05,0.67)$
and
$|B'\rangle=e^{i \phi_{B'}}(-0.09,0.57,$ $-0.34,-0.02,0.17,0.54,0.11,0.47)$.

The interference angles $\phi_{AB}=\phi_{B}-\phi_{A}=76.57^{\circ}$, $\phi_{AB'}=\phi_{B'}-\phi_{A}=103.86^{\circ}$, $\phi_{A'B}=\phi_{B}-\phi_{A'}=84.42^{\circ}$ and $\phi_{A'B'}=\phi_{B'}-\phi_{A'}=85.94^{\circ}$ complete the Hilbert space representation in $\mathbb{C}^8$. A complete modeling in the Fock space $\mathbb{C}^8 \oplus (\mathbb{C}^8\otimes \mathbb{C}^8)$ satisfying Equations (\ref{FockSpaceSolutionAB}), (\ref{FockSpaceSolutionAB'}), (\ref{FockSpaceSolutionA'B}) and (\ref{FockSpaceSolutionA'B'}) is given by an entangled state characterized by $\sum_{i,j=5}^{8}c_{ij}^2=0.31^2$, $\sum_{i=5}^{8}\sum_{i=1}^{4}c_{ij}^{2}=0.64^2$, $\sum_{i=1}^{4}\sum_{j=5}^{8}c_{ij}^2=0.62^2$ and $\sum_{i,j=1}^{4}c_{ij}^{2}=0.34^2$, and convex weights 
$m_{AB}=0.51$,
$n_{AB}=0.86$, 
$m_{AB'}=0.77$, 
$n_{AB'}=0.63$, 
$m_{A'B}=1$,  $n_{A'B}=0$, $m_{A'B'}=0.54$, and $n_{A'B'}=0.84$.

The theoretic analysis on the representatibility of the data in Tables 1--4 is thus concluded. We stress that the majority of these data can be faithfully modeled by using the mathematical formalism of quantum theory in Fock space. We finally observe that a big amount of the collected data can be modeled by using only the first sector of Fock space, while almost all the weights of $n^2$-type in first sector prevail over the weights of $m^2$-type in second sector. The reasons of this will be clear after the discussion in Section \ref{results}.

\section{Confirmations of Fock space framework \label{results}}
Our experimental data on conjunctions and negations of natural concepts confirm that classical probability does not generally work when humans combine concepts, as we have seen in the previous sections. However, we have also proved here that the deviations from classicality cannot be reduced to overextension and underextension, while our quantum-theoretic framework in Fock space has received remarkable corroboration. Hence, we think it worth to summarize and stress the novelties that have emerged in this article with respect to our approach.

We have recently put forward an explanatory hypothesis with respect to the deviations from classical logical reasoning that have been observed in human cognition \cite{IQSA2}. According to our explanatory hypothesis, human reasoning is a specifically structured superposition of two processes, a `logical reasoning' and an `emergent reasoning'. The former `logical reasoning' combines cognitive entities, such as concepts, combinations of concepts, or propositions,  by applying the rules of logic, though 
generally in a probabilistic way. The latter `emergent reasoning' enables formation of combined cognitive entities as 
newly emerging entities, in the case of concepts, new concepts, in the case of propositions, new propositions, 
carrying new meaning, linked to the meaning of the constituent cognitive entities, but with a linkage not defined by the algebra of logic. The two mechanisms act simultaneously and in superposition in human thought during a reasoning process, the first one is guided by an algebra of `logic', the second one follows a mechanism of `emergence'. In this perspective, human reasoning can be mathematically formalized in a two-sector Fock space. More specifically, first sector of Fock space models `conceptual emergence', while second sector of Fock space models a conceptual combination from the combining concepts by requiring 
the rules of logic for the logical connective used for the combining to be satisfied in a probabilistic setting. The relative prevalence of emergence  or logic in a specific cognitive process is measured by the `degree of participation' of second and first sectors, respectively. The abundance of evidence of deviations from classical logical reasoning in concrete human decisions (paradoxes, fallacies, effects, contradictions), together with our results, led us to draw the conclusion that emergence constitutes the dominant dynamics of human reasoning, while logic is only a secondary form of dynamics. 

Now, if one reflects on how we represented conceptual negation in Section \ref{brussels}, one realizes at once that its modeling directly and naturally follows from the general assumption stated above. Indeed, suppose that a given subject is asked to estimate whether a given exemplar $x$ is a member of the concepts $A$, $B'$, `$A \ {\rm and} \ B'$ (a completely equivalent explanation can be given for the conjunctions `$A'$ and $B$' and `$A'$ and $B'$'). Then, our quantum mathematics can be interpreted by assuming that a `quantum logical thought' acts, where the subject considers two copies of $x$ and estimates whether the first copy belongs to $A$ and the second copy of $x$ `does not' belong to $B$. But also a `quantum conceptual thought' acts, where the subject estimates whether the exemplar $x$ belongs to the newly emergent concept `$A \ {\rm and} \ B'$'. The place whether these superposed processes can be suitably structured is the two-sector Fock space. First sector of Fock space hosts the latter process, second sector hosts the former, and the weights $m_{AB'}^2$ and $n_{AB'}^2$ indicate whether the overall process is mainly guided by logic or emergence.

The second confirmation of our quantum-conceptual framework comes from the significantly stable deviations from Equations (\ref{condbis01})--(\ref{condbis05}) discussed in Section \ref{res-experiments}. These deviations indeed occur at a different, deeper, level than  overextension and underextension. We think we have identified a general mechanism determining how concepts are formed in the human mind. And this would already be convincing even without mentioning a Fock space modeling. But, this very stable pattern can exactly be explained in our two-sector Fock space framework by assuming that emergence plays a primary role in the human reasoning process, but also aspects of logic are systematically present. Indeed, suppose that, for every exemplar $x$ and every $X=A,A',Y=B,B'$,  $n_{XY}^2=1$ and $m_{XY}^2=0$, that is, the decision process only occurs in first sector of Fock space. This assumption corresponds, from our quantum modeling perspective (see  (\ref{FockSpaceSolutionAB})--(\ref{FockSpaceSolutionA'B'})), to a situation where only emergence is present. Thus, we can write  $\mu(X \ {\rm and } \ Y)=\frac{1}{2}({\mu(X)+\mu(Y)})$. Indeed, since interference in these equations can be negative as well as positive, and there is a priori no reason to suppose that there would be more of the one than the other, we can neglect the interference parts of the equations, because it is reasonable to suppose that they will cancel out when summing on all the terms of the equations of our classicality conditions. A simple calculation shows that, for every $X=A,A',Y=B,B'$, $I_{X}=I_{Y}=-0.5$ and $I_{ABA'B'}=-1$, in this case. Then, an immediate comparison with the experimental values of $I_{X}$, $I_{Y}$ and $I_{ABA'B'}$ in Section \ref{res-experiments} reveals that a component of second sector of Fock space is also present, which is generally smaller than the component of first sector but systematic across all exemplars. The consequence is immediate: both emergence and logic play a role in the decision process -- emergence is dominant, but also logic is systematically present. We believe that this finding is really a fundamental one, and it deserves further investigation in the future.

The third strong confirmation of this two-layered structure of human thought and its representation in two-sector Fock space comes from the peculiarities of conceptual negation. Indeed, being pushed to cope with conceptual conjunction and negation simultaneously, we have found a new insight which we had not noticed before, namely, the emergent non-classical properties of the conjunction {\it Fruits And Vegetables} are naturally accounted for in first sector of Fock space, while the fact that {\it Not Vegetables} does not have a well defined prototype, but it is rather the negation of {\it Vegetables}, is accounted for in second sector of Fock space, where logic occurs. In both cases, the Fock space model has naturally suggested us the right directions to follow.

The fourth corroboration derives from the fact that our Fock space indicates how and why introducing entanglement. In our previous attempts to model conceptual combinations, we had not recognised that by representing the combined concept by a tensor product vector $|A\rangle \otimes |B\rangle$, we impliticly assumed that membership weights probabilities are factorised in second sector, that is, the membership weights $\mu(A)$ and $\mu(B)$ correspond to independent events in this sector. In Section  \ref{entanglement} we have showed that, if one introduces entangled states to represent combined concepts in second sector, one is able to fully reproduce all classicality conditions  (\ref{condbis01})--(\ref{condbis05}) in this sector. And, more, one can formalize the fact that, for certain exemplars, the probabilities associated with memberships of, say {\it Fruits} and {\it Vegetables}, are not independent. Therefore, Fock space has suggested how to capture this relevant aspect in depth.

To conclude we can safely say that the success of our two-sector Fock space fremework goes far beyond faithful represention of a set of experimental data. It has allowed us to capture some fundamental aspects of the mechanisms through which concepts are formed, combine and interact in our thought.

\section{A remark on standard quantum logic in second sector\label{quantumlogic}}
We have claimed in Section \ref{results} that human thought is in our quantum-theoretic framework the superposition of two layers, a `quantum conceptual thought', guided by `emergence', and a `quantum logical thought', guided by `logic'. And we have made the latter explicit by representing in second sector of Fock space the decision measurement of a subject estimating the negation $A'$ of a concept $A$ by the orthogonal projection operator $\mathbbmss{1}-M$, where $M$ represents the decision measurement on the membership with respect to $A$ -- standard quantum logical negation of a quantum proposition represented by $M$ would exactly be represented by $\mathbbmss{1}-M$. On the other hand, we have proved in Section \ref{entanglement} that the five classicality conditions (\ref{condbis01})--(\ref{condbis05}) in Theorem 3$'$ are satisfied in this second sector, which would induce to maintain that
it is standard classical logic, in its probabilistic version, that holds in second sector. How are these two findings connected? Is it possible to specify the precise sense in which the term `quantum logic' should be interpreted in second sector of Fock space? These are the questions that we address in this section.

Let us preliminary call a `standard quantum logic' the complete orthocomplemented lattice $( {\mathcal L}({\mathcal H}),\subset, \wedge, \vee, {}^{\perp} )$ of closed subspaces of a complex Hilbert space ${\cal H}$, where $\subset$ denotes set-theoretical inclusion, $\wedge$ and $\vee$ respectively denote meet and join in ${\mathcal L}({\mathcal H})$, and ${\cal A}^{\perp}$ is the orthogonal complement of the closed subspace $\cal A$. The lattice $( {\mathcal L}({\mathcal H}),\subset, \wedge, \vee, {}^{\perp} )$ is isomomorphic to the complete orthocomplemented lattice of orthogonal projection operators on a complex Hilbert space ${\cal H}$, hence we will denote these two lattices by the same symbols, for the sake of brevity.

Suppose now that the closed subspaces $\cal A$ and $\cal B$ respectively correspond to the orthogonal projection operators $P_A$ and $P_B$, such that $P_AP_B=P_BP_A$ (equivalently, $[P_A,P_B]=0$), that is, suppose that $P_A$ and $P_B$ commute. Then, the five conditions (\ref{condbis01})--(\ref{condbis05}) in Theorem 3$'$ that we have identified as classicality conditions are exactly the same conditions that would express commutativity in ${\mathcal L}({\mathcal H})$. Indeed, a commutativity condition is expressed by
\begin{eqnarray} \label{commutativity}
{\cal A}=({\cal A} \wedge {\cal B})\vee({\cal A} \wedge {\cal B}^\perp)
\end{eqnarray}
Consider now the Boolean sub-lattice generated by the closed subspaces ${\cal A},{\cal B},{\cal A}^{\perp},{\cal B}^{\perp}$, and let $\mu$ be a probability measure over this Boolean sub-lattice. Then, Equation (\ref{commutativity}) implies that $\mu({\cal A})=\mu({\cal A} \wedge {\cal B})+\mu({\cal A} \wedge {\cal B}^{\perp})$, which perfectly corresponds to Equation (\ref{condbis01}). The other compatibility conditions following from Equation (\ref{commutativity}) correspond to Equations (\ref{condbis02})--(\ref{condbis05}).

Thus, there is a deep structural equivalence, which gives the impression that in fact the five conditions are conditions rather of `commutativity' within a quantum logic setting, than of conditions for a pure classical setting. This is exactly the meaning we should give to the locution `quantum logic' which holds in sector two of Fock space, namely that standard classical logical rules hold in this sector, and these logical rules correspond to the logical rules of specific Boolean sublattices of a standard quantum logic.

Summing up, what we identified in second sector of Fock space has the structure of a `quantum logic', whenever only commuting propositions are considered, and this quantum logical structure has a classical probabilistic counterpart satisfying conditions (\ref{condbis01})--(\ref{condbis05}) in Theorem 3$'$. This finding on the interpretation of quantum logical structures clarifies some aspects of our previous quantum-theoretic modeling \cite{a2009a,ags2013,s2014a,s2014b}. But we think this interesting aspect of our quantum-theoretic approach to concepts deserves further investigation.

\section{Appendices}
\appendix

\section{Fundamentals of modeling in a classical framework\label{classical}}
We introduce in this section the elementary measure-theoretic notions that are needed to express the classicality of experimental data coming from the membership weights of two concepts $A$ and $B$ with respect to the conceptual negation `${\rm not} \ B$' and the conjunctions `$A \ \rm{and} \ B$', `$A \ \rm{and} \ {\rm not} \ B$', `${\rm not} \ A \ \rm{and} \ B$' and `${\rm not} \ A \ \rm{and} \ {\rm not} \ B$'. As we have anticipated in Section \ref{experiment}, by `classicality of a collection of experimental date' we actually mean the possibility to represent them in a `classical', or `Kolmogorovian', probability model. We avoid in our presentation superfuous technicalities, but aim to be synthetic and rigorous at the same time.

Let us start by the definition of a $\sigma$-algebra over a set.

\medskip
\noindent
{\it First definition}. A \emph{$\sigma$-algebra} over a set $\Omega$ is a non-empty collection $\sigma(\Omega)$ of subsets of $\Omega$ that is closed under complementation and countable unions of its members. It is a Boolean algebra, completed to include countably infinite operations.

\medskip
\noindent
Measure structures are the most general classical structures devised by mathematicians and physicists to structure weights. A Kolmogorovian probability measure is such a measure applied to statistical data. It is called `Kolmogorovian', because Andrey Kolmogorov was the first to axiomatize probability theory in this way \cite{k1933}.

\medskip
\noindent
{\it Second definition}. A \emph{measure} $P$ is a function defined on a $\sigma$-algebra $\sigma(\Omega)$ over a set $\Omega$ and taking values in the extended interval $[0,\infty]$ such that the following three conditions are satisfied:

(i) the empty set has measure zero;

(ii) if  $E_1$, $E_2$, $E_3$, $\dots$ is a countable sequence of pairwise disjoint sets in $\sigma(\Omega)$, the measure of the union of all the $E_i$ is equal to the sum of the measures of each $E_i$ \emph{(countable additivity}, or \emph{$\sigma$-additivity)};

(iii) the triple $(\Omega,\sigma(\Omega),P)$ satisfying (i) and (ii) is then called a measure space, and the members of $\sigma(\Omega)$ are called measurable sets.

A \emph{Kolmogorovian probability measure} is a measure with total measure one. A Kolmogorovian probability space $(\Omega,\sigma(\Omega),P)$ is a measure space $(\Omega,\sigma(\Omega),P)$ such that $P$ is a Kolmogorovian probability. The three conditions expressed in a mathematical way are:
\begin{equation} \label{KolmogorovianMeasure}
P(\emptyset)=0 \quad P(\bigcup_{i=1}^\infty E_i)=\sum_{i=1}^\infty P(E_i) \quad P(\Omega)=1
\end{equation}

\medskip
\noindent
Let us now come to the possibility to represent a set of experimental data on two concepts and their conjunction in a classical Kolmogorovian probability model.

\medskip
\noindent
{\it Third definition}. We say that the membership weights $\mu(A)$, $\mu(B)$ and $\mu(A\ {\rm and}\ B)$ of the exemplar $x$ with respect to the pair of concepts $A$ and $B$ and their conjunction `$A$ and $B$', respectively, can be represented in a \emph{classical Kolmogorovian probability model} if there exists a Kolmogorovian probability space $(\Omega,\sigma(\Omega),P)$ and events $E_A, E_B \in \sigma(\Omega)$ of the events algebra $\sigma(\Omega)$ such that
\begin{equation}
P(E_A) = \mu(A) \quad P(E_B) = \mu(B) \quad {\rm and} \quad P(E_A \cap E_B) = \mu(A\ {\rm and}\ B)
\end{equation}

\medskip
\noindent
Let us finally come to the representability a set of experimental data on a concept and its negation in a classical Kolmogorovian probability model.

\medskip
\noindent
{\it Fourth definition}. We say that the membership weights $\mu(B)$ and $\mu({\rm not} \ B)$ of the exemplar $x$ with respect to the concept $B$ and its negation `${\rm not} \ B$', respectively, can be represented in a classical Kolmogorovian probability model if there exists a Kolmogorovian probability space $(\Omega,\sigma(\Omega),P)$ and an event $E_B \in \sigma(\Omega)$ of the events algebra $\sigma(\Omega)$ such that
\begin{equation} \label{definitionnegation}
P(E_B) = \mu(B) \quad P(\Omega \setminus E_{B}) = \mu({\rm not} \ B)
\end{equation}

\section{Quantum mathematics for conceptual modeling\label{quantum}}
We illustrate in this section how the mathematical formalism of quantum theory can be applied to model situations outside the microscopic quantum world, more specifically, in the representation of concepts and their combinations. As in Appendix \ref{classical}, we will limit technicalities to the essential.

When the quantum mechanical formalism is applied for modeling purposes, each considered entity  -- in our case a concept -- is associated with a complex Hilbert space ${\cal H}$, that is, a vector space over the field ${\mathbb C}$ of complex numbers, equipped with an inner product $\langle \cdot |  \cdot \rangle$ that maps two vectors $\langle A|$ and $|B\rangle$ onto a complex number $\langle A|B\rangle$. We denote vectors by using the bra-ket notation introduced by Paul Adrien Dirac, one of the pioneers of quantum theory \cite{d1958}. Vectors can be `kets', denoted by $\left| A \right\rangle $, $\left| B \right\rangle$, or `bras', denoted by $\left\langle A \right|$, $\left\langle B \right|$. The inner product between the ket vectors $|A\rangle$ and $|B\rangle$, or the bra-vectors $\langle A|$ and $\langle B|$, is realized by juxtaposing the bra vector $\langle A|$ and the ket vector $|B\rangle$, and $\langle A|B\rangle$ is also called a `bra-ket', and it satisfies the following properties:

(i) $\langle A |  A \rangle \ge 0$;

(ii) $\langle A |  B \rangle=\langle B |  A \rangle^{*}$, where $\langle B |  A \rangle^{*}$ is the complex conjugate of $\langle A |  B \rangle$;

(iii) $\langle A |(z|B\rangle+t|C\rangle)=z\langle A |  B \rangle+t \langle A |  C \rangle $, for $z, t \in {\mathbb C}$,
where the sum vector $z|B\rangle+t|C\rangle$ is called a `superposition' of vectors $|B\rangle$ and $|C\rangle$ in the quantum jargon.

From (ii) and (iii) follows that inner product $\langle \cdot |  \cdot \rangle$ is linear in the ket and anti-linear in the bra, i.e. $(z\langle A|+t\langle B|)|C\rangle=z^{*}\langle A | C\rangle+t^{*}\langle B|C \rangle$.

We recall that the `absolute value' of a complex number is defined as the square root of the product of this complex number times its complex conjugate, that is, $|z|=\sqrt{z^{*}z}$. Moreover, a complex number $z$ can either be decomposed into its cartesian form $z=x+iy$, or into its polar form $z=|z|e^{i\theta}=|z|(\cos\theta+i\sin\theta)$.  As a consequence, we have $|\langle A| B\rangle|=\sqrt{\langle A|B\rangle\langle B|A\rangle}$. We define the `length' of a ket (bra) vector $|A\rangle$ ($\langle A|$) as $|| |A\rangle ||=||\langle A |||=\sqrt{\langle A |A\rangle}$. A vector of unitary length is called a `unit vector'. We say that the ket vectors $|A\rangle$ and $|B\rangle$ are `orthogonal' and write $|A\rangle \perp |B\rangle$ if $\langle A|B\rangle=0$.

We have now introduced the necessary mathematics to state the first modeling rule of quantum theory, as follows.

\medskip
\noindent{\it First quantum modeling rule:} A state $A$ of an entity -- in our case a concept -- modeled by quantum theory is represented by a ket vector $|A\rangle$ with length 1, that is $\langle A|A\rangle=1$.

\medskip
\noindent
An orthogonal projection $M$ is a linear operator on the Hilbert space, that is, a mapping $M: {\cal H} \rightarrow {\cal H}, |A\rangle \mapsto M|A\rangle$ which is Hermitian and idempotent. The latter means that, for every $|A\rangle, |B\rangle \in {\cal H}$ and $z, t \in {\mathbb C}$, we have:

(i) $M(z|A\rangle+t|B\rangle)=zM|A\rangle+tM|B\rangle$ (linearity);

(ii) $\langle A|M|B\rangle=\langle B|M|A\rangle^{*}$ (hermiticity);

(iii) $M \cdot M=M$ (idempotency).

The identity operator $\mathbbmss{1}$ maps each vector onto itself and is a trivial orthogonal projection. We say that two orthogonal projections $M_k$ and $M_l$ are orthogonal operators if each vector contained in $M_k({\cal H})$ is orthogonal to each vector contained in $M_l({\cal H})$, and we write $M_k \perp M_l$, in this case. The orthogonality of the projection operators $M_{k}$ and $M_{l}$ can also be expressed by $M_{k}M_{l}=0$, where $0$ is the null operator. A set of orthogonal projection operators $\{M_k\ \vert k=1,\ldots,n\}$ is called a `spectral family' if all projectors are mutually orthogonal, that is, $M_k \perp M_l$ for $k \not= l$, and their sum is the identity, that is, $\sum_{k=1}^nM_k=\mathbbmss{1}$.

The above definitions give us the necessary mathematics to state the second modeling rule of quantum theory, as follows.

\medskip
\noindent
{\it Second quantum modeling rule:} A measurable quantity $Q$ of an entity -- in our case a concept -- modeled by quantum theory, and having a set of possible real values $\{q_1, \ldots, q_n\}$ is represented by a spectral family $\{M_k\ \vert k=1, \ldots, n\}$ in the following way. If the entity -- in our case a concept -- is in a state represented by the vector $|A\rangle$, then the probability of obtaining the value $q_k$ in a measurement of the measurable quantity $Q$ is $\langle A|M_k|A\rangle=||M_k |A\rangle||^{2}$. This formula is called the `Born rule' in the quantum jargon. Moreover, if the value $q_k$ is actually obtained in the measurement, then the initial state is changed into a state represented by the vector
\begin{equation}
|A_k\rangle=\frac{M_k|A\rangle}{||M_k|A\rangle||}
\end{equation}
This change of state is called `collapse' in the quantum jargon.

\medskip
\noindent
The tensor product ${\cal H}_{A} \otimes {\cal H}_{B}$ of two Hilbert spaces ${\cal H}_{A}$ and ${\cal H}_{B}$ is the Hilbert space generated by the set $\{|A_i\rangle \otimes |B_j\rangle\}$, where $|A_i\rangle$ and $|B_j\rangle$ are vectors of ${\cal H}_{A}$ and ${\cal H}_{B}$, respectively, which means that a general vector of this tensor product is of the form $\sum_{ij}|A_i\rangle \otimes |B_j\rangle$. This gives us the necessary mathematics to introduce the third modeling rule.

\medskip
\noindent
{\it Third quantum modeling rule:} A state $C$ of a compound entity -- in our case a combined concept -- is represented by a unit vector $|C\rangle$ of the tensor product ${\cal H}_{A} \otimes {\cal H}_{B}$ of the two Hilbert spaces ${\cal H}_{A}$ and ${\cal H}_{B}$ containing the vectors that represent the states of the component entities -- concepts.

\medskip
\noindent
The above means that we have $|C\rangle=\sum_{ij}c_{ij}|A_i\rangle \otimes |B_j\rangle$, where $|A_i\rangle$ and $|B_j\rangle$ are unit vectors of ${\cal H}_{A}$ and ${\cal H}_{B}$, respectively, and $\sum_{i,j}|c_{ij}|^{2}=1$. We say that the state $C$ represented by $|C\rangle$ is a product state if it is of the form $|A\rangle \otimes |B\rangle$ for some $|A\rangle \in {\cal H}_{A}$ and $|B\rangle \in {\cal H}_{B}$. Otherwise, $C$ is called an `entangled state'.

\medskip
\noindent
The Fock space is a specific type of Hilbert space, originally introduced in quantum field theory. For most states of a quantum field the number of identical quantum entities is not conserved but is a variable quantity. The Fock space copes with this situation in allowing its vectors to be superpositions of vectors pertaining to different sectors for fixed numbers of identical quantum entities. More explicitly, the $k$-th sector of a Fock space describes a fixed number of $k$ identical quantum entities, and it is of the form ${\cal H}\otimes \ldots \otimes{\cal H}$ of the tensor product of $k$ identical Hilbert spaces ${\cal H}$. The Fock space $F$ itself is the direct sum of all these sectors, hence
\begin{equation} \label{fockspace}
{\cal F}=\oplus_{k=1}^j\otimes_{l=1}^k{\cal H}
\end{equation}
For our modeling we have only used Fock space for the `two' and `one quantum entity' case, hence ${\cal F}={\cal H}\oplus({\cal H}\otimes{\cal H})$. This is due to considering only combinations of two concepts. The sector ${\cal H}$ is called the `sector 1', while the sector ${\cal H}\otimes{\cal H}$ is called the `sector 2'. A unit vector $|F\rangle \in {\cal F}$ is then written as $|F\rangle = ne^{i\gamma}|C\rangle+me^{i\delta}(|A\rangle\otimes|B\rangle)$, where $|A\rangle, |B\rangle$ and $|C\rangle$ are unit vectors of ${\cal H}$, and such that $n^2+m^2=1$. For combinations of $j$ concepts, the general form of Fock space expressed in Equation (\ref{fockspace}) will have to be used.

The quantum modeling above can be generalized by allowing states to be represented by the so called `density operators' and measurements to be represented by the so called `positive operator valued measures'. However, we will dot dwell on this extension here, for the sake of brevity.

\section{Data modeling tables and statistical analysis\label{tables}}
\setlength{\tabcolsep}{1.3pt}
\begin{table}[H]
\begin{center}
\tiny
\begin{tabular}{|cccccccccccccccccc|}
\hline
\multicolumn{18}{|l|}{\it $A$=Home Furnishing, $B$=Furniture} \\
\hline				
{\it Exemplar}	 &	$\mu(A)$ & $\mu(B)$ & $\mu(A')$	& $\mu(B')$ & $\mu(A \ {\rm and} \ B)$ &   $\mu(A \ {\rm and} \ B')$ 
& $\mu(A' \ {\rm and} \ B)$ & $\mu(A' \ {\rm and} \ B')$ & $\Delta_{AB}$ & $\Delta_{AB'}$ & $\Delta_{A'B}$ & $\Delta_{A'B'}$ &
$I_{ABA'B'}$ & $I_{A}$ & $I_{B}$ & $I_{A'}$ & $I_{B'}$ \\ 
\hline
Mantelpiece		&	0.9	&	0.61	&	0.12	&	0.5	&	0.71	&	0.75	&	0.21	&	0.21	&	0.1	&	0.25	&	0.09	&	0.09	&	-0.89	&	-0.56	&	-0.31	&	-0.31	&	-0.46	\\
Window Seat		&	0.5	&	0.48	&	0.47	&	0.55	&	0.45	&	0.49	&	0.39	&	0.41	&	-0.03	&	-0.01	&	-0.08	&	-0.06	&	-0.74	&	-0.44	&	-0.36	&	-0.33	&	-0.35	\\
Painting		&	0.8	&	0.49	&	0.35	&	0.64	&	0.64	&	0.6	&	0.33	&	0.38	&	0.15	&	-0.04	&	-0.03	&	0.03	&	-0.94	&	-0.44	&	-0.48	&	-0.35	&	-0.33	\\
Light Fixture		&	0.88	&	0.6	&	0.16	&	0.51	&	0.73	&	0.63	&	0.33	&	0.16	&	0.13	&	0.11	&	0.16	&	0	&	-0.84	&	-0.48	&	-0.45	&	-0.33	&	-0.28	\\
Kitchen Counter		&	0.67	&	0.49	&	0.31	&	0.62	&	0.55	&	0.54	&	0.38	&	0.33	&	0.06	&	-0.08	&	0.06	&	0.01	&	-0.79	&	-0.42	&	-0.44	&	-0.39	&	-0.24	\\
Bath Tub		&	0.73	&	0.51	&	0.28	&	0.46	&	0.59	&	0.59	&	0.36	&	0.29	&	0.08	&	0.13	&	0.08	&	0.01	&	-0.83	&	-0.45	&	-0.44	&	-0.37	&	-0.41	\\
Deck Chair		&	0.73	&	0.9	&	0.27	&	0.2	&	0.74	&	0.41	&	0.54	&	0.18	&	0.01	&	0.21	&	0.27	&	-0.03	&	-0.86	&	-0.42	&	-0.38	&	-0.44	&	-0.39	\\
Shelves		&	0.85	&	0.93	&	0.24	&	0.13	&	0.84	&	0.39	&	0.53	&	0.08	&	-0.01	&	0.26	&	0.29	&	-0.05	&	-0.83	&	-0.38	&	-0.43	&	-0.36	&	-0.34	\\
Rug		&	0.89	&	0.58	&	0.18	&	0.61	&	0.7	&	0.68	&	0.41	&	0.21	&	0.13	&	0.07	&	0.24	&	0.04	&	-1	&	-0.48	&	-0.54	&	-0.45	&	-0.28	\\
Bed		&	0.76	&	0.93	&	0.26	&	0.11	&	0.79	&	0.36	&	0.61	&	0.14	&	0.03	&	0.26	&	0.35	&	0.03	&	-0.9	&	-0.39	&	-0.48	&	-0.49	&	-0.39	\\
Wall-Hangings		&	0.87	&	0.46	&	0.21	&	0.68	&	0.55	&	0.71	&	0.35	&	0.24	&	0.09	&	0.03	&	0.14	&	0.03	&	-0.85	&	-0.39	&	-0.44	&	-0.38	&	-0.27	\\
Space Rack		&	0.38	&	0.43	&	0.63	&	0.62	&	0.41	&	0.49	&	0.43	&	0.58	&	0.04	&	0.11	&	0	&	-0.04	&	-0.9	&	-0.53	&	-0.41	&	-0.37	&	-0.44	\\
Ashtray		&	0.74	&	0.4	&	0.32	&	0.64	&	0.49	&	0.6	&	0.36	&	0.39	&	0.09	&	-0.04	&	0.04	&	0.07	&	-0.84	&	-0.34	&	-0.45	&	-0.43	&	-0.35	\\
Bar		&	0.72	&	0.63	&	0.37	&	0.51	&	0.61	&	0.61	&	0.4	&	0.4	&	-0.01	&	0.11	&	0.03	&	0.03	&	-1.03	&	-0.51	&	-0.39	&	-0.43	&	-0.51	\\
Lamp		&	0.94	&	0.64	&	0.15	&	0.49	&	0.75	&	0.7	&	0.4	&	0.2	&	0.11	&	0.21	&	0.25	&	0.05	&	-1.05	&	-0.51	&	-0.51	&	-0.45	&	-0.41	\\
Wall Mirror		&	0.91	&	0.76	&	0.13	&	0.45	&	0.83	&	0.66	&	0.44	&	0.14	&	0.07	&	0.21	&	0.31	&	0.01	&	-1.06	&	-0.58	&	-0.51	&	-0.45	&	-0.35	\\
Door Bell		&	0.75	&	0.33	&	0.32	&	0.79	&	0.5	&	0.64	&	0.34	&	0.51	&	0.17	&	-0.11	&	0.02	&	0.19	&	-0.99	&	-0.39	&	-0.51	&	-0.53	&	-0.36	\\
Hammock		&	0.62	&	0.66	&	0.41	&	0.41	&	0.6	&	0.5	&	0.56	&	0.31	&	-0.02	&	0.09	&	0.16	&	-0.09	&	-0.98	&	-0.48	&	-0.5	&	-0.47	&	-0.41	\\
Desk		&	0.78	&	0.95	&	0.31	&	0.09	&	0.78	&	0.33	&	0.75	&	0.15	&	-0.01	&	0.24	&	0.44	&	0.06	&	-1	&	-0.32	&	-0.58	&	-0.59	&	-0.39	\\
Refrigerator		&	0.74	&	0.73	&	0.26	&	0.41	&	0.66	&	0.55	&	0.46	&	0.25	&	-0.06	&	0.14	&	0.21	&	-0.01	&	-0.93	&	-0.47	&	-0.4	&	-0.46	&	-0.39	\\
Park Bench		&	0.53	&	0.66	&	0.59	&	0.46	&	0.55	&	0.29	&	0.56	&	0.39	&	0.02	&	-0.17	&	-0.03	&	-0.07	&	-0.79	&	-0.31	&	-0.45	&	-0.36	&	-0.22	\\
Waste Paper Basket		&	0.69	&	0.54	&	0.36	&	0.63	&	0.59	&	0.41	&	0.46	&	0.49	&	0.04	&	-0.22	&	0.1	&	0.13	&	-0.95	&	-0.31	&	-0.51	&	-0.59	&	-0.27	\\
Sculpture		&	0.83	&	0.46	&	0.34	&	0.66	&	0.58	&	0.73	&	0.46	&	0.36	&	0.11	&	0.07	&	0.13	&	0.03	&	-1.13	&	-0.48	&	-0.58	&	-0.49	&	-0.43	\\
Sink Unit		&	0.71	&	0.57	&	0.34	&	0.58	&	0.6	&	0.56	&	0.38	&	0.38	&	0.03	&	-0.01	&	0.04	&	0.04	&	-0.91	&	-0.46	&	-0.41	&	-0.41	&	-0.36	\\
\hline
\end{tabular}
\normalsize
\end{center}
{\rm Table 1.} Representation of the membership weights in the case of the concepts {\it Home Furnishing} and {\it Furniture}.
\end{table}

\begin{table}[H]
\begin{center}
\tiny
\begin{tabular}{|cccccccccccccccccc|}
\hline
\multicolumn{18}{|l|}{\it $A$=Spices, $B$=Herbs} \\
\hline					
{\it Exemplar}	 &	$\mu(A)$ & $\mu(B)$ & $\mu(A')$	& $\mu(B')$ & $\mu(A \ {\rm and} \ B)$ &   $\mu(A \ {\rm and} \ B')$ 
& $\mu(A' \ {\rm and} \ B)$ & $\mu(A' \ {\rm and} \ B')$ & $\Delta_{AB}$ & $\Delta_{AB'}$ & $\Delta_{A'B}$ & $\Delta_{A'B'}$ &
$I_{ABA'B'}$ & $I_{A}$ & $I_{B}$ & $I_{A'}$ & $I_{B'}$ \\ 
\hline
Molasses		&	0.36	&	0.13	&	0.67	&	0.84	&	0.24	&	0.54	&	0.25	&	0.73	&	0.11	&	0.18	&	0.12	&	0.06	&	-0.75	&	-0.41	&	-0.36	&	-0.31	&	-0.43	\\
Salt		&	0.67	&	0.04	&	0.36	&	0.92	&	0.24	&	0.69	&	0.09	&	0.6	&	0.19	&	0.02	&	0.04	&	0.24	&	-0.61	&	-0.26	&	-0.28	&	-0.33	&	-0.37	\\
Peppermint		&	0.67	&	0.93	&	0.38	&	0.1	&	0.7	&	0.38	&	0.55	&	0.15	&	0.03	&	0.28	&	0.18	&	0.05	&	-0.78	&	-0.41	&	-0.33	&	-0.33	&	-0.43	\\
Curry		&	0.96	&	0.28	&	0.04	&	0.78	&	0.54	&	0.88	&	0.16	&	0.21	&	0.26	&	0.1	&	0.13	&	0.18	&	-0.79	&	-0.45	&	-0.42	&	-0.34	&	-0.31	\\
Oregano		&	0.81	&	0.86	&	0.21	&	0.13	&	0.79	&	0.4	&	0.5	&	0.08	&	-0.03	&	0.28	&	0.29	&	-0.05	&	-0.76	&	-0.38	&	-0.43	&	-0.36	&	-0.35	\\
MSG		&	0.44	&	0.12	&	0.59	&	0.85	&	0.23	&	0.58	&	0.24	&	0.73	&	0.11	&	0.13	&	0.12	&	0.13	&	-0.76	&	-0.36	&	-0.34	&	-0.37	&	-0.45	\\
Chili Pepper		&	0.98	&	0.53	&	0.05	&	0.56	&	0.8	&	0.9	&	0.28	&	0.13	&	0.27	&	0.34	&	0.23	&	0.08	&	-1.1	&	-0.73	&	-0.54	&	-0.35	&	-0.46	\\
Mustard		&	0.65	&	0.28	&	0.39	&	0.71	&	0.49	&	0.65	&	0.23	&	0.46	&	0.21	&	0	&	-0.05	&	0.08	&	-0.83	&	-0.49	&	-0.44	&	-0.3	&	-0.41	\\
Mint		&	0.64	&	0.96	&	0.43	&	0.09	&	0.79	&	0.31	&	0.64	&	0.11	&	0.14	&	0.23	&	0.21	&	0.03	&	-0.85	&	-0.46	&	-0.47	&	-0.32	&	-0.34	\\
Cinnamon		&	1	&	0.49	&	0.02	&	0.51	&	0.69	&	0.79	&	0.21	&	0.15	&	0.19	&	0.28	&	0.19	&	0.13	&	-0.84	&	-0.48	&	-0.41	&	-0.34	&	-0.43	\\
Parsley		&	0.54	&	0.9	&	0.54	&	0.09	&	0.68	&	0.26	&	0.73	&	0.18	&	0.14	&	0.18	&	0.19	&	0.09	&	-0.84	&	-0.4	&	-0.5	&	-0.36	&	-0.35	\\
Saccarin		&	0.34	&	0.14	&	0.68	&	0.88	&	0.24	&	0.54	&	0.24	&	0.8	&	0.1	&	0.19	&	0.1	&	0.12	&	-0.81	&	-0.43	&	-0.34	&	-0.36	&	-0.46	\\
Poppy Seeds		&	0.82	&	0.47	&	0.29	&	0.54	&	0.59	&	0.66	&	0.31	&	0.28	&	0.12	&	0.13	&	0.02	&	-0.02	&	-0.84	&	-0.43	&	-0.43	&	-0.29	&	-0.4	\\
Pepper		&	0.99	&	0.47	&	0.1	&	0.58	&	0.7	&	0.9	&	0.18	&	0.14	&	0.23	&	0.32	&	0.08	&	0.04	&	-0.91	&	-0.61	&	-0.41	&	-0.21	&	-0.46	\\
Turmeric		&	0.88	&	0.53	&	0.11	&	0.43	&	0.74	&	0.69	&	0.28	&	0.21	&	0.21	&	0.26	&	0.16	&	0.1	&	-0.91	&	-0.54	&	-0.49	&	-0.38	&	-0.47	\\
Sugar		&	0.45	&	0.34	&	0.59	&	0.77	&	0.35	&	0.56	&	0.25	&	0.65	&	0.01	&	0.11	&	-0.09	&	0.06	&	-0.81	&	-0.46	&	-0.26	&	-0.31	&	-0.44	\\
Vinegar		&	0.3	&	0.11	&	0.76	&	0.88	&	0.15	&	0.41	&	0.26	&	0.83	&	0.04	&	0.11	&	0.16	&	0.07	&	-0.65	&	-0.26	&	-0.31	&	-0.33	&	-0.36	\\
Sesame Seeds		&	0.8	&	0.49	&	0.3	&	0.59	&	0.59	&	0.7	&	0.34	&	0.29	&	0.1	&	0.11	&	0.04	&	-0.01	&	-0.91	&	-0.49	&	-0.44	&	-0.33	&	-0.4	\\
Lemon Juice		&	0.28	&	0.2	&	0.74	&	0.81	&	0.15	&	0.43	&	0.39	&	0.81	&	-0.05	&	0.15	&	0.19	&	0.07	&	-0.78	&	-0.3	&	-0.34	&	-0.46	&	-0.43	\\
Chocolate		&	0.27	&	0.21	&	0.78	&	0.8	&	0.2	&	0.46	&	0.38	&	0.78	&	-0.01	&	0.19	&	0.16	&	-0.01	&	-0.81	&	-0.39	&	-0.36	&	-0.37	&	-0.44	\\
Horseradish		&	0.61	&	0.67	&	0.48	&	0.28	&	0.61	&	0.4	&	0.53	&	0.33	&	0	&	0.12	&	0.04	&	0.04	&	-0.86	&	-0.4	&	-0.47	&	-0.37	&	-0.44	\\
Vanilla		&	0.76	&	0.51	&	0.3	&	0.49	&	0.63	&	0.61	&	0.33	&	0.35	&	0.11	&	0.13	&	0.03	&	0.05	&	-0.91	&	-0.48	&	-0.44	&	-0.38	&	-0.48	\\
Chives		&	0.66	&	0.89	&	0.43	&	0.26	&	0.76	&	0.28	&	0.64	&	0.31	&	0.1	&	0.02	&	0.21	&	0.06	&	-0.99	&	-0.38	&	-0.51	&	-0.53	&	-0.33	\\
Root Ginger		&	0.84	&	0.56	&	0.23	&	0.44	&	0.69	&	0.59	&	0.41	&	0.23	&	0.13	&	0.14	&	0.18	&	-0.01	&	-0.91	&	-0.43	&	-0.54	&	-0.41	&	-0.37	\\
\hline
\end{tabular}
\normalsize
\end{center}
{\rm Table 2.} Representation of the membership weights in the case of the concepts {\it Spices} and {\it Herbs}.
\end{table}

\begin{table}[H]
\begin{center}
\tiny
\begin{tabular}{|cccccccccccccccccc|}
\hline
\multicolumn{18}{|l|}{\it $A$=Pets, $B$=Farmyard Animals} \\
\hline					
{\it Exemplar}	 &	$\mu(A)$ & $\mu(B)$ & $\mu(A')$	& $\mu(B')$ & $\mu(A \ {\rm and} \ B)$ &   $\mu(A \ {\rm and} \ B')$ 
& $\mu(A' \ {\rm and} \ B)$ & $\mu(A' \ {\rm and} \ B')$ & $\Delta_{AB}$ & $\Delta_{AB'}$ & $\Delta_{A'B}$ & $\Delta_{A'B'}$ &
$I_{ABA'B'}$ & $I_{A}$ & $I_{B}$ & $I_{A'}$ & $I_{B'}$ \\ 
\hline
Goldfish		&	0.93	&	0.17	&	0.12	&	0.81	&	0.43	&	0.91	&	0.18	&	0.43	&	0.26	&	0.1	&	0.06	&	0.31	&	-0.94	&	-0.41	&	-0.43	&	-0.48	&	-0.53	\\
Robin		&	0.28	&	0.36	&	0.71	&	0.64	&	0.31	&	0.35	&	0.46	&	0.46	&	0.04	&	0.08	&	0.1	&	-0.18	&	-0.59	&	-0.39	&	-0.41	&	-0.22	&	-0.18	\\
Blue-tit		&	0.25	&	0.31	&	0.76	&	0.71	&	0.18	&	0.39	&	0.44	&	0.56	&	-0.08	&	0.14	&	0.13	&	-0.15	&	-0.56	&	-0.31	&	-0.3	&	-0.24	&	-0.24	\\
Collie Dog		&	0.95	&	0.77	&	0.03	&	0.35	&	0.86	&	0.56	&	0.25	&	0.11	&	0.09	&	0.21	&	0.23	&	0.09	&	-0.79	&	-0.48	&	-0.34	&	-0.34	&	-0.33	\\
Camel		&	0.16	&	0.26	&	0.89	&	0.75	&	0.2	&	0.31	&	0.51	&	0.68	&	0.04	&	0.16	&	0.26	&	-0.08	&	-0.7	&	-0.36	&	-0.46	&	-0.3	&	-0.24	\\
Squirrel		&	0.3	&	0.39	&	0.74	&	0.65	&	0.28	&	0.26	&	0.46	&	0.59	&	-0.03	&	-0.04	&	0.07	&	-0.06	&	-0.59	&	-0.24	&	-0.34	&	-0.31	&	-0.2	\\
Guide Dog for Blind		&	0.93	&	0.33	&	0.13	&	0.69	&	0.55	&	0.73	&	0.16	&	0.33	&	0.23	&	0.03	&	0.04	&	0.2	&	-0.76	&	-0.35	&	-0.39	&	-0.36	&	-0.36	\\
Spider		&	0.31	&	0.39	&	0.73	&	0.63	&	0.31	&	0.31	&	0.44	&	0.51	&	0	&	0	&	0.05	&	-0.12	&	-0.58	&	-0.31	&	-0.36	&	-0.23	&	-0.19	\\
Homing Pigeon		&	0.41	&	0.71	&	0.61	&	0.34	&	0.56	&	0.25	&	0.59	&	0.34	&	0.16	&	-0.09	&	-0.03	&	0	&	-0.74	&	-0.41	&	-0.44	&	-0.31	&	-0.25	\\
Monkey		&	0.39	&	0.18	&	0.65	&	0.79	&	0.2	&	0.49	&	0.29	&	0.61	&	0.03	&	0.09	&	0.11	&	-0.04	&	-0.59	&	-0.29	&	-0.31	&	-0.25	&	-0.31	\\
Circus Horse		&	0.3	&	0.48	&	0.74	&	0.6	&	0.34	&	0.35	&	0.53	&	0.48	&	0.04	&	0.05	&	0.04	&	-0.13	&	-0.69	&	-0.39	&	-0.38	&	-0.26	&	-0.23	\\
Prize Bull		&	0.13	&	0.76	&	0.88	&	0.26	&	0.43	&	0.28	&	0.83	&	0.34	&	0.29	&	0.14	&	0.06	&	0.08	&	-0.86	&	-0.57	&	-0.49	&	-0.28	&	-0.35	\\
Rat		&	0.2	&	0.36	&	0.85	&	0.68	&	0.21	&	0.28	&	0.54	&	0.63	&	0.01	&	0.08	&	0.18	&	-0.05	&	-0.65	&	-0.29	&	-0.39	&	-0.31	&	-0.23	\\
Badger		&	0.16	&	0.28	&	0.88	&	0.73	&	0.14	&	0.26	&	0.44	&	0.66	&	-0.03	&	0.1	&	0.16	&	-0.07	&	-0.5	&	-0.24	&	-0.3	&	-0.23	&	-0.19	\\
Siamese Cat		&	0.99	&	0.5	&	0.05	&	0.53	&	0.74	&	0.75	&	0.18	&	0.24	&	0.24	&	0.23	&	0.13	&	0.19	&	-0.9	&	-0.5	&	-0.41	&	-0.36	&	-0.46	\\
Race Horse		&	0.29	&	0.7	&	0.71	&	0.39	&	0.51	&	0.31	&	0.65	&	0.31	&	0.23	&	0.03	&	-0.05	&	-0.08	&	-0.79	&	-0.54	&	-0.46	&	-0.26	&	-0.24	\\
Fox		&	0.13	&	0.3	&	0.86	&	0.68	&	0.18	&	0.29	&	0.46	&	0.59	&	0.04	&	0.16	&	0.16	&	-0.09	&	-0.51	&	-0.33	&	-0.34	&	-0.19	&	-0.19	\\
Donkey		&	0.29	&	0.9	&	0.78	&	0.15	&	0.56	&	0.18	&	0.81	&	0.23	&	0.28	&	0.03	&	0.04	&	0.08	&	-0.78	&	-0.45	&	-0.48	&	-0.26	&	-0.25	\\
Field Mouse		&	0.16	&	0.41	&	0.83	&	0.59	&	0.23	&	0.24	&	0.43	&	0.58	&	0.06	&	0.08	&	0.02	&	-0.01	&	-0.46	&	-0.3	&	-0.24	&	-0.18	&	-0.23	\\
Ginger Tom-cat		&	0.82	&	0.51	&	0.21	&	0.54	&	0.59	&	0.58	&	0.26	&	0.29	&	0.08	&	0.03	&	0.05	&	0.08	&	-0.71	&	-0.34	&	-0.34	&	-0.34	&	-0.32	\\
Husky in Slead team		&	0.64	&	0.51	&	0.37	&	0.53	&	0.56	&	0.51	&	0.44	&	0.29	&	0.06	&	-0.01	&	0.07	&	-0.08	&	-0.8	&	-0.43	&	-0.49	&	-0.36	&	-0.28	\\
Cart Horse		&	0.27	&	0.86	&	0.76	&	0.15	&	0.53	&	0.2	&	0.84	&	0.23	&	0.26	&	0.05	&	0.08	&	0.08	&	-0.79	&	-0.46	&	-0.5	&	-0.31	&	-0.28	\\
Chicken		&	0.23	&	0.95	&	0.8	&	0.06	&	0.58	&	0.11	&	0.81	&	0.18	&	0.34	&	0.05	&	0.01	&	0.11	&	-0.68	&	-0.46	&	-0.44	&	-0.19	&	-0.23	\\
Doberman Guard Dog		&	0.88	&	0.76	&	0.14	&	0.27	&	0.8	&	0.55	&	0.45	&	0.23	&	0.04	&	0.28	&	0.31	&	0.09	&	-1.03	&	-0.47	&	-0.49	&	-0.54	&	-0.51	\\
\hline
\end{tabular}
\normalsize
\end{center}
{\rm Table 3.} Representation of the membership weights in the case of the concepts {\it Pets} and {\it Farmyard Animals}.
\end{table}

\begin{table}[H]
\begin{center}
\tiny
\begin{tabular}{|cccccccccccccccccc|}
\hline
\multicolumn{18}{|l|}{\it $A$=Fruits, $B$=Vegetables} \\
\hline					
{\it Exemplar}	 &	$\mu(A)$ & $\mu(B)$ & $\mu(A')$	& $\mu(B')$ & $\mu(A \ {\rm and} \ B)$ &   $\mu(A \ {\rm and} \ B')$ 
& $\mu(A' \ {\rm and} \ B)$ & $\mu(A' \ {\rm and} \ B')$ & $\Delta_{AB}$ & $\Delta_{AB'}$ & $\Delta_{A'B}$ & $\Delta_{A'B'}$ &
$I_{ABA'B'}$ & $I_{A}$ & $I_{B}$ & $I_{A'}$ & $I_{B'}$ \\ 
\hline
Apple		&	1	&	0.23	&	0	&	0.82	&	0.6	&	0.89	&	0.13	&	0.18	&	0.38	&	0.07	&	0.13	&	0.18	&	-0.79	&	-0.49	&	-0.5	&	-0.3	&	-0.24	\\
Parsley		&	0.02	&	0.78	&	0.99	&	0.25	&	0.45	&	0.1	&	0.84	&	0.44	&	0.43	&	0.08	&	0.06	&	0.19	&	-0.83	&	-0.53	&	-0.51	&	-0.29	&	-0.29	\\
Olive		&	0.53	&	0.63	&	0.47	&	0.44	&	0.65	&	0.34	&	0.51	&	0.36	&	0.12	&	-0.11	&	0.04	&	-0.08	&	-0.86	&	-0.46	&	-0.53	&	-0.41	&	-0.26	\\
Chili Pepper		&	0.19	&	0.73	&	0.83	&	0.35	&	0.51	&	0.2	&	0.68	&	0.44	&	0.33	&	0.01	&	-0.06	&	0.09	&	-0.83	&	-0.53	&	-0.46	&	-0.29	&	-0.29	\\
Broccoli		&	0.09	&	1	&	0.94	&	0.06	&	0.59	&	0.09	&	0.9	&	0.25	&	0.49	&	0.03	&	-0.04	&	0.19	&	-0.83	&	-0.58	&	-0.49	&	-0.21	&	-0.28	\\
Root Ginger		&	0.14	&	0.71	&	0.81	&	0.33	&	0.46	&	0.14	&	0.71	&	0.43	&	0.33	&	0	&	0	&	0.1	&	-0.74	&	-0.46	&	-0.46	&	-0.33	&	-0.24	\\
Pumpkin		&	0.45	&	0.78	&	0.51	&	0.26	&	0.66	&	0.21	&	0.63	&	0.18	&	0.21	&	-0.05	&	0.11	&	-0.09	&	-0.68	&	-0.43	&	-0.51	&	-0.29	&	-0.13	\\
Raisin		&	0.88	&	0.27	&	0.13	&	0.76	&	0.53	&	0.75	&	0.25	&	0.34	&	0.26	&	-0.01	&	0.12	&	0.21	&	-0.86	&	-0.39	&	-0.51	&	-0.46	&	-0.33	\\
Acorn		&	0.59	&	0.4	&	0.49	&	0.64	&	0.46	&	0.49	&	0.38	&	0.51	&	0.06	&	-0.1	&	-0.03	&	0.02	&	-0.84	&	-0.36	&	-0.44	&	-0.39	&	-0.36	\\
Mustard		&	0.07	&	0.39	&	0.87	&	0.6	&	0.29	&	0.23	&	0.55	&	0.75	&	0.22	&	0.16	&	0.16	&	0.15	&	-0.81	&	-0.44	&	-0.45	&	-0.43	&	-0.38	\\
Rice		&	0.12	&	0.46	&	0.9	&	0.52	&	0.21	&	0.23	&	0.59	&	0.59	&	0.09	&	0.11	&	0.13	&	0.07	&	-0.61	&	-0.32	&	-0.34	&	-0.28	&	-0.29	\\
Tomato		&	0.34	&	0.89	&	0.64	&	0.19	&	0.7	&	0.2	&	0.74	&	0.23	&	0.36	&	0.01	&	0.1	&	0.04	&	-0.86	&	-0.56	&	-0.55	&	-0.33	&	-0.24	\\
Coconut		&	0.93	&	0.32	&	0.17	&	0.7	&	0.56	&	0.69	&	0.2	&	0.34	&	0.24	&	-0.01	&	0.03	&	0.17	&	-0.79	&	-0.33	&	-0.44	&	-0.37	&	-0.33	\\
Mushroom		&	0.12	&	0.66	&	0.9	&	0.38	&	0.33	&	0.13	&	0.66	&	0.5	&	0.21	&	0.01	&	0	&	0.12	&	-0.61	&	-0.33	&	-0.33	&	-0.26	&	-0.24	\\
Wheat		&	0.17	&	0.51	&	0.8	&	0.52	&	0.34	&	0.21	&	0.61	&	0.56	&	0.17	&	0.04	&	0.11	&	0.04	&	-0.73	&	-0.38	&	-0.44	&	-0.38	&	-0.26	\\
Green Pepper		&	0.23	&	0.61	&	0.81	&	0.41	&	0.49	&	0.24	&	0.61	&	0.43	&	0.26	&	0.01	&	0	&	0.02	&	-0.76	&	-0.5	&	-0.49	&	-0.23	&	-0.26	\\
Watercress		&	0.14	&	0.76	&	0.89	&	0.25	&	0.49	&	0.1	&	0.79	&	0.35	&	0.35	&	-0.04	&	0.03	&	0.1	&	-0.73	&	-0.45	&	-0.51	&	-0.24	&	-0.2	\\
Peanut		&	0.62	&	0.29	&	0.48	&	0.75	&	0.48	&	0.55	&	0.25	&	0.53	&	0.18	&	-0.07	&	-0.04	&	0.05	&	-0.8	&	-0.41	&	-0.43	&	-0.3	&	-0.33	\\
Black Pepper		&	0.21	&	0.41	&	0.81	&	0.61	&	0.38	&	0.21	&	0.5	&	0.63	&	0.17	&	0.01	&	0.09	&	0.01	&	-0.71	&	-0.38	&	-0.46	&	-0.31	&	-0.23	\\
Garlic		&	0.13	&	0.79	&	0.88	&	0.24	&	0.53	&	0.1	&	0.75	&	0.45	&	0.4	&	-0.03	&	-0.04	&	0.21	&	-0.83	&	-0.5	&	-0.49	&	-0.33	&	-0.31	\\
Yam		&	0.38	&	0.66	&	0.71	&	0.43	&	0.59	&	0.24	&	0.65	&	0.44	&	0.21	&	-0.14	&	-0.01	&	0.01	&	-0.91	&	-0.45	&	-0.58	&	-0.38	&	-0.24	\\
Elderberry		&	0.51	&	0.39	&	0.54	&	0.61	&	0.45	&	0.41	&	0.46	&	0.48	&	0.06	&	-0.09	&	0.07	&	-0.07	&	-0.8	&	-0.36	&	-0.52	&	-0.39	&	-0.28	\\
Almond		&	0.76	&	0.29	&	0.28	&	0.72	&	0.48	&	0.61	&	0.24	&	0.48	&	0.18	&	-0.11	&	-0.04	&	0.19	&	-0.8	&	-0.33	&	-0.42	&	-0.43	&	-0.37	\\
Lentils		&	0.11	&	0.66	&	0.89	&	0.38	&	0.38	&	0.11	&	0.7	&	0.53	&	0.26	&	0	&	0.04	&	0.15	&	-0.71	&	-0.38	&	-0.41	&	-0.33	&	-0.26	\\
\hline
\end{tabular}
\normalsize
\end{center}
{\rm Table 4.} Representation of the membership weights in the case of the concepts {\it Fruits} and {\it Vegetables}.
\end{table}

\begin{table}[H]
\begin{center}
\scriptsize
\begin{tabular}{|c|c||c|c||c|c||c|c|}
\hline
\multicolumn{8}{|l|}{Deviation of $\mu(A)$ from $\mu(A \ {\rm and} \ B)+\mu(A \ {\rm and} \ B')$} \\
\hline		
(Home Furnishing, Furniture)	& p-value	&		(Spices, Herbs)	&	p-value	&	(Pets, Farmyard Animals)	&	p-value	&	(Fruits, Vegetables)	&	p-value	\\
\hline
Mantelpiece	&	5.61E-09	&	Molasses	&	9.73E-07	&	Goldfish	&	2.52E-05	&	Apple	&	1.78E-08	\\
Window Seat	&	1.08E-05	&	Salt	&	7.01E-04	&	Robin	&	1.78E-06	&	Parsley	&	5.14E-07	\\
Painting	&	4.19E-07	&	Peppermint	&	5.57E-06	&	Blue-tit	&	3.56E-06	&	Olive	&	1.98E-05	\\
Light Fixture	&	4.20E-06	&	Curry	&	6.51E-05	&	Collie Dog	&	7.90E-06	&	Chili Pepper	&	1.85E-07	\\
Kitchen Counter	&	6.10E-05	&	Oregano	&	1.78E-06	&	Camel	&	5.92E-05	&	Broccoli	&	4.35E-09	\\
Bath Tub	&	4.17E-06	&	MSG	&	5.93E-05	&	Squirrel	&	5.23E-04	&	Root Ginger	&	1.44E-06	\\
Deck Chair	&	5.35E-06	&	Chili Pepper	&	4.00E-12	&	Guide Dog for Blind	&	6.72E-04	&	Pumpkin	&	9.35E-06	\\
Shelves	&	2.20E-06	&	Mustard	&	2.26E-05	&	Spider	&	4.19E-05	&	Raisin	&	1.13E-06	\\
Rug	&	1.81E-09	&	Mint	&	1.56E-05	&	Homing Pigeon	&	4.87E-05	&	Acorn	&	1.04E-05	\\
Bed	&	5.81E-07	&	Cinnamon	&	8.90E-08	&	Monkey	&	7.21E-04	&	Mustard	&	6.05E-07	\\
Wall-Hangings	&	7.75E-07	&	Parsley	&	2.05E-05	&	Circus Horse	&	3.71E-07	&	Rice	&	7.87E-05	\\
Space Rack	&	2.02E-08	&	Saccarin	&	1.64E-06	&	Prize Bull	&	2.02E-08	&	Tomato	&	1.77E-07	\\
Ashtray	&	2.73E-06	&	Poppy Seeds	&	5.63E-05	&	Rat	&	1.05E-03	&	Coconut	&	1.61E-03	\\
Bar	&	3.07E-08	&	Pepper	&	2.70E-07	&	Badger	&	3.79E-04	&	Mushroom	&	3.40E-05	\\
Lamp	&	4.80E-08	&	Turmeric	&	1.62E-08	&	Siamese Cat	&	4.20E-06	&	Wheat	&	3.76E-06	\\
Wall Mirror	&	1.95E-10	&	Sugar	&	1.52E-07	&	Race Horse	&	1.47E-07	&	Green Pepper	&	3.96E-07	\\
Door Bell	&	5.94E-07	&	Vinegar	&	5.93E-04	&	Fox	&	2.26E-05	&	Watercress	&	2.60E-07	\\
Hammock	&	2.35E-06	&	Sesame Seeds	&	5.49E-07	&	Donkey	&	1.79E-06	&	Peanut	&	7.62E-05	\\
Desk	&	2.94E-05	&	Lemon Juice	&	2.79E-05	&	Field Mouse	&	1.20E-05	&	Black Pepper	&	6.54E-06	\\
Refrigerator	&	2.41E-07	&	Chocolate	&	3.06E-06	&	Ginger Tom-cat	&	9.79E-06	&	Garlic	&	2.38E-07	\\
Park Bench	&	1.09E-06	&	Horseradish	&	1.55E-06	&	Husky in Slead team	&	1.56E-05	&	Yam	&	3.68E-08	\\
Waste Paper Basket	&	2.41E-06	&	Vanilla	&	4.28E-07	&	Cart Horse	&	2.62E-08	&	Elderberry	&	8.82E-05	\\
Sculpture	&	1.43E-06	&	Chives	&	7.42E-06	&	Chicken	&	8.62E-08	&	Almond	&	1.04E-04	\\
Sink Unit	&	3.97E-07	&	Root Ginger	&	5.02E-06	&	Doberman Guard Dog	&	1.29E-04	&	Lentils	&	9.59E-07	\\
\hline
\end{tabular}
\normalsize
\end{center}
{\rm Table 5a.} Calculation of the p-values corresponding to the deviation $I_{A}$ between $\mu(A)$ and $\mu(A \ {\rm and} \ B)+\mu(A \ {\rm and} \ B')$. By applying a Bonferroni correction procedure, the null hypothesis can be rejected for a p-value less than $\frac{0.05}{24}=2.08\cdot 10^{-3}$. 
\end{table}

\begin{table}[H]
\begin{center}
\scriptsize
\begin{tabular}{|c|c||c|c||c|c||c|c|}
\hline
\multicolumn{8}{|l|}{Deviation of $\mu(B)$ from $\mu(A \ {\rm and} \ B)+\mu(A' \ {\rm and} \ B)$} \\
\hline		
(Home Furnishing, Furniture)	& p-value	&		(Spices, Herbs)	&	p-value	&	(Pets, Farmyard Animals)	&	p-value	&	(Fruits, Vegetables)	&	p-value	\\
\hline
Mantelpiece	&	1.09E-05	&	Molasses	&	4.21E-06	&	Goldfish	&	1.89E-07	&	Apple	&	9.00E-07	\\
Window Seat	&	1.89E-05	&	Salt	&	9.37E-05	&	Robin	&	8.27E-07	&	Parsley	&	9.14E-08	\\
Painting	&	2.20E-07	&	Peppermint	&	2.50E-04	&	Blue-tit	&	2.90E-06	&	Olive	&	6.81E-08	\\
Light Fixture	&	2.99E-06	&	Curry	&	8.55E-06	&	Collie Dog	&	2.00E-06	&	Chili Pepper	&	1.60E-07	\\
Kitchen Counter	&	5.12E-06	&	Oregano	&	1.25E-06	&	Camel	&	1.35E-07	&	Broccoli	&	1.22E-06	\\
Bath Tub	&	1.43E-06	&	MSG	&	1.79E-06	&	Squirrel	&	5.85E-06	&	Root Ginger	&	7.24E-06	\\
Deck Chair	&	1.07E-04	&	Chili Pepper	&	2.20E-07	&	Guide Dog for Blind	&	4.55E-06	&	Pumpkin	&	1.17E-07	\\
Shelves	&	7.84E-07	&	Mustard	&	5.57E-06	&	Spider	&	6.15E-05	&	Raisin	&	4.75E-08	\\
Rug	&	4.99E-09	&	Mint	&	1.80E-06	&	Homing Pigeon	&	1.18E-06	&	Acorn	&	9.32E-08	\\
Bed	&	2.73E-06	&	Cinnamon	&	5.03E-06	&	Monkey	&	2.95E-05	&	Mustard	&	4.61E-08	\\
Wall-Hangings	&	3.93E-06	&	Parsley	&	7.52E-06	&	Circus Horse	&	2.27E-05	&	Rice	&	9.76E-06	\\
Space Rack	&	2.16E-08	&	Saccarin	&	2.63E-06	&	Prize Bull	&	2.17E-06	&	Tomato	&	4.63E-07	\\
Ashtray	&	8.83E-07	&	Poppy Seeds	&	3.05E-06	&	Rat	&	4.08E-06	&	Coconut	&	3.58E-06	\\
Bar	&	3.07E-05	&	Pepper	&	4.70E-06	&	Badger	&	8.59E-05	&	Mushroom	&	1.39E-04	\\
Lamp	&	8.94E-07	&	Turmeric	&	2.06E-07	&	Siamese Cat	&	1.98E-05	&	Wheat	&	5.03E-07	\\
Wall Mirror	&	1.05E-06	&	Sugar	&	2.45E-06	&	Race Horse	&	5.03E-06	&	Green Pepper	&	8.15E-07	\\
Door Bell	&	6.27E-07	&	Vinegar	&	1.89E-05	&	Fox	&	9.66E-05	&	Watercress	&	1.26E-07	\\
Hammock	&	4.82E-06	&	Sesame Seeds	&	6.89E-07	&	Donkey	&	2.41E-06	&	Peanut	&	3.58E-06	\\
Desk	&	5.97E-06	&	Lemon Juice	&	1.64E-06	&	Field Mouse	&	1.29E-03	&	Black Pepper	&	3.97E-07	\\
Refrigerator	&	3.82E-05	&	Chocolate	&	9.12E-06	&	Ginger Tom-cat	&	1.64E-06	&	Garlic	&	1.15E-05	\\
Park Bench	&	2.16E-08	&	Horseradish	&	1.31E-05	&	Husky in Slead team	&	1.01E-07	&	Yam	&	1.62E-09	\\
Waste Paper Basket	&	7.21E-08	&	Vanilla	&	1.22E-07	&	Cart Horse	&	2.26E-07	&	Elderberry	&	1.98E-08	\\
Sculpture	&	4.46E-08	&	Chives	&	3.81E-07	&	Chicken	&	4.98E-05	&	Almond	&	2.17E-06	\\
Sink Unit	&	3.64E-06	&	Root Ginger	&	1.02E-08	&	Doberman Guard Dog	&	2.40E-06	&	Lentils	&	7.69E-06	\\
\hline
\end{tabular}
\normalsize
\end{center}
{\rm Table 5b.} Calculation of the p-values corresponding to the deviation $I_{B}$ between $\mu(B)$ and $\mu(A \ {\rm and} \ B)+\mu(A' \ {\rm and} \ B)$. By applying a Bonferroni correction procedure, the null hypothesis can be rejected for a p-value less than $\frac{0.05}{24}=2.08\cdot 10^{-3}$. 
\end{table}

\begin{table}[H]
\begin{center}
\scriptsize
\begin{tabular}{|c|c||c|c||c|c||c|c|}
\hline
\multicolumn{8}{|l|}{Deviation of $\mu(A')$ from $\mu(A' \ {\rm and} \ B)+\mu(A' \ {\rm and} \ B')$} \\
\hline		
(Home Furnishing, Furniture)	& p-value	&		(Spices, Herbs)	&	p-value	&	(Pets, Farmyard Animals)	&	p-value	&	(Fruits, Vegetables)	&	p-value	\\
\hline
Mantelpiece	&	2.38E-05	&	Molasses	&	6.35E-05	&	Goldfish	&	1.51E-05	&	Apple	&	9.93E-05	\\
Window Seat	&	4.88E-04	&	Salt	&	1.63E-04	&	Robin	&	2.85E-03	&	Parsley	&	3.03E-05	\\
Painting	&	6.06E-05	&	Peppermint	&	3.01E-03	&	Blue-tit	&	4.54E-04	&	Olive	&	1.33E-06	\\
Light Fixture	&	5.64E-04	&	Curry	&	7.06E-04	&	Collie Dog	&	1.39E-05	&	Chili Pepper	&	3.47E-05	\\
Kitchen Counter	&	1.55E-05	&	Oregano	&	1.33E-04	&	Camel	&	5.17E-04	&	Broccoli	&	7.60E-03	\\
Bath Tub	&	9.61E-05	&	MSG	&	2.91E-05	&	Squirrel	&	5.10E-05	&	Root Ginger	&	1.71E-04	\\
Deck Chair	&	2.96E-04	&	Chili Pepper	&	8.75E-05	&	Guide Dog for Blind	&	2.75E-05	&	Pumpkin	&	2.55E-04	\\
Shelves	&	6.06E-05	&	Mustard	&	2.14E-03	&	Spider	&	1.06E-02	&	Raisin	&	7.20E-06	\\
Rug	&	1.09E-05	&	Mint	&	2.62E-03	&	Homing Pigeon	&	8.10E-04	&	Acorn	&	1.68E-05	\\
Bed	&	3.00E-05	&	Cinnamon	&	3.65E-04	&	Monkey	&	5.59E-03	&	Mustard	&	1.75E-06	\\
Wall-Hangings	&	1.12E-04	&	Parsley	&	1.33E-04	&	Circus Horse	&	5.80E-04	&	Rice	&	4.90E-03	\\
Space Rack	&	3.28E-06	&	Saccarin	&	6.42E-06	&	Prize Bull	&	3.59E-04	&	Tomato	&	4.41E-04	\\
Ashtray	&	1.55E-05	&	Poppy Seeds	&	2.05E-03	&	Rat	&	1.29E-03	&	Coconut	&	1.86E-04	\\
Bar	&	2.34E-05	&	Pepper	&	7.60E-03	&	Badger	&	2.48E-02	&	Mushroom	&	2.62E-03	\\
Lamp	&	2.26E-05	&	Turmeric	&	5.96E-04	&	Siamese Cat	&	2.71E-04	&	Wheat	&	2.76E-05	\\
Wall Mirror	&	5.58E-06	&	Sugar	&	2.81E-04	&	Race Horse	&	1.28E-03	&	Green Pepper	&	2.75E-02	\\
Door Bell	&	1.64E-05	&	Vinegar	&	5.31E-05	&	Fox	&	9.46E-02	&	Watercress	&	1.41E-03	\\
Hammock	&	2.04E-04	&	Sesame Seeds	&	1.17E-03	&	Donkey	&	6.94E-03	&	Peanut	&	1.20E-05	\\
Desk	&	1.36E-05	&	Lemon Juice	&	1.35E-07	&	Field Mouse	&	2.22E-02	&	Black Pepper	&	1.10E-03	\\
Refrigerator	&	2.55E-05	&	Chocolate	&	3.03E-05	&	Ginger Tom-cat	&	2.79E-04	&	Garlic	&	6.14E-05	\\
Park Bench	&	2.93E-05	&	Horseradish	&	2.80E-06	&	Husky in Slead team	&	3.22E-05	&	Yam	&	1.13E-06	\\
Waste Paper Basket	&	7.87E-09	&	Vanilla	&	2.07E-06	&	Cart Horse	&	2.98E-04	&	Elderberry	&	1.58E-05	\\
Sculpture	&	2.41E-06	&	Chives	&	1.08E-06	&	Chicken	&	2.66E-02	&	Almond	&	6.63E-06	\\
Sink Unit	&	7.90E-06	&	Root Ginger	&	7.32E-04	&	Doberman Guard Dog	&	2.64E-07	&	Lentils	&	6.54E-05	\\
\hline
\end{tabular}
\normalsize
\end{center}
{\rm Table 5c.} Calculation of the p-values corresponding to the deviation $I_{A'}$ between $\mu(A')$ and $\mu(A' \ {\rm and} \ B)+\mu(A' \ {\rm and} \ B')$. By applying a Bonferroni correction procedure, the null hypothesis can be rejected for a p-value less than $\frac{0.05}{24}=2.08\cdot 10^{-3}$. 
\end{table}

\begin{table}[H]
\begin{center}
\scriptsize
\begin{tabular}{|c|c||c|c||c|c||c|c|}
\hline
\multicolumn{8}{|l|}{Deviation of $\mu(B')$ from $\mu(A \ {\rm and} \ B')+\mu(A' \ {\rm and} \ B')$} \\
\hline		
(Home Furnishing, Furniture)	& p-value	&		(Spices, Herbs)	&	p-value	&	(Pets, Farmyard Animals)	&	p-value	&	(Fruits, Vegetables)	&	p-value	\\
\hline
Mantelpiece	&	1.09E-06	&	Molasses	&	3.14E-07	&	Goldfish	&	2.89E-07	&	Apple	&	1.46E-05	\\
Window Seat	&	8.03E-05	&	Salt	&	1.56E-05	&	Robin	&	2.23E-02	&	Parsley	&	5.26E-05	\\
Painting	&	5.38E-05	&	Peppermint	&	1.56E-05	&	Blue-tit	&	1.63E-04	&	Olive	&	4.77E-04	\\
Light Fixture	&	1.16E-03	&	Curry	&	1.83E-04	&	Collie Dog	&	2.06E-04	&	Chili Pepper	&	5.16E-05	\\
Kitchen Counter	&	3.02E-03	&	Oregano	&	1.11E-03	&	Camel	&	5.93E-03	&	Broccoli	&	1.83E-04	\\
Bath Tub	&	4.21E-06	&	MSG	&	2.60E-07	&	Squirrel	&	2.64E-03	&	Root Ginger	&	2.14E-03	\\
Deck Chair	&	4.45E-06	&	Chili Pepper	&	1.71E-07	&	Guide Dog for Blind	&	5.83E-04	&	Pumpkin	&	5.09E-03	\\
Shelves	&	5.06E-04	&	Mustard	&	4.08E-06	&	Spider	&	2.57E-02	&	Raisin	&	2.38E-05	\\
Rug	&	1.17E-05	&	Mint	&	1.13E-04	&	Homing Pigeon	&	9.81E-03	&	Acorn	&	1.37E-05	\\
Bed	&	4.44E-05	&	Cinnamon	&	5.03E-07	&	Monkey	&	9.47E-03	&	Mustard	&	3.07E-05	\\
Wall-Hangings	&	7.21E-04	&	Parsley	&	6.14E-05	&	Circus Horse	&	1.87E-03	&	Rice	&	6.15E-04	\\
Space Rack	&	1.38E-06	&	Saccarin	&	6.59E-08	&	Prize Bull	&	4.22E-04	&	Tomato	&	5.09E-04	\\
Ashtray	&	5.03E-06	&	Poppy Seeds	&	1.87E-04	&	Rat	&	1.14E-02	&	Coconut	&	2.81E-04	\\
Bar	&	1.74E-08	&	Pepper	&	3.52E-04	&	Badger	&	1.18E-02	&	Mushroom	&	4.84E-04	\\
Lamp	&	2.24E-07	&	Turmeric	&	1.32E-06	&	Siamese Cat	&	3.69E-05	&	Wheat	&	1.45E-04	\\
Wall Mirror	&	1.52E-04	&	Sugar	&	2.60E-06	&	Race Horse	&	7.53E-03	&	Green Pepper	&	4.63E-03	\\
Door Bell	&	5.64E-06	&	Vinegar	&	1.50E-05	&	Fox	&	1.97E-02	&	Watercress	&	1.68E-03	\\
Hammock	&	2.32E-04	&	Sesame Seeds	&	3.41E-05	&	Donkey	&	4.90E-03	&	Peanut	&	1.16E-04	\\
Desk	&	6.37E-05	&	Lemon Juice	&	1.16E-06	&	Field Mouse	&	1.18E-02	&	Black Pepper	&	3.76E-03	\\
Refrigerator	&	1.71E-05	&	Chocolate	&	8.26E-08	&	Ginger Tom-cat	&	1.20E-03	&	Garlic	&	4.62E-05	\\
Park Bench	&	1.04E-04	&	Horseradish	&	7.46E-08	&	Husky in Slead team	&	2.62E-03	&	Yam	&	3.20E-05	\\
Waste Paper Basket	&	2.88E-06	&	Vanilla	&	1.36E-06	&	Cart Horse	&	3.42E-04	&	Elderberry	&	3.72E-04	\\
Sculpture	&	1.28E-04	&	Chives	&	6.15E-05	&	Chicken	&	1.92E-03	&	Almond	&	2.73E-06	\\
Sink Unit	&	1.89E-04	&	Root Ginger	&	6.54E-05	&	Doberman Guard Dog	&	1.15E-05	&	Lentils	&	6.94E-05	\\
\hline
\end{tabular}
\normalsize
\end{center}
{\rm Table 5d.} Calculation of the p-values corresponding to the deviation $I_{B'}$ between $\mu(B')$ and $\mu(A \ {\rm and} \ B')+\mu(A' \ {\rm and} \ B')$. By applying a Bonferroni correction procedure, the null hypothesis can be rejected for a p-value less than $\frac{0.05}{24}=2.08\cdot 10^{-3}$. 
\end{table}

\begin{table}[H]
\begin{center}
\scriptsize
\begin{tabular}{|c|c||c|c||c|c||c|c|}
\hline
\multicolumn{8}{|l|}{Deviation of $\mu(A \ {\rm and} \ B)+\mu(A \ {\rm and} \ B')+\mu(A' \ {\rm and} \ B)+\mu(A' \ {\rm and} \ B')$  from 1} \\
\hline		
(Home Furnishing, Furniture)	& p-value	&		(Spices, Herbs)	&	p-value	&	(Pets, Farmyard Animals)	&	p-value	&	(Fruits, Vegetables)	&	p-value	\\
\hline
Mantelpiece	&	4.34E-09	&	Molasses	&	4.33E-07	&	Goldfish	&	3.98E-09	&	Apple	&	1.24E-08	\\
Window Seat	&	5.12E-06	&	Salt	&	4.04E-05	&	Robin	&	1.09E-05	&	Parsley	&	1.03E-08	\\
Painting	&	1.07E-07	&	Peppermint	&	7.51E-07	&	Blue-tit	&	3.54E-07	&	Olive	&	1.44E-08	\\
Light Fixture	&	5.01E-07	&	Curry	&	3.31E-06	&	Collie Dog	&	1.55E-07	&	Chili Pepper	&	2.81E-09	\\
Kitchen Counter	&	4.63E-06	&	Oregano	&	1.95E-07	&	Camel	&	1.19E-05	&	Broccoli	&	6.15E-09	\\
Bath Tub	&	1.13E-07	&	MSG	&	5.67E-07	&	Squirrel	&	1.99E-05	&	Root Ginger	&	1.79E-06	\\
Deck Chair	&	3.04E-07	&	Chili Pepper	&	2.19E-10	&	Guide Dog for Blind	&	2.43E-05	&	Pumpkin	&	6.60E-07	\\
Shelves	&	7.84E-08	&	Mustard	&	5.67E-07	&	Spider	&	5.24E-05	&	Raisin	&	7.37E-09	\\
Rug	&	8.18E-09	&	Mint	&	9.21E-08	&	Homing Pigeon	&	5.23E-06	&	Acorn	&	8.26E-09	\\
Bed	&	3.69E-08	&	Cinnamon	&	2.42E-08	&	Monkey	&	5.05E-04	&	Mustard	&	9.61E-08	\\
Wall-Hangings	&	1.09E-06	&	Parsley	&	2.61E-07	&	Circus Horse	&	7.10E-07	&	Rice	&	3.68E-06	\\
Space Rack	&	3.77E-08	&	Saccarin	&	1.20E-07	&	Prize Bull	&	1.27E-08	&	Tomato	&	1.82E-09	\\
Ashtray	&	9.08E-08	&	Poppy Seeds	&	1.24E-06	&	Rat	&	5.07E-05	&	Coconut	&	6.54E-07	\\
Bar	&	2.27E-09	&	Pepper	&	8.33E-07	&	Badger	&	3.01E-04	&	Mushroom	&	5.77E-06	\\
Lamp	&	1.87E-08	&	Turmeric	&	5.34E-08	&	Siamese Cat	&	1.34E-07	&	Wheat	&	9.21E-08	\\
Wall Mirror	&	2.20E-09	&	Sugar	&	5.03E-07	&	Race Horse	&	5.77E-08	&	Green Pepper	&	6.54E-07	\\
Door Bell	&	1.62E-07	&	Vinegar	&	3.40E-05	&	Fox	&	9.66E-04	&	Watercress	&	6.57E-08	\\
Hammock	&	7.17E-07	&	Sesame Seeds	&	1.07E-07	&	Donkey	&	2.38E-06	&	Peanut	&	3.51E-07	\\
Desk	&	7.94E-07	&	Lemon Juice	&	4.30E-07	&	Field Mouse	&	8.23E-04	&	Black Pepper	&	9.33E-07	\\
Refrigerator	&	5.49E-07	&	Chocolate	&	5.18E-08	&	Ginger Tom-cat	&	6.79E-07	&	Garlic	&	2.51E-07	\\
Park Bench	&	1.39E-08	&	Horseradish	&	5.03E-08	&	Husky in Slead team	&	3.99E-08	&	Yam	&	1.60E-10	\\
Waste Paper Basket	&	1.38E-09	&	Vanilla	&	6.49E-08	&	Cart Horse	&	8.26E-09	&	Elderberry	&	1.60E-07	\\
Sculpture	&	7.78E-09	&	Chives	&	1.80E-08	&	Chicken	&	1.53E-06	&	Almond	&	9.09E-08	\\
Sink Unit	&	2.66E-07	&	Root Ginger	&	6.10E-08	&	Doberman Guard Dog	&	3.86E-08	&	Lentils	&	3.47E-07	\\
\hline
\end{tabular}
\normalsize
\end{center}
{\rm Table 5e.} Calculation of the p-values corresponding to the deviation $I_{ABA'B'}$ between $\mu(A \ {\rm and} \ B)+\mu(A \ {\rm and} \ B')+\mu(A' \ {\rm and} \ B)+\mu(A' \ {\rm and} \ B')$ and 1. By applying a Bonferroni correction procedure, the null hypothesis can be rejected for a p-value less than $\frac{0.05}{24}=2.08\cdot 10^{-3}$. 
\end{table}


\end{document}